\documentclass[journal]{IEEEtran}

\usepackage{amsmath,epsfig,lipsum}
\usepackage{subfigure}
\usepackage{multirow}
\usepackage{booktabs}
\usepackage{makecell}
\usepackage{cite}
\usepackage{color}
\usepackage{algorithmic}
\usepackage{graphicx}
\usepackage{textcomp}
\usepackage{xcolor}
\usepackage{placeins}
\usepackage{float}
\usepackage{tabularx,colortbl}
\usepackage{graphicx}
\usepackage{epstopdf}
\usepackage{cuted}
\usepackage{amssymb}
\usepackage{bm}
\usepackage[none]{hyphenat}
\ifCLASSINFOpdf
\else
\fi

\begin{document}\sloppy{}
	\title{Deformable Wiener Filter for Future Video Coding}
	\author{Xuewei~Meng, 
		Chuanmin~Jia, 
		Xinfeng~Zhang,~\IEEEmembership{Senior~Member,~IEEE,}
		Shanshe~Wang,~\IEEEmembership{Member,~IEEE,}
		and~Siwei~Ma,~\IEEEmembership{Senior~Member,~IEEE}

        \thanks{
		This work was supported in part by the National Natural Science Foundation of China under grant 62031013, 62025101, 61931014, 62088102, the National Key R\&D Program of China 2021YFF0900503, and in part by the High Performance Computing Platform of Peking University, which are gratefully acknowledged. The associate editor coordinating the review of this manuscript and approving it for publication was Dr. Jingning Han. (Corresponding author: Siwei Ma and Chuanmin Jia.)

        Xuewei Meng was with the National Engineering Research Center of Visual Technology, School of Computer Science, Peking University, Beijing 100871, China, and is now with the Core Media Technology, Disney Streaming. Chuanmin Jia is with the Wangxuan Institute of Computer Technology, Peking University, Beijing 100871, China. Shanshe Wang and Siwei Ma are with the National Engineering Research Center of Visual Technology, School of Computer Science, Peking University, Beijing 100871, China, and also with Information Technology R\&D Innovation Center of Peking University, Shaoxing 312000, China, and Peng Cheng Laboratory, Shenzhen 518066, China.~(e-mail: \{xwmeng, cmjia, sswang, swma\}@pku.edu.cn).
		
		Xinfeng Zhang is with the School of Computer Science and Technology, University of Chinese Academy of Sciences, Beijing 100190, China~(e-mail: xfzhang@ucas.ac.cn).}
        }
	
	\markboth{Journal of \LaTeX\ Class Files,~Vol.~XX, No.~X, December~2021}%
	{Shell \MakeLowercase{\textit{et al.}}: Bare Demo of IEEEtran.cls for IEEE Journals}
	\maketitle
	\begin{abstract}
		In-loop filters have attracted increasing attention due to the remarkable noise-reduction capability in the hybrid video coding framework. However, the existing in-loop filters in Versatile Video Coding~(VVC) mainly take advantage of the image local similarity. Although some non-local based in-loop filters can make up for this shortcoming, the widely-used unsupervised parameter estimation method by non-local filters limits the performance. In view of this, we propose a deformable Wiener Filter~(DWF). It combines the local and non-local characteristics and supervisedly trains the filter coefficients based on the Wiener Filter theory. In the filtering process, local adjacent samples and non-local similar samples are first derived for each sample of interest. Then the to-be-filtered samples are classified into specific groups based on the patch-level noise and sample-level characteristics. Samples in each group share the same filter coefficients. After that, the local and non-local reference samples are adaptively fused based on the classification results. Finally, the filtering operation with outlier data constraints is conducted for each to-be-filtered sample. Moreover, the performance of the proposed DWF is analyzed with different reference sample derivation schemes in detail. Simulation results show that the proposed approach achieves 1.16\%, 1.92\%, and 2.67\% bit-rate savings on average compared to the VTM-11.0 for All Intra, Random Access, and Low-Delay B configurations, respectively.
	\end{abstract}
	
	\begin{IEEEkeywords}
		Non-local, local, Wiener Theory, in-loop filter, versatile video coding.
	\end{IEEEkeywords}
	
	\IEEEpeerreviewmaketitle
	
	\makeatletter
	\newcommand{\thickhline}{%
		\noalign {\ifnum 0=`}\fi \hrule height 1pt
		\futurelet \reserved@a \@xhline
	}
	
	\section{Introduction}
	\IEEEPARstart {T}{he} increasing requirements for high-resolution videos and the limited transmission and memory capacity have motivated a significant interest in advanced video compression technologies~\cite{VVC,han2021technical,ma2022evolution,zhang2022implicitly,lei2022joint,meng2021spatio,lei2022coarse,meng2022parametric,lin2022high,meng2020edge}. To meet these requirements, a new video coding standard, Versatile Video Coding~(H.266/VVC)~\cite{VVC}, was developed by the Joint Video Experts Team~(JVET) since October 2017 and was finalized in July 2020. VVC achieves about 35\% bit-rate savings, compared to High Efficiency Video Coding~(H.265/HEVC)~\cite{HEVC}. Due to the widely used block-wise operation and coarse quantization in the current video coding standards, blocking and ringing artifacts have been inevitably induced in the compressed frame, which significantly degrades the objective and subjective quality. To suppress these compression artifacts, in-loop filtering algorithms have been comprehensively explored during the development of video coding standards. In-loop filters can not only improve the quality of reconstructed frames, but also provide high-quality reference frames for the subsequent pictures, yielding more precise motion compensation.

	\begin{figure}[t!]
		\begin{center}
			\noindent
			\includegraphics[width=3.45in]{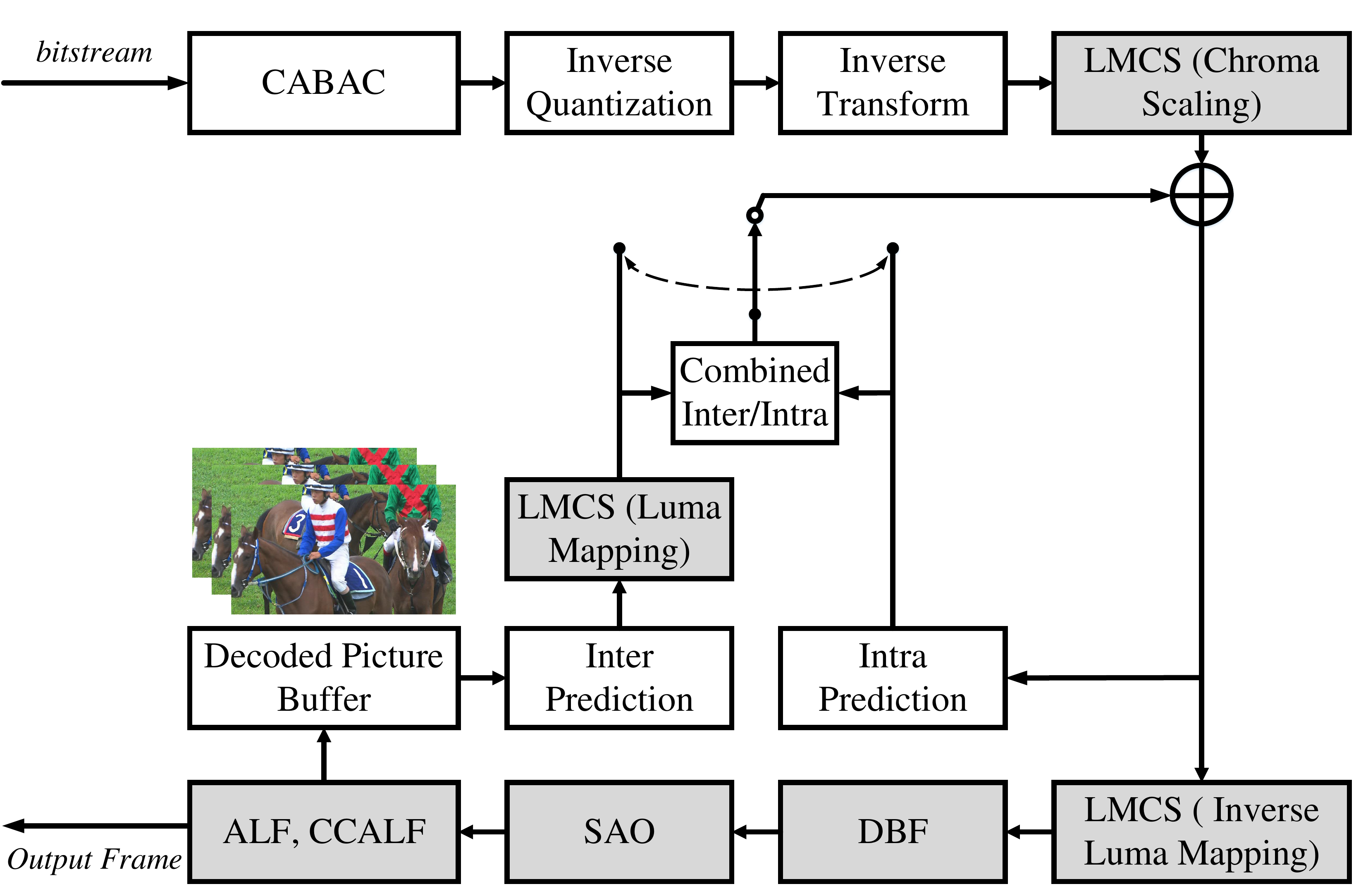}
			\caption{Illustration of the VVC decoder block diagram, with the gray boxes corresponding to in-loop filters.}
			\label{fig::VVC-framework}
		\end{center}
		\vspace{-3mm}
	\end{figure}

	There are four kinds of in-loop filters~\cite{karczewicz2021vvc} in VVC, i.e., deblocking filter~(DBF)~\cite{norkin2012hevc}, sample adaptive offset~(SAO)~\cite{SAO}, luma mapping with chroma scaling~(LMCS)~\cite{lu2020luma}, and adaptive loop filter~(ALF)~\cite{tsai2013adaptive,GALF-JVET,karczewicz2016geometry,temporal-ALF,nonlinear-ALF}. These filters are depicted in Fig.~\ref{fig::VVC-framework}. DBF applies a set of low-pass filters to the boundaries of the coding unit, the prediction unit, and the transform unit. It can remove the blocking artifacts introduced by the block-based motion compensation, and improve the subjective quality of reconstructed frames obviously. SAO is applied to reduce the mean sample distortion of a region by conditionally adding an offset to the reconstructed samples after DBF. These offsets need to be coded and transmitted to the decoder. LMCS and ALF are newly adopted by VVC. LMCS does not particularly target coding artifact reduction, but rather tries to enhance coding efficiency by better utilizing the dynamic range. ALF aims to improve the quality of reconstructed video signals by applying a spatial filtering process. Specifically, a filtered sample is calculated by taking the weighted average of its reference samples. The weight coefficients are trained in the encoder using the reconstructed samples after SAO and the original samples, following the principle of minimizing the mean square error~(MSE). After that, the filter coefficients derived in the encoder are transmitted to the decoder. To improve the adaptability, to-be-filtered samples with similar features are grouped into the same class. Samples in the same class share the same filter coefficients. 
	
	How to select the proper reference samples and group the to-be-filtered samples have been hot topics in the video compression area. For the reference sample derivation, various filter shapes have been explored, such as square shape, cross shape~\cite{tsai2013adaptive}, directional shape~\cite{zheng2011directional}, diamond shape which is adopted by VVC, and so on. These filter shapes are all based on the image local correlations. Also, many sample grouping approaches have been proposed based on the local characteristics, including the block-based sample grouping~\cite{watanabe2009loop}, local binary pattern based classification~\cite{liu2012adaptive}, sub-sampled Laplacian classification~\cite{karczewicz2021vvc} which is used in the current VVC, multiple-feature-based classification~\cite{erfurt2018multiple}, and so on. Although a variety of pre-defined fixed filter shapes can be selected by the encoder, the adaptability is still insufficient since the available filter shapes are limited in the practical codec. In addition to ALF, a cross-component adaptive loop filter~(CCALF)~\cite{CCALF,CCALF-JVET,meng2021optimizedNNN} is also introduced by VVC to improve the chroma fidelity by exploiting correlations between the luma and chroma channels. Specifically, CCALF applies a linear filter to the luma channel to determine a chroma residual that is added to a previously reconstructed chroma signal. During the development of VVC, several other in-loop filters were also discussed, such as bilateral filter~\cite{wennersten2017bilateral,strom2019bilateral}, Hadamard filter~\cite{P0078-hadamard}, and diffusion filter~\cite{rasch2018signal,rasch2020signal}.
    
	These approaches listed above mostly take advantage of the local smoothness in the image. The non-local self-similarity of images is not fully taken into account. Because the compression artifacts created by video compression are so complicated, there is still a large space to improve the quality of the compressed video. Given the excellent efficiency of the well-known non-local means~(NLM) filter~\cite{NLM} in image processing, some non-local-based in-loop filters were investigated in video coding. Matsumura \textit{et al.} first introduced NLM~\cite{NLMinHM} into HEVC through a sample-independent filtering technique to compensate for the limitations of previous in-loop filters that are exclusively based on local correlation. Following that, the well-known block-matching and 3D filtering method~(BM3D)~\cite{BM3D} was proposed by stacking non-local similar patches into 3D matrices and shrinking 3D transform coefficients of similar patches based on the image-sparse prior model to eliminate noise. In the video compression area, Zhang \textit{et al.}~\cite{NALF, zhang2015nonlocal, ma2016nonlocal} proposed a non-local adaptive loop filter~(NALF) utilizing image non-local prior knowledge, which improves the traditional low-rank-based filters by adaptively estimating compression noise for every similar patch group. After that, a structure-driven adaptive non-local filter~(SANF)~\cite{SANF} was proposed as a simplified version of NALF employing the global noise level for all similar patch groups. These non-local-based methods were also discussed in the JVET meeting~\cite{K0160, K0053, K0236}. However, the high coding complexity and the floating-point filtering operation make it unsuitable to be applied in video coding standards. Taking SANF as an example, the block matching and singular value decomposition~(SVD) process impose high computational demands. Although significant progress has been made in reducing the complexity of block matching~\cite{meng2018optimized,jia2020fast}, the SVD operation still makes hardware design problematic. Furthermore, these unsupervised non-local algorithms cannot effectively remove sophisticated compression noise, limiting content flexibility and coding performance. 
	
	Through the above analysis, both local and non-local approaches have specific drawbacks, but they can complement each other in many aspects. For instance, non-local operations will be more effective for images with sufficient repetitive details, especially screen content videos. For images with complex textures and less repetitive content, the local filter may be a preferable choice. How to make a trade-off between local and non-local filtering operations has not been explored in video compression.
	
	In this paper, we propose a deformable Wiener Filter~(DWF) that adaptively combines the local and non-local samples and determines filter coefficients using the Wiener Theory. To the best of our knowledge, DWF is the first attempt to integrate local and non-local samples in the Wiener Filter in video coding. Specifically, local adjacent samples and non-local similar samples are first derived for each sample of interest. Moreover, to improve the filtering adaptability, to-be-filtered samples are classified into specific groups based on patch-level noise and sample-level characteristics. Following that, the local and non-local reference samples are fused based on the classification result. Filter coefficients are calculated by solving the Wiener-Hopf equation and transmitted to the decoder. Finally, the filtering operation with outlier data constraints is conducted for each to-be-filtered sample. Compared to the VVC reference software VTM-11.0~\cite{VTM11.0}, the proposed DWF can achieve -1.16\%, -1.92\%, and -2.67\% coding gain for All Intra~(AI), Random Access~(RA), and Low Delay B~(LDB) configurations, respectively. For screen content videos, -2.56\%, -3.39\%, and -4.03\% performance gain can be observed for AI, RA, and LDB configurations, respectively. We also investigate the optimization implementation to reduce the computational complexity of the proposed DWF. With the fast algorithm, the proposed method shows -1.16\%, -1.78\%, and -2.46\% coding gain with 1402\%, 1175\%, and 1131\% decoding time increments.

	The remainder of this paper is organized as follows. Section II introduces the problem formulation of the Wiener-Theory-based filtering. Section III introduces the proposed DWF scheme. Section IV shows the experimental results and discussions. Finally, Section V concludes this paper.

 \begin{figure*}[t!]
		\begin{center}
			\noindent
			\includegraphics[width=7.1in]{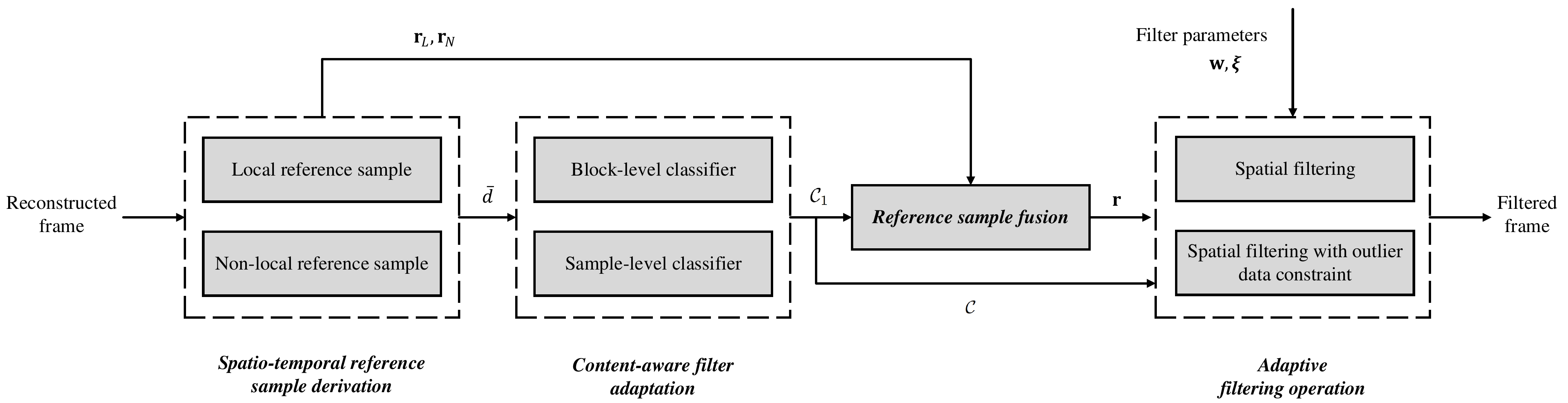}
			\caption{Illustration of the framework of the proposed DWF at the decoder.}
			\label{fig::framework}
		\end{center}
		\vspace{-3mm}
	\end{figure*}
 
	\section{Problem Formulation} \label{formulation}
	The purpose of the loop filter is to provide an estimate of the original video from the reconstructed data with compression noise. We consider the compressed video as a noisy image sequence $z: P\times T \in \mathbb{R}$. The compressed frame at time $t$ can be defined as,
	\begin{equation}
	z(t) = H\left(s(t)\right) + \zeta(t), \text{ }\text{ }\text{ }\text{ } t\in T
	\end{equation}
	\noindent where $s$ is the source~(original) video, and $\zeta$ is usually the additive Gaussian white noise. $H(\cdot)$ denotes a function representing the coding effect. $t$ is the coordinate belonging to the time domain $T \in \mathbb{Z}$. The concept of the Wiener Filter is to estimate the original signal by minimizing the MSE between the original frame $s(t)$ and the reconstructed frame $z(t)$. Let us assume the following notations.

	\begin{itemize}
	    \item [(1)] To-be-filtered sample: $z(\textbf{p}, t)$. $\textbf{p}=(x,y)$ represents the position of samples in spatial domain $P \in \mathbb{Z}^2$. $\textbf{p}$ belongs to the location set $\mathcal{S}$ of to-be-filtered samples. 
	    \item [(2)] Original sample: $s(\textbf{p}, t)$
	    \item [(3)] $N$-tap filter coefficients: $\textbf{w}=[w_{0},w_{1},w_{2},  ... ,w_{N-1}]$
	    \item [(4)] Reference samples for the to-be-filtered sample: $\textbf{r}(\textbf{p}, t) = [r_{0},r_{1},r_{2},...,r_{N-1}]$
	    \item [(5)] Filtered sample: $f(\textbf{p}, t)$
	    \begin{equation}\label{filtering}
	        f(\textbf{p}, t) = \sum_{j=0}^{N-1}w_{j}\times{r_{j}}.
	    \end{equation}
	\end{itemize}
	
	The filter coefficients $\textbf{w}$ can be derived by solving the following optimization problem,
	\begin{equation}\label{LMSE}
	\min_{\textbf{w}}  \sum_{\textbf{p} \in \mathcal{S}} \left (\textbf{w} \odot \textbf{r}(\textbf{p}, t) - s(\textbf{p},t)\right )^{2},
	\end{equation}
	\noindent where $\odot$ is the inner product. By solving the Wiener-Hopf equations, the filter coefficients $\textbf{w}$ can be calculated by,
	\begin{equation}\label{Wiener}
       \textbf{w} = \textbf{R}_{z,z}^{-1} \textbf{R}_{z,s},
    \end{equation}
    \noindent where $\textbf{R}_{z,z}$ denotes the auto-correlation matrix of the to-be-filtered samples with the location in $\mathcal{S}$. $\textbf{R}_{z,s}$ is the cross-correlation vector of the to-be-filtered and original samples with the location in $\mathcal{S}$. Based on the above illustration, $\textbf{r}$ and $\mathcal{S}$ are the main aspects of the discrete Wiener Filter. Although Eqn~(\ref{Wiener}) can derive the optimal filter coefficients for certain ${\textbf{r}}$ and $\mathcal{S}$, the optimality of ${\textbf{r}}$ and $\mathcal{S}$ are difficult to guarantee. {Generally speaking, reference sample derivation and sample grouping methods should cooperate with each other.}

	\section{Proposed Deformable Wiener Filter}
	In this section, the details of the proposed DWF are presented. The DWF framework is given first. Then we describe its main modules. Finally, the implementation study and parameter selection are provided.
	
	\subsection{Framework of Deformable Wiener Filter}
	Based on the analysis of local and non-local filters, we develop an efficient DWF to explore the performance potential of the Wiener Filter. Fig.~\ref{fig::framework} shows the framework. Specifically, we apply a Wiener Filter in the spatial domain to the reconstructed video $z$. The filtered sample is calculated by taking the weighted average of corresponding reference samples. Firstly, we derive the neighboring and non-adjacent samples in the spatial or temporal domain, based on image local and non-local similarities. Then to-be-filtered samples with similar characteristics are grouped into the same class and share the same filter coefficients for better content adaptability. After that, the reference sample vector is generated by fusing the local and non-local samples based on the block complexity. Finally, the filtering operation is conducted with the reference sample vector, sample classification results, and the decoded filter parameters. The key modules, \textit{spatio-temporal reference sample derivation}, \textit{content-aware filter adaptation}, \textit{reference sample fusion}, and \textit{adaptive filtering operation} are detailed in the following subsections.
	
	\subsection{Spatio-temporal Reference Sample Derivation} \label{noiseAna}
	Unlike the widely-used local-based reference sample derivation approach employing diamond or square filter shapes, we construct the reference sample vector $\textbf{r}$ leveraging the local samples $\textbf{r}_{L}$ and non-local samples $\textbf{r}_{N}$, as illustrated in Fig~\ref{fig::blk-matching}. The derivation details of local and non-local samples are given below.
	
	We use $R_{L}$ and $R_{N}$ to represent the number of local and non-local reference samples, respectively. Local samples are selected based on the distance towards the to-be-filtered sample in the spatial domain, which can be represented by 
	$\textbf{r}_{L}(\textbf{p}, t) = [r_{0}, r_{1}, r_{2},  ... , r_{R_{L}-1}]$. 
	
	Non-local samples are derived by block matching in the specific searching area. We first divide the current reconstructed frame $z(t)$ into patches with size $B_{s}\times B_{s}$. These patches are extracted every $m$ samples in raster scanning order, and $m$ is referred to as neighbor distance, which is used to describe the extent of overlapping. For each patch, derive $R_{N}$ similar patches in the pre-defined searching area according to the $L2$ norm of image patches. The $L2$ norm is calculated by,
	\begin{equation}
	d\left (\textbf{z}(\textbf{p}, t), \textbf{z}(\textbf{p}^{\prime}, t^{\prime})\right ) = {\left \Vert \textbf{z}(\textbf{p}, t) - \textbf{z}(\textbf{p}^{\prime}, t^{\prime})\right \Vert}^{2}_{2},
	\end{equation}
	\noindent where $\textbf{z}(\textbf{p}, t)$ is the current patch. $(\textbf{p},t)$ represents the coordinate of the upper left sample of the current patch in the reconstructed frame $z(t)$. $\textbf{z}(\textbf{p}^{\prime}, t^{\prime})$ is the candidate reference patch in the pre-defined searching area, where $\textbf{p}^{\prime} \in \Omega_{p}$ and $t^{\prime} \in \Omega_{t}$. For I-frames, the similar patches are searched in the search window~(with size $W_{1}\times W_{1}$) neighboring the current patch $\textbf{z}(\textbf{p}, t)$ on the current frame. For B-frames and P-frames, similar patches can be derived on the current frame as well as reference frames. The search window size on the reference frame is $W_{2}\times W_{2}$. When searching on the current frame, $\Omega_{p}$ can be defined as,
	\begin{equation}
	\Omega_{p} = \{(i,j)\in [x - \frac{W_{1}}{2}, x+\frac{W_{1}}{2}) \times [y - \frac{W_{1}}{2}, y+\frac{W_{1}}{2}) \},
	\end{equation}
	\noindent where $x$ and $y$ are the coordinates of the upper left sample of the current patch in the reconstructed frame. When searching on the reference frames, $\Omega_{p}$ can be defined as,
	\begin{equation}
	\Omega_{p} = \{(i,j)\in [x - \frac{W_{2}}{2}, x+\frac{W_{2}}{2}) \times [y - \frac{W_{2}}{2}, y+\frac{W_{2}}{2}) \}.
	\end{equation}
	\noindent $\Omega_{t}$ can be derived by,
	\begin{equation}
	\Omega_{t} = \begin{cases}
 	             \{t\} & \mathrm{I-frame,} \\
 	             \{t, t_{1}, t_{2},..., t_{N_{ref}}\} & \mathrm{B-frame \text{ } or \text{ } P-frame.}
	             \end{cases}
	\end{equation}
	\noindent where $N_{ref}$ is the number of reference frames. For each patch, a patch group $G$ is constructed by including the $R_{N}$ similar patches with the $R_{N}$-smallest $L2$ norm distance. Then, the group is arranged into a matrix,
	\begin{equation}
	\textit{\textbf{P}}_G = [\textbf{z}_{G}^{(0)}, \textbf{z}_{G}^{(1)},..., \textbf{z}_{G}^{(R_{N}-1)}],
	\end{equation}
	where $\textit{\textbf{P}}_G$ is a matrix with the size of ${B_{s}}^{2}\times R_{N}$ by arranging every patch in the group $G$ as a column vector. For the to-be-filtered sample with location $j$ in the patch, non-local reference samples can be denoted by $\textbf{r}_{N}(\textbf{p}, t) = [\textbf{z}_{G}^{(0,j)}, \textbf{z}_{G}^{(1,j)}, ..., \textbf{z}_{G}^{(R_{N}-1,j)}]$. The reference sample vector of luma component is denoted by $\textbf{r}(\textbf{p}, t) = [r_{0},r_{1},r_{2},...,r_{R_{L}-1},\textbf{z}_{G}^{(0,j)},\textbf{z}_{G}^{(1,j)},...,\textbf{z}_{G}^{(R_{N}-1,j)}]$.

    {For chroma component, the reference sample derivation process is similar to that of the luma component. For local reference sample derivation process, the number of chroma local reference sample is $R_{L} / 2$. For non-local reference sample derivation process, the non-local similar blocks of chroma component are only searched in the search window~(with size $\frac{W_{1}}{2} \times \frac{W_{1}}{2}$) on the current frame. The number of chroma non-local reference sample is $R_{N} / 2$. The reference sample vector of chroma component can be denoted by $\textbf{r}(\textbf{p}, t) = [r_{0},r_{1},r_{2},...,r_{R_{L} / 2-1},\textbf{z}_{G}^{(0,j)},\textbf{z}_{G}^{(1,j)},...,\textbf{z}_{G}^{(R_{N} / 2 - 1,j)}]$.}
	
	\begin{figure}[t!]
		\begin{center}
			\noindent
			\includegraphics[width=3.4in]{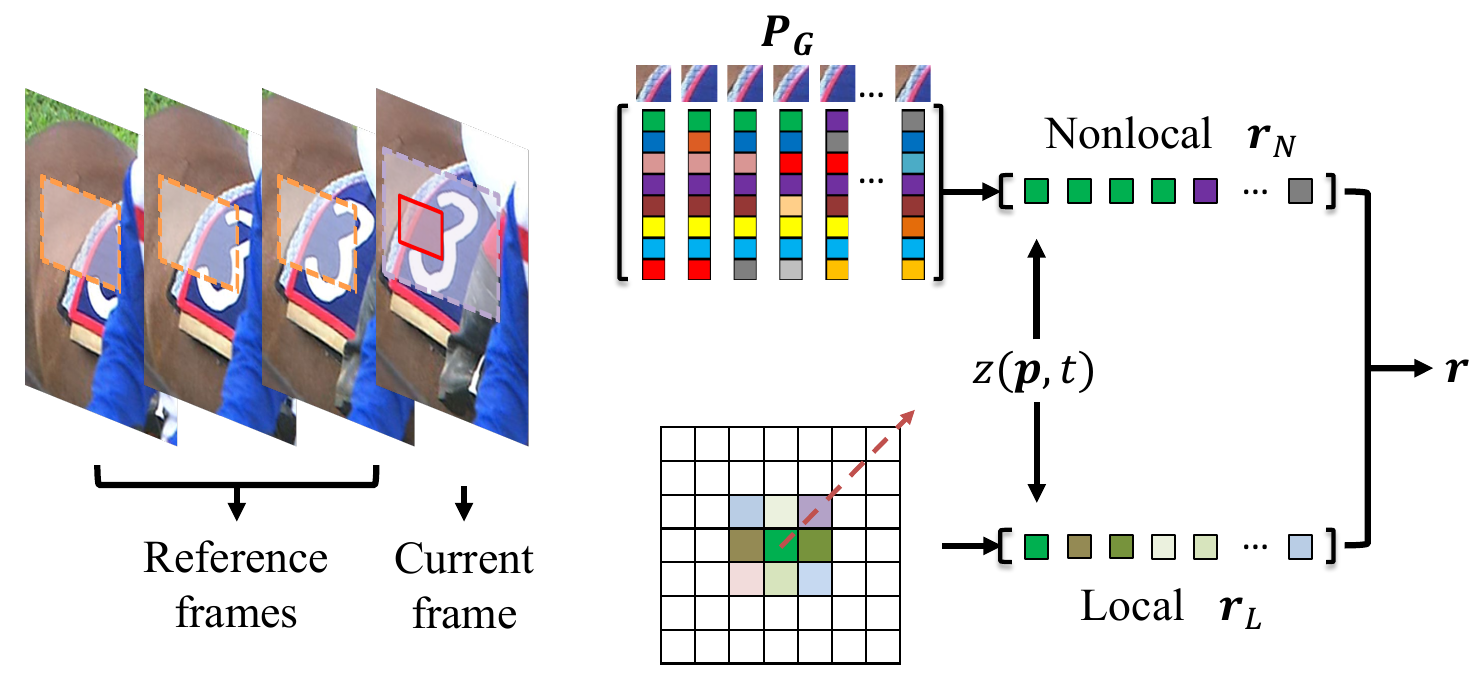}
			\caption{Illustration of the \textit{spatio-temporal reference sample derivation}. The red solid block is the current patch. The purple dash block represents the searching area on the current frame. The yellow dash blocks are the search area on the reference frames. $\textbf{P}_{G}$: similar patch group; $z(\textbf{p}, t)$: to-be-filtered sample; $\textbf{r}_N$: non-local reference samples; $\textbf{r}_L$: local reference samples; $\textbf{r}$: reference sample vector.}
			\label{fig::blk-matching}
		\end{center}
		\vspace{-3mm}
	\end{figure}

 \begin{figure*}[!t]
		\begin{center}
			\noindent
			\subfigure[]{
				\includegraphics[width=2.25in]{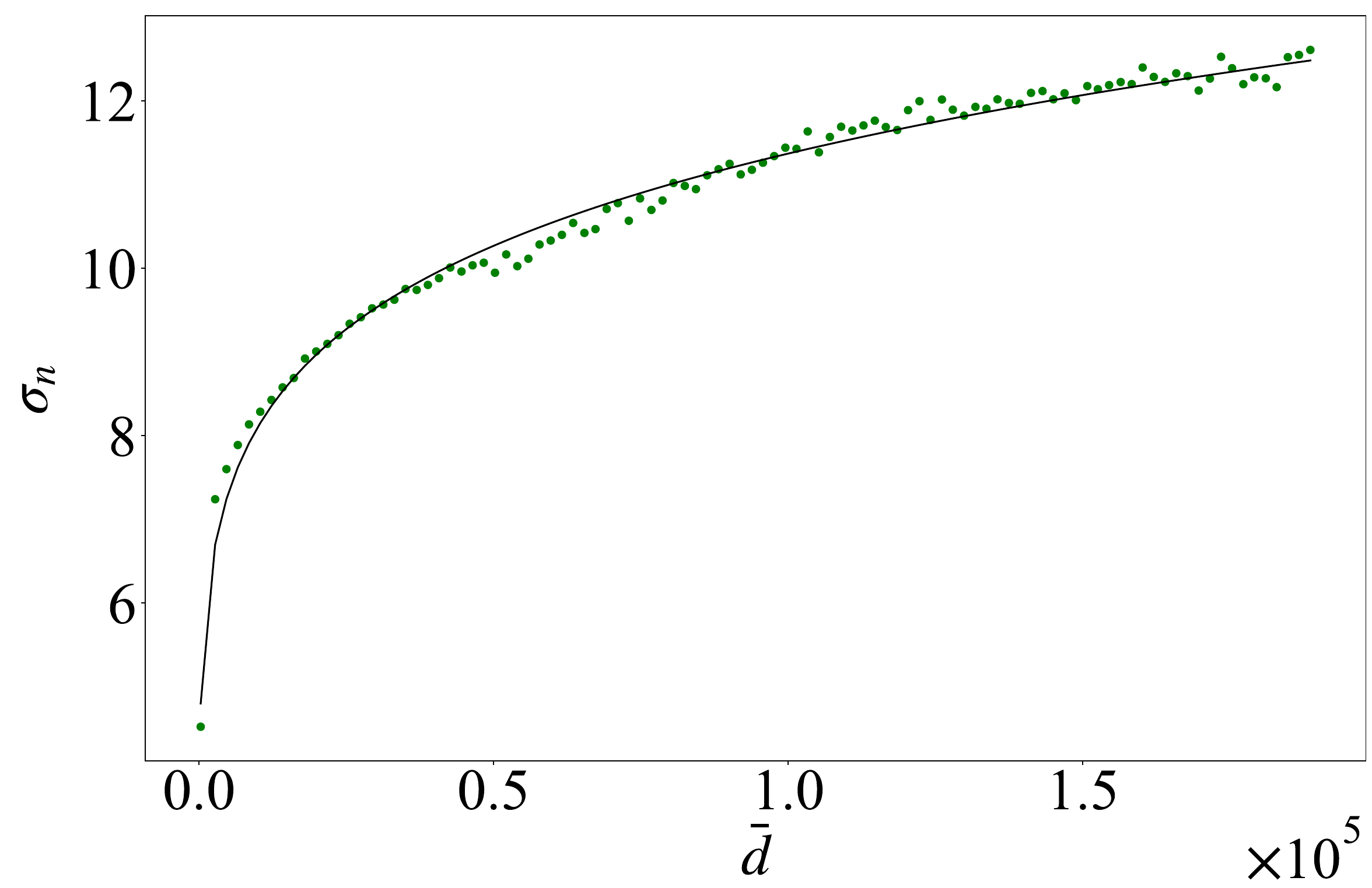}
			}
			\subfigure[]{
				\includegraphics[width=2.25in]{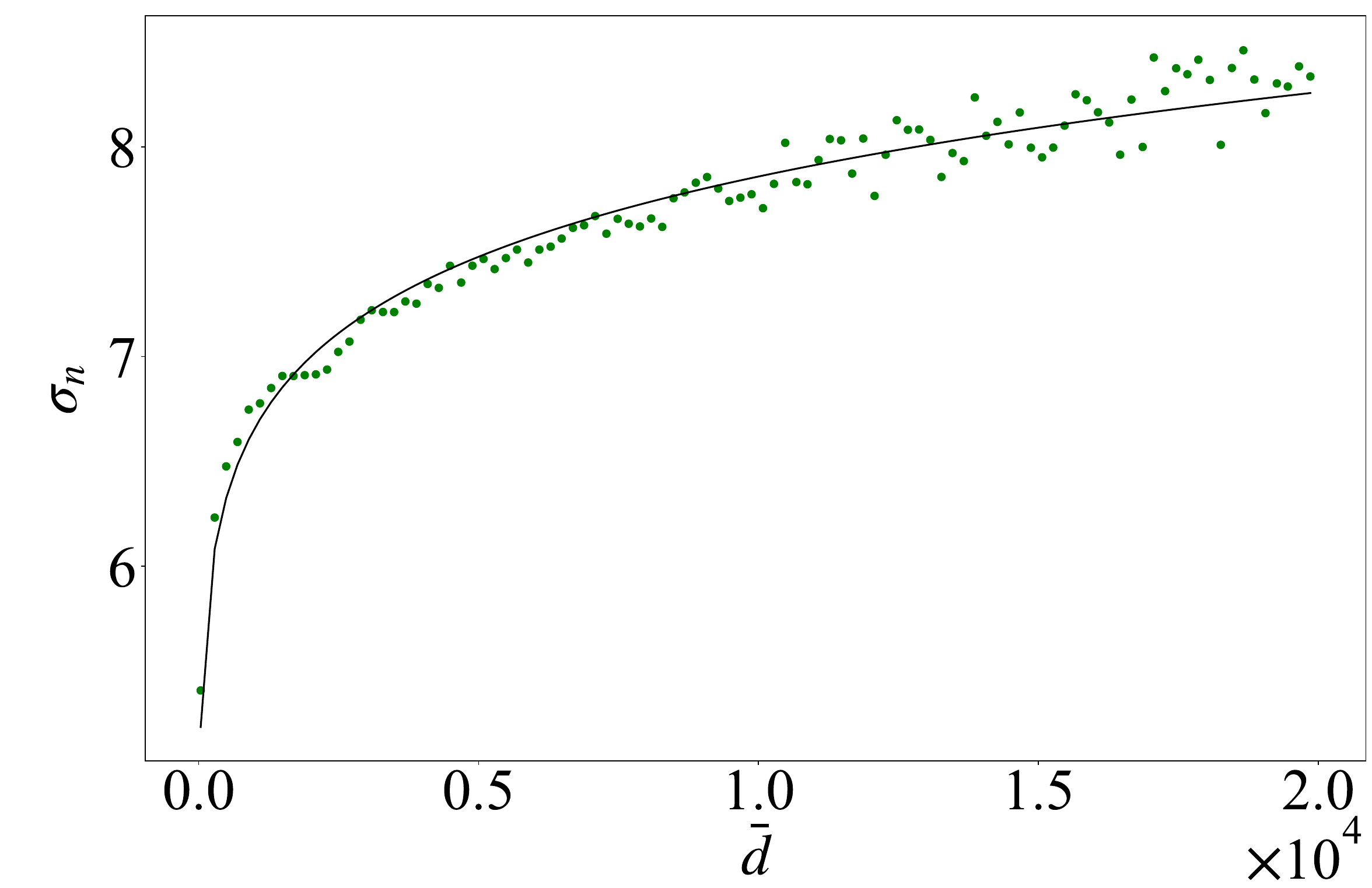}
			}
			\subfigure[]{
				\includegraphics[width=2.25in]{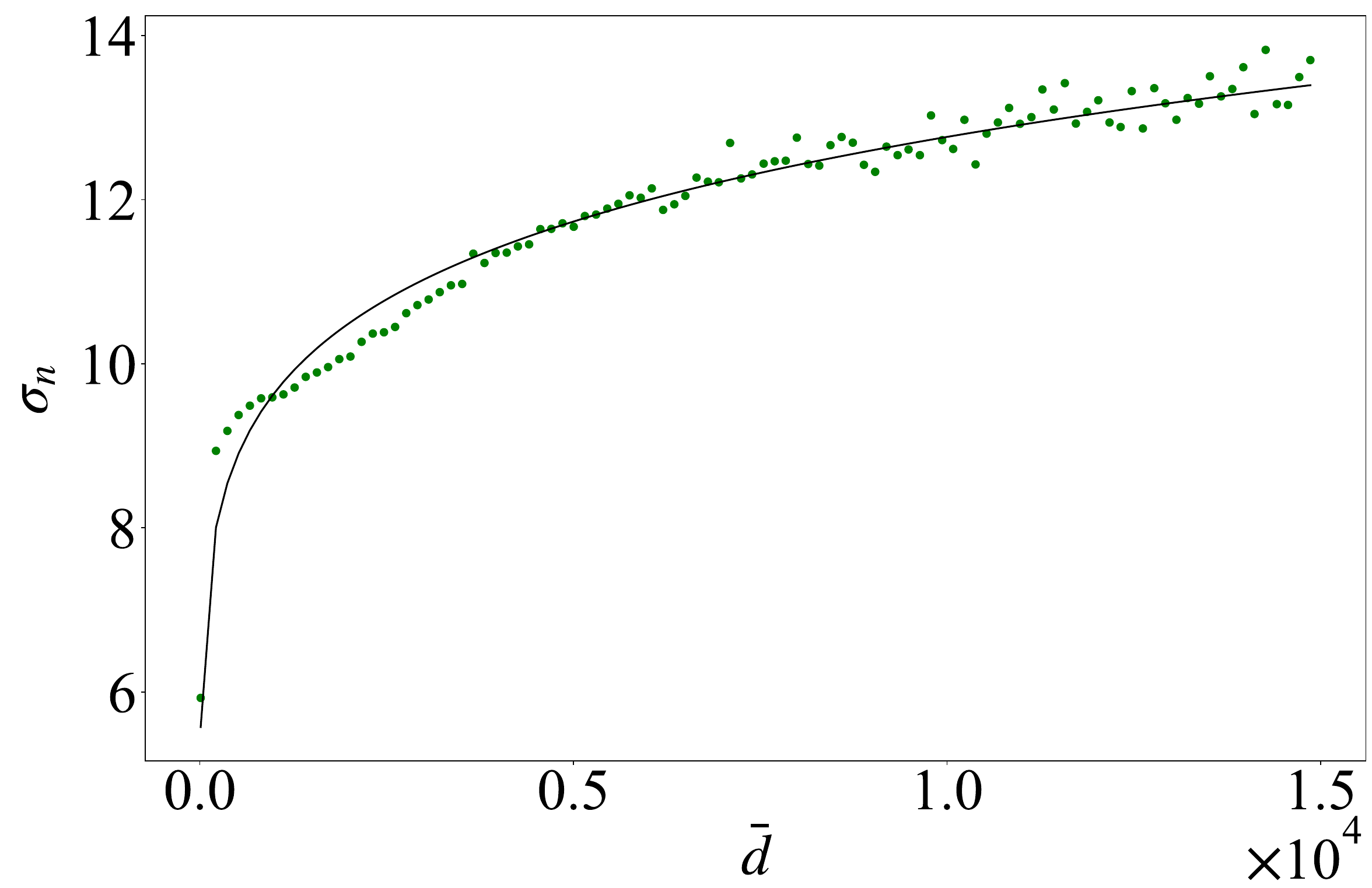}
			}
			\subfigure[]{
				\includegraphics[width=2.25in]{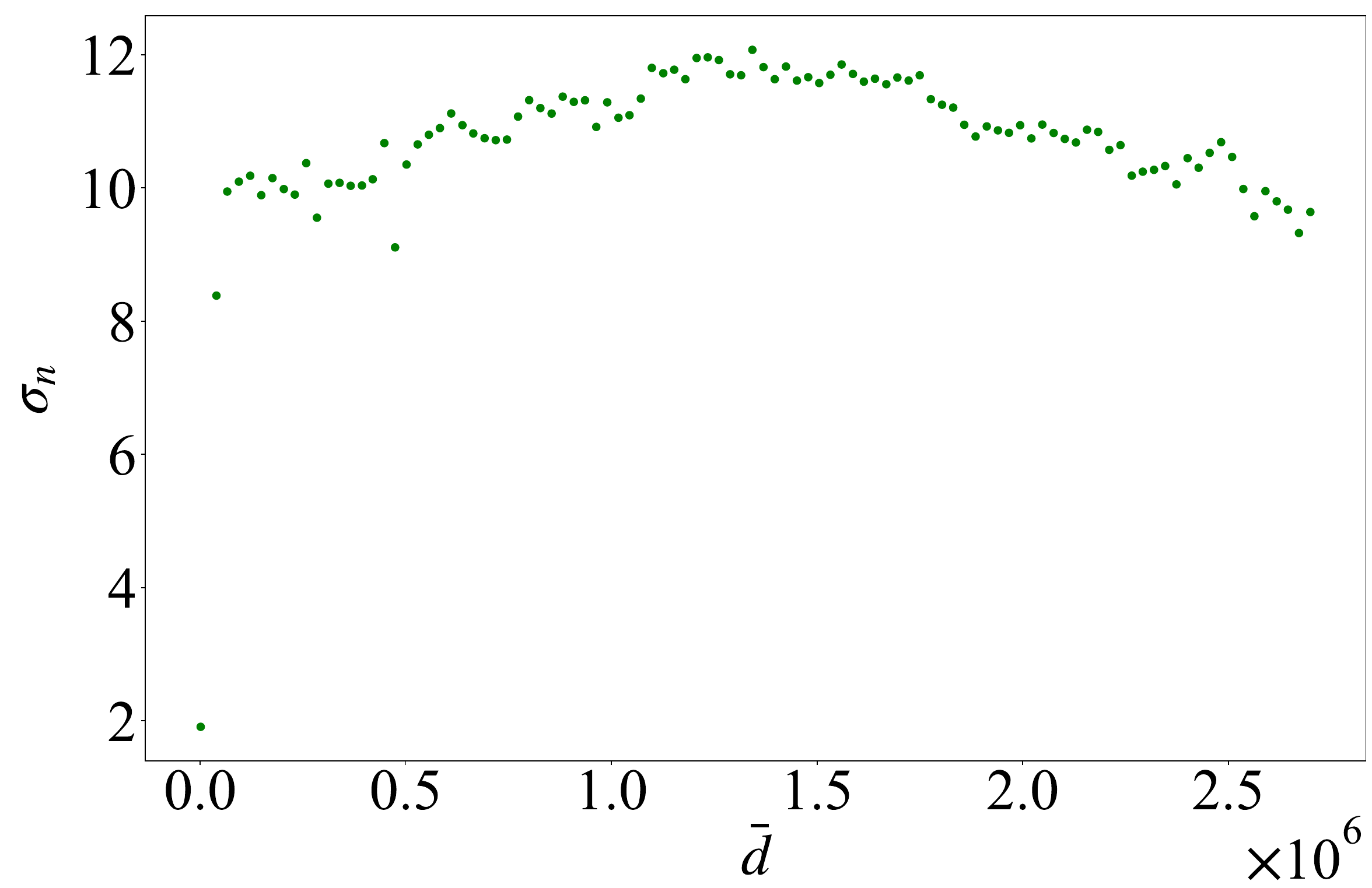}
			}
			\subfigure[]{
				\includegraphics[width=2.25in]{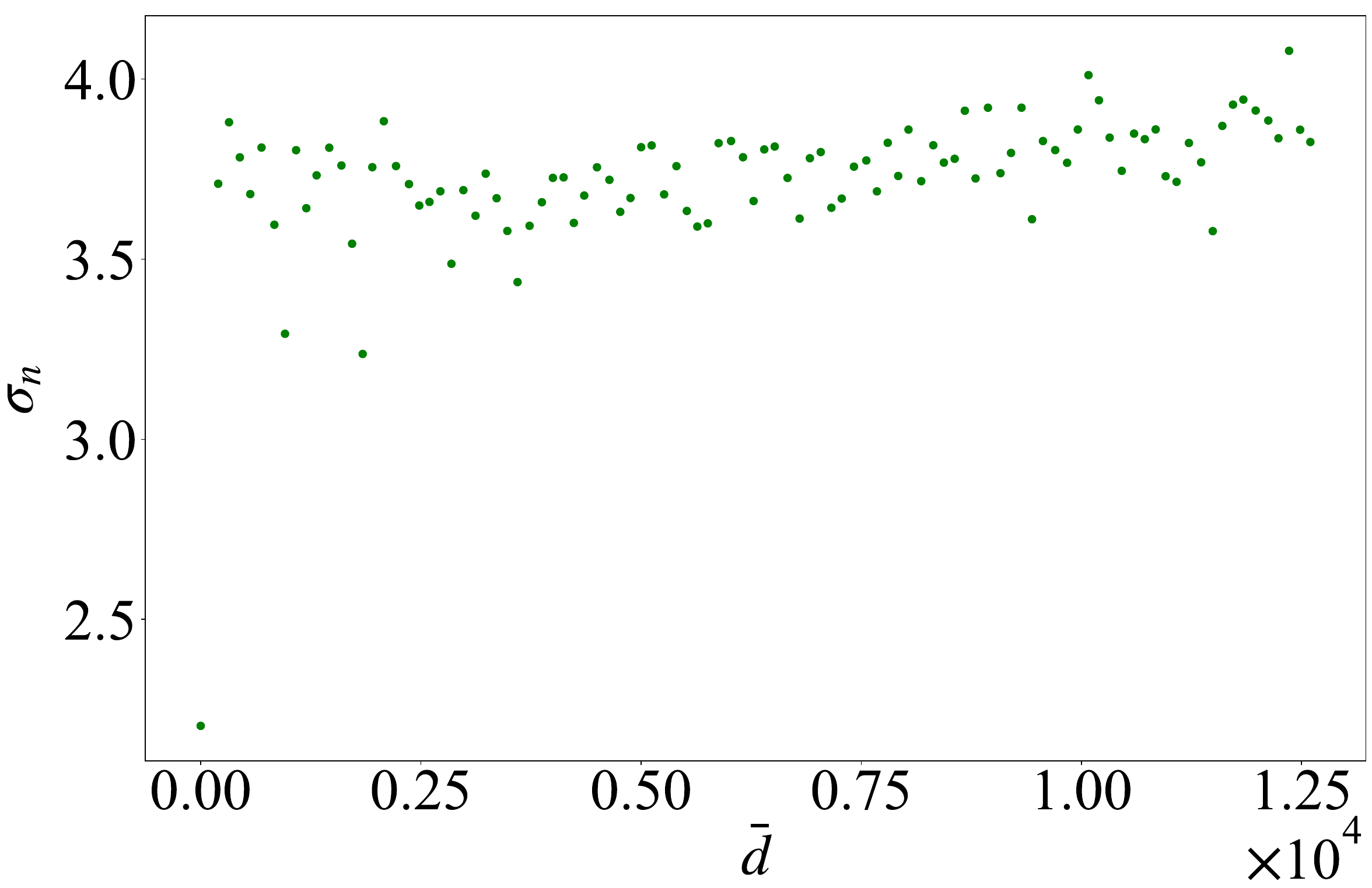}
			}
			\subfigure[]{
				\includegraphics[width=2.25in]{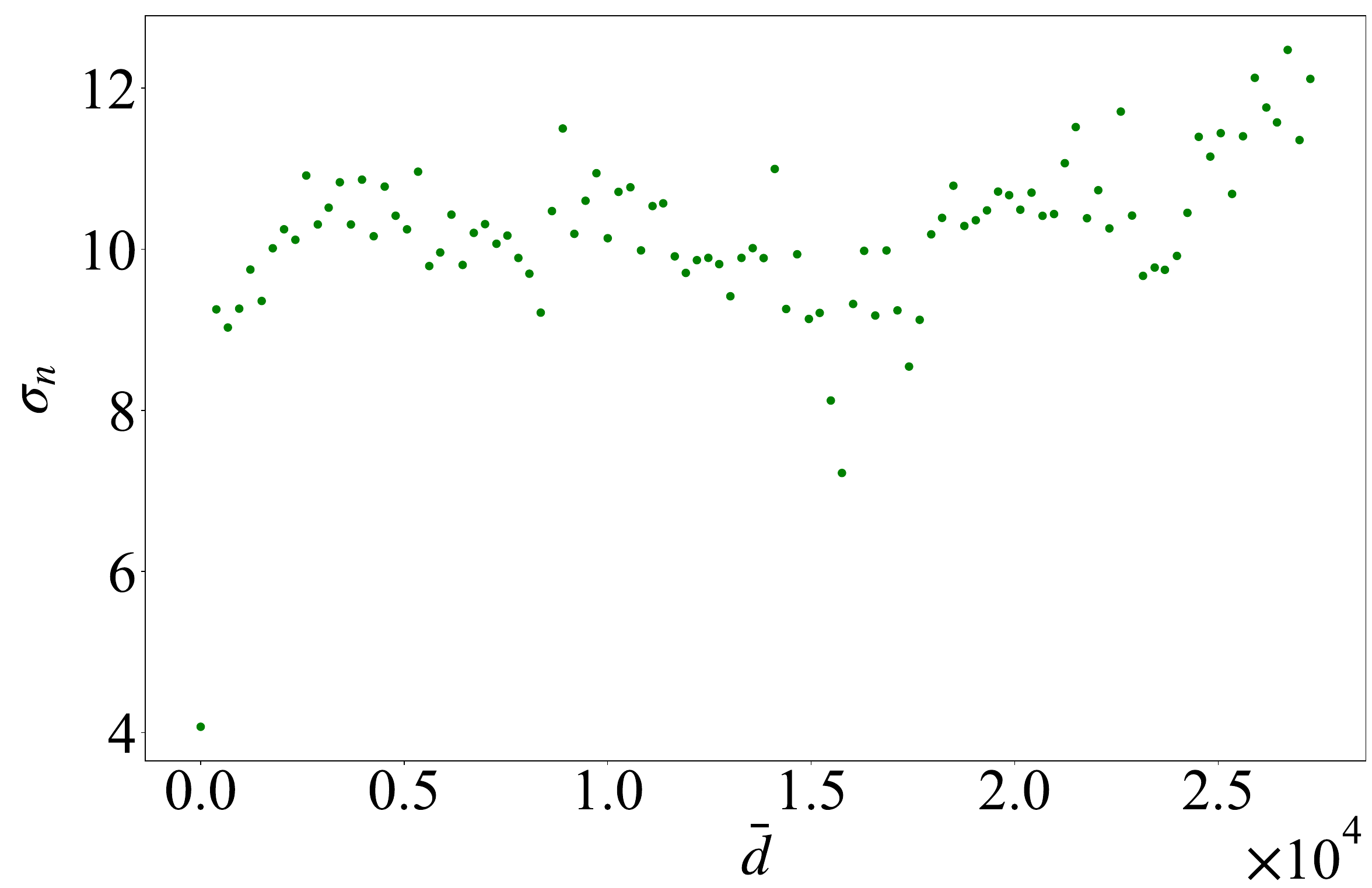}
			}
			\caption{Illustration of the relation between compression noise $\sigma_{n}$ and $L2$ norm $\bar{d}$ of similar image patches at QP = 32 under AI, RA, and LDB configurations. (a) AI, natural sequence; (b) RA, natural sequence; (c) LDB, natural sequence; (d) AI, screen content sequence; (e) RA, screen content sequence; (f) LDB, screen content sequence.}
			\label{fig::noise_SSD}
		\end{center}
		\vspace{-5mm}
	\end{figure*}
 
	\subsection{Content-aware Filter Adaptation} \label{classification} 
	As described in \cite{zhang2017high}, sample classification does influence the coding efficiency of ALF significantly. However, sample classification is not an independent problem. It should be incorporated with the reference sample derivation method. Due to the proposed deformable reference sample derivation scheme, the widely utilized local-based classification approach cannot meet our needs. Here, we propose a content-aware filter adaptation approach that combines the block-level and sample-level characteristics and adaptively applies alternative classification methods for various video content.
	
	{We first assume that patches with comparable noise levels can be grouped together and denoised by the same filter.} However, video compression noise is too complicated to be simply formulated by spatial-domain-independent noise like Gaussian noise, salt-and-pepper noise, etc. In general, compression noise is highly correlated to video content and quantization parameter~(QP). Due to the block-wise motion compensation, blocks with complex structures and rich textures tend to have larger residuals and higher noise levels. The standard deviation is a common criterion to describe the content complexity of blocks. However, the standard deviation calculation for each patch may cause a high computational burden for the decoder. Moreover, the content complexity is usually related to the similarity between the current patch and its reference patches, i.e., high-complexity blocks usually contain less similar blocks leading to a larger $L2$ norm, and vice versa. In view of this, we investigate the empirical model between the noise level and $L2$ norm between similar patches and the current patch. For the current patch ${\textbf{z}}(\textbf{p},t)$, we utilize $\bar{d}$ to indicate the average of $L2$ norms between the current patch and its similar patches,
	\begin{equation}
	    \bar{d} = \frac{1}{R_{N}}\sum_{i=0}^{R_{N}-1}{d({\textbf{z}}(\textbf{p},t), {\textbf{z}}_{G}^{(i)})},
	\end{equation}
	\noindent where $R_{N}$ is the number of non-local reference samples. ${\textbf{z}}_{G}^{(i)}$ represents the $i$-th similar patch. The noise level is represented by the standard deviation of the error signal as \cite{NALF}. The standard deviation is calculated by, 
	\begin{equation}
	    \sigma = \frac{1}{L}\sum_{l=1}^{L} \sqrt{\sum_{i=1}^{w}\sum_{j=1}^{h}\frac{\left(I_{l}(i,j)-\mu_{l}\right)^2}{w\times h}},
	\end{equation}
    \begin{equation}
        \mu_{l} = \frac{1}{w\times h}\sum_{i=1}^{w}\sum_{j=1}^{h}I_{l}(i,j),
    \end{equation}
	\noindent where $I_{l}(i,j)$ denotes the sample with location $(i,j)$ in image patch $I_l$. $w(w=6)$ and $h(h=6)$ are the width and height of patch $I_l$. $L$ is the number of image patches $I_l$ in a frame. We use $\sigma_{n}$ to represent the standard deviation of compression noise. For the calculation of $\sigma_{n}$, $I_{l}$ is extracted from the error signal without overlap. The error signal $e$ is the absolute difference between original video $s$ and reconstructed video $z$, which is calculated by,
	\begin{equation}
        e = |s - z|.
    \end{equation}

 \begin{figure*}[!t]
		\begin{center}
			\noindent
			\subfigure[]{
				\includegraphics[width=1.655in]{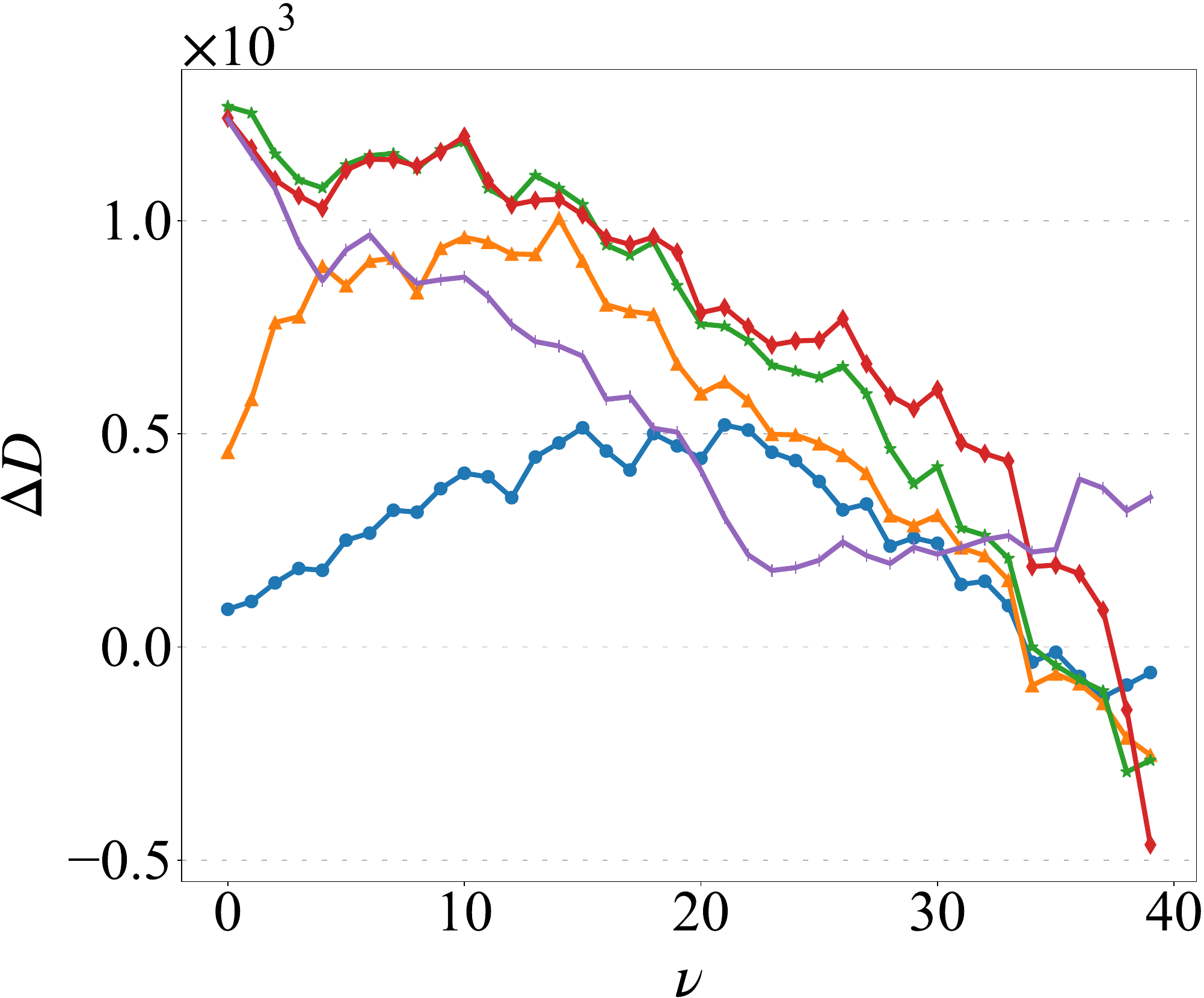}
			}
			\subfigure[]{
				\includegraphics[width=1.655in]{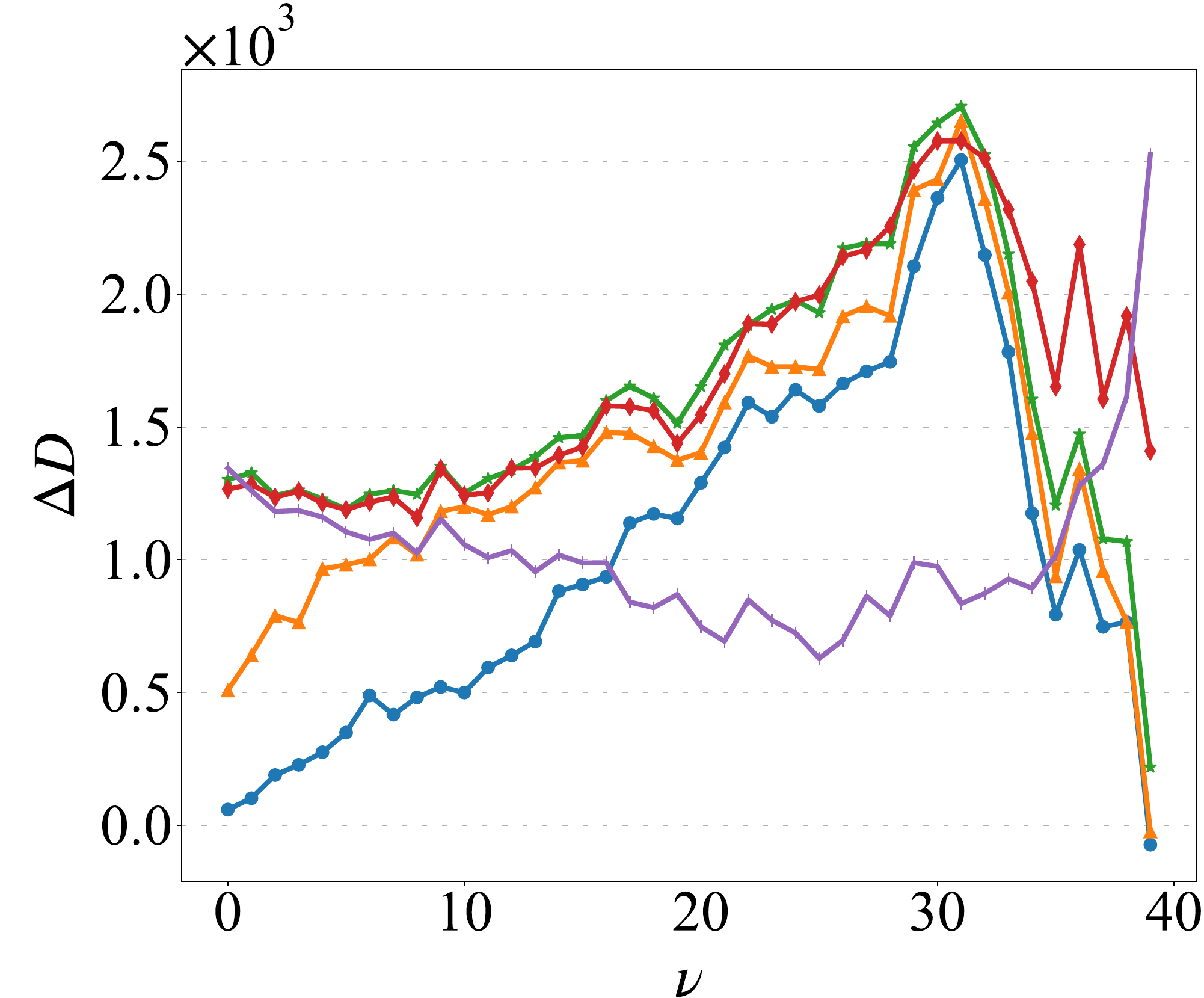}
			}
			\subfigure[]{
				\includegraphics[width=1.655in]{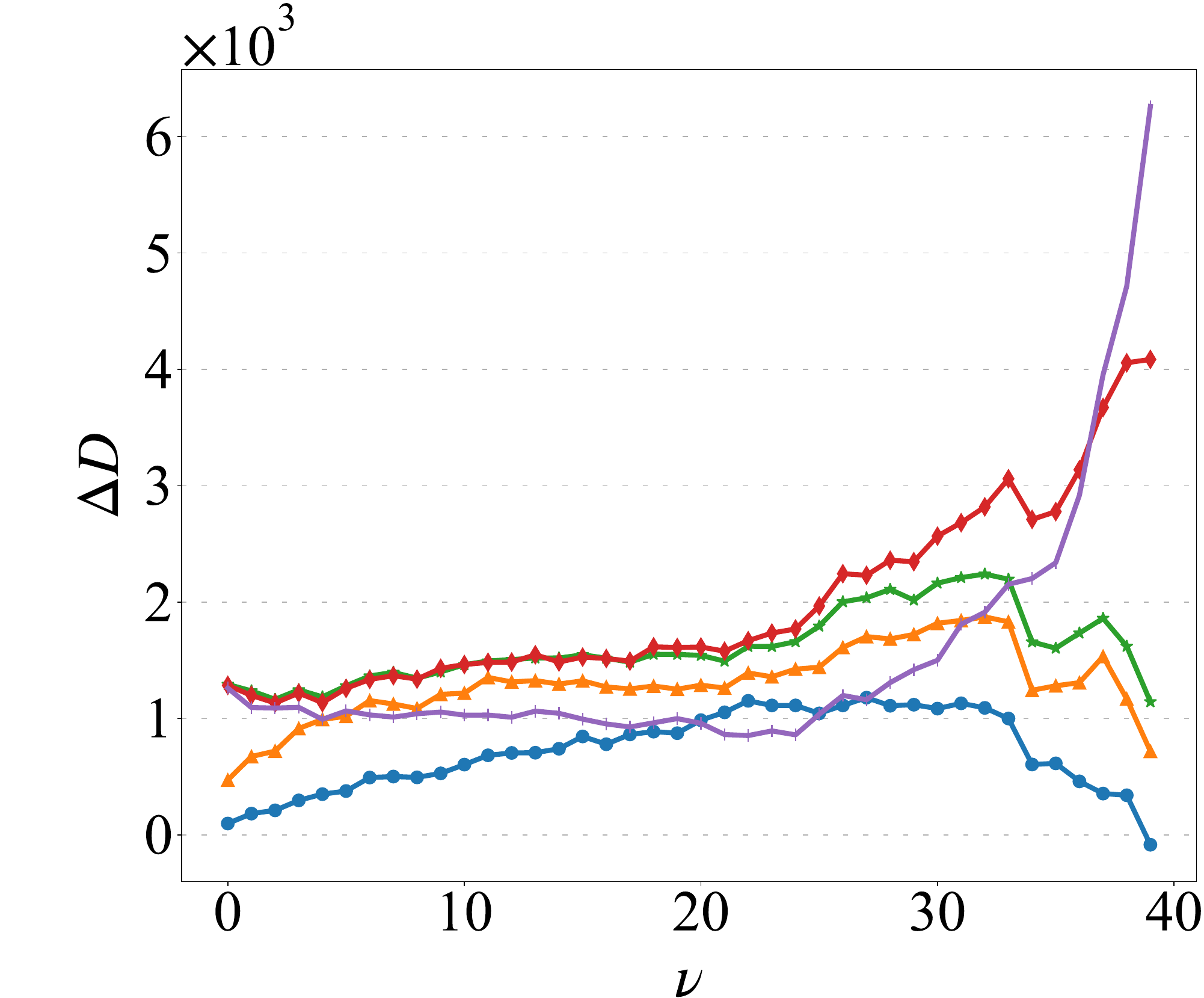}
			}
			\subfigure[]{
				\includegraphics[width=1.655in]{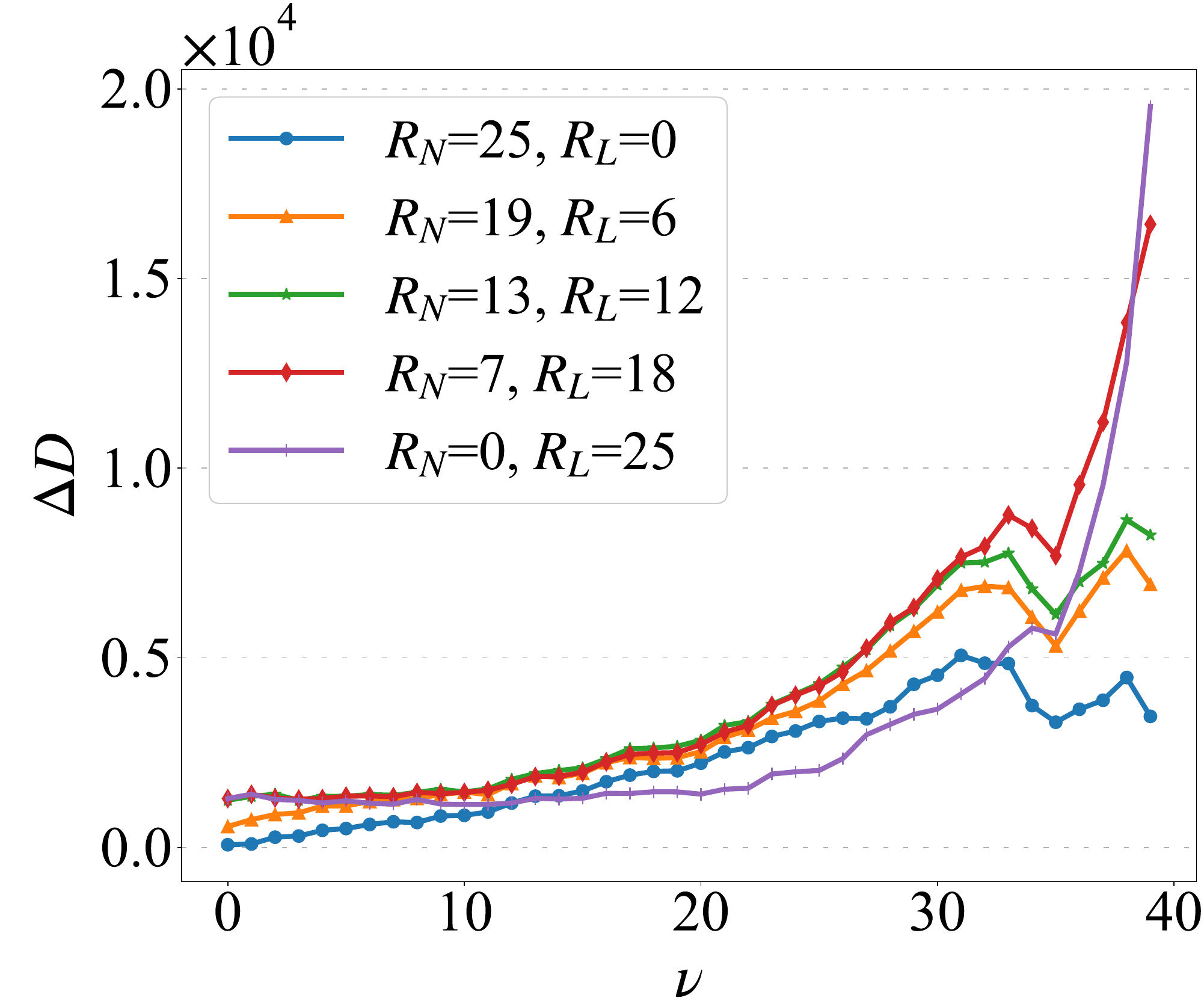}
			}
			\caption{Illustration of the influence of reference sample fusion on distortion reduction $\Delta D$ versus patch complexity $\nu$. (a) ALF ON, QP = 27; (b) ALF ON, QP = 32; (c) ALF OFF, QP = 27; (d) ALF OFF, QP = 32.}
			\label{fig::ref_ana_anchor}
		\end{center}
		\vspace{-3mm}
	\end{figure*}

	We conduct the experiment on the sequences compressed by VTM-11.0 under AI, RA, and LDB configurations. Natural sequences selected in the experiment are \textit{BQSquare}, \textit{BasketballPass}, \textit{BQMall}, \textit{BasketballDrill}, \textit{Johnny}, and \textit{FourPeople}. Screen content videos, \textit{ChineseEditing} and \textit{Console}, are also included. For each sequence, the first one second of each sequence is encoded at QP \{22, 23, 24, 25, ..., 42\}. As shown in Fig.~\ref{fig::noise_SSD}, the relationship between compression noise level $\sigma_{n}$ and average $L2$ norm $\bar{d}$ can be well fitted by a power function~($y = ax^{b}$) at certain QP and configuration for natural sequences. {Hence, the noise level can be estimated based on the average $L2$ norm $\bar{d}$ of reconstructed image patches.} As the $L2$ norm between similar patches and the current patch need to be calculated in the \textit{spatio-temporal reference sample derivation} process, estimating the noise level based on the $\bar{d}$ rarely introduces additional complexity.

	In view of this, we design a block-level classifier based on the $\bar{d}$, which is formulated by,
	\begin{equation}
	\mathcal{C}_{1} = i \text{ } \mathrm{if} \text{ } t_{i} < \bar{d} \leq t_{i+1} \text{ } \mathrm{for} \text{ } i \in\{0, 1, ..., K_{1}-1\},
	\end{equation}
	\noindent where $K_{1}$ is the number of classes of the block-level classifier. $\{t_{0}, t_{1}, ..., t_{K_{1}-1}\}$ is the offline trained threshold for different frame types and QPs. 
	
	As the block-level classifier may be insufficient to describe the characteristics of reconstructed samples, we design a sample-level classifier, which is formulated by,
	\begin{equation}
	\mathcal{C}_{2} = \left\lfloor \frac{K_2}{2^{BD}}\times z(\textbf{p},t) \right\rfloor,
	\end{equation}
	\noindent where $BD$ is the bit depth of the internal video of the codec. $K_{2}$ represents the number of classes of the sample-level classifier. 
	
	Natural sequences usually contain complex textures and are governed by multiple specific local features. To better capture these characteristics, we apply a classification method by the product of classifier $\mathcal{C}_{1}$ and $\mathcal{C}_{2}$,
	\begin{equation}
	\mathcal{C} = (\mathcal{C}_{1},\mathcal{C}_{2}) \in \{0,1,...,K_{1}-1\} \times \{0,1,...,K_{2}-1\},
	\end{equation}
	\noindent where $\mathcal{C}$ is the classification result of the current sample $z(\textbf{p},t)$. The number of classes $K$ is $K_{1} \times K_{2}$~($K_{1}=5$, $K_{2}=8$, $K=40$).
	
    We also investigate the diverse characteristics of natural sequences and screen content videos. As depicted in Fig.~\ref{fig::noise_SSD}, it is clear that the relation between the compression noise $\sigma_{n}$ and $\bar{d}$ of screen content videos is difficult to be well fitted by a simple function. Considering the relatively simple video content, the classification method for screen content is formulated by, 
    \begin{equation}
	\mathcal{C} = \left\lfloor \frac{X}{2^{BD}}\times z(\textbf{p},t) \right\rfloor,
	\end{equation}
	\noindent where $X$ is the number of classes, which is empirically set to be 32 for screen content videos. The detection of screen content frames is based on the method proposed in JVET~\cite{M0255}. 
	
	In the encoder, $K$ or $X$ filters are trained at most. For a better trade-off between filter coefficient signaling overhead and distortion reduction, the filters are adaptively merged in the encoder based on the Rate-Distortion Cost~(RDCost). The RDCost $J_{RD}$ is calculated by,
	\begin{equation}
	    J_{RD} = D + \lambda \times R,
	\end{equation}
	\noindent where $D$ is the distortion between the original image and the reconstructed image filtered by DWF. $R$ represents the bits introduced by the signaling of filtering parameters in DWF. $\lambda$ controls the trade-off between $D$ and $R$, and is usually determined by QP and frame type. 
 
    {In one extreme, all classes have their own filter. In the other extreme, all classes share the same filter. The merging process is based on the greedy algorithm, in which all the possible merging schemes are conducted. And the merging scheme with the smallest $J_{RD}$ is selected by the encoder. Its corresponding filter coefficients and the relationship between DWF filters and classes are transmitted to the decoder. In the decoder, there is no need to conduct the merging process. The decoder only needs to parse these syntaxes, find the corresponding filter for each class, and conduct the DWF filtering process. }

	\subsection{Reference Sample Fusion}
	As we mentioned before, local and non-local approaches can complement each other. How to combine these local and non-local samples is a crucial issue. In this paper, we construct the reference sample vector $\textbf{r}$ by adjusting the number of samples in $\textbf{r}_N$ and $\textbf{r}_L$, i.e., $R_{N}$ and $R_{L}$. The adjustment scheme is based on the block-level complexity, which is off-line investigated using the method below.

	Based on the observation in Fig.~\ref{fig::noise_SSD}, the noise level can be estimated by the average of $L2$ norm between the current patch and its similar patches. And the high-complexity blocks tend to have fewer similar blocks and larger $L2$ norms. We assume the optimal reference sample vector is related to the similarity of the current patch and reference patches. Hence, in the offline experiment, to-be-filtered samples are first classified into multiple classes based on the average $L2$ norm $\bar{d}$. In particular, $\bar{d}$ is calculated for each patch. The patch size $B_{s}$ is set to be 6. And the patch step $m$ is also 6. Then these $\bar{d}$ values of patches in the whole frame are sorted. Finally, patches are divided into equal numbers according to the sorted results, i.e., patches with similar $\bar{d}$ are classified into the same group. We use $\nu$ ($\nu \in \{0,1,2,...,39\}$) to represent the group index. $\mathcal{S}_{\nu}$ is the location set of samples in the group $\nu$. $\nu$ can also represent the patch complexity. The larger the $\nu$, the higher the patch complexity. For each group, the filter coefficients are trained by minimizing the MSE between the reconstructed and original samples using Eqn.~(\ref{LMSE}). And the filtered samples are calculated by the weighted average of reference samples using Eqn.~(\ref{filtering}). 
	
	We use the distortion reduction as the effect of the filter, which is calculated by,
	\begin{equation}
	\Delta D_{\nu} = \sum_{\textbf{p} \in \mathcal{S}_{\nu}} \left ((z(\textbf{p}, t) - s(\textbf{p},t))^{2} - (f(\textbf{p}, t) - s(\textbf{p},t))^{2} \right ).
	\end{equation}
	\noindent The test sequences are \textit{BQSquare}, \textit{BasketballPass}, \textit{BQMall}, \textit{BasketballDrill}, \textit{Johnny}, and \textit{FourPeople}. For each sequence, the first one second is encoded at QP \{22,27,32,37\}. As shown in Fig.~\ref{fig::ref_ana_anchor}, it can be seen that local samples are not always the best choice for Wiener Filter. The combination of local and non-local samples shows better performance in most cases.
	
	Specifically, when ALF is enabled, all-local reference samples~($R_{N}=0, R_{L}=25$) tend to show better performance for both high-complexity and low-complexity blocks. And all-non-local reference samples~($R_{N}=25, R_{L}=0$) are better to be used by median-complexity blocks. The reason for this is that when a block contains a lot of complex textures, it's usually difficult to derive adequate similar blocks or the similarity of selected blocks is relatively low. In this phenomenon, applying non-local operations directly may result in additional artifacts, and local operations may be an appropriate choice. For a flat area with a low noise level, too many identical blocks can be chosen for the filtering of the current block, making the purely non-local operation meaningless. Compared to other parameters, $R_{N}=7, R_{L}=18$ and $R_{N}=13, R_{L}=12$ shows better performance for low-complexity and median-complexity blocks. When ALF is disabled, all-local samples always exhibit a larger distortion reduction compared to all-non-local samples. While the combination of local and non-local samples also shows better performance, especially for low-complexity and median-complexity blocks. {Hence, we can conclude that the reference sample derivation is an important aspect that can influence the filtering effect significantly.}
    
    Based on the above observation, we adjust the number of local and non-local reference samples based on the block complexity. For blocks with high and low complexity, more local reference samples are used. For median-complexity blocks, non-local samples are better choices. By incorporating the block-level classifier described in Section \ref{classification}, the number of local and non-local reference samples $R_{L}$ and $R_{N}$ are determined for each complexity level. As the complexity distribution is different for different resolutions, the $R_{L}$ and $R_{N}$ are refined based on the image resolution. We use $N_{pix}$ to represent the number of samples in a frame. A look-up table is used to store the selected $R_{L}$, which is depicted in Table \ref{table::lut}. The number of non-local reference samples $R_{N}$ is $N - R_{L}$, where $N$ is the filter tap. 
	
	\begin{table}[t!]
		\centering
		\begin{center}
			\caption{Illustration of the setting of $R_{L}$ in the proposed DWF.} \label{table::lut}
			\begin{tabular}{c c c c}
				\thickhline
				\hline
				         &$\mathcal{C}_1 = 0$     & $0 < \mathcal{C}_1 < 4$     & $\mathcal{C}_1 \geq 4$\\
			    \hline
				$N_{pix} \geq 1920\times1080$   & 19       & 17           & 19   \\
				$N_{pix} \leq 1280\times720$    & 3        & 0            & 3    \\
				otherwise                       & 15       & 13           & 15   \\
				\thickhline
			\end{tabular}
		\end{center}
		\vspace{-3mm}
	\end{table}
	
	\begin{figure*}[!t]
		\begin{center}
			\noindent
			\subfigure[]{
				\includegraphics[width=2.2in]{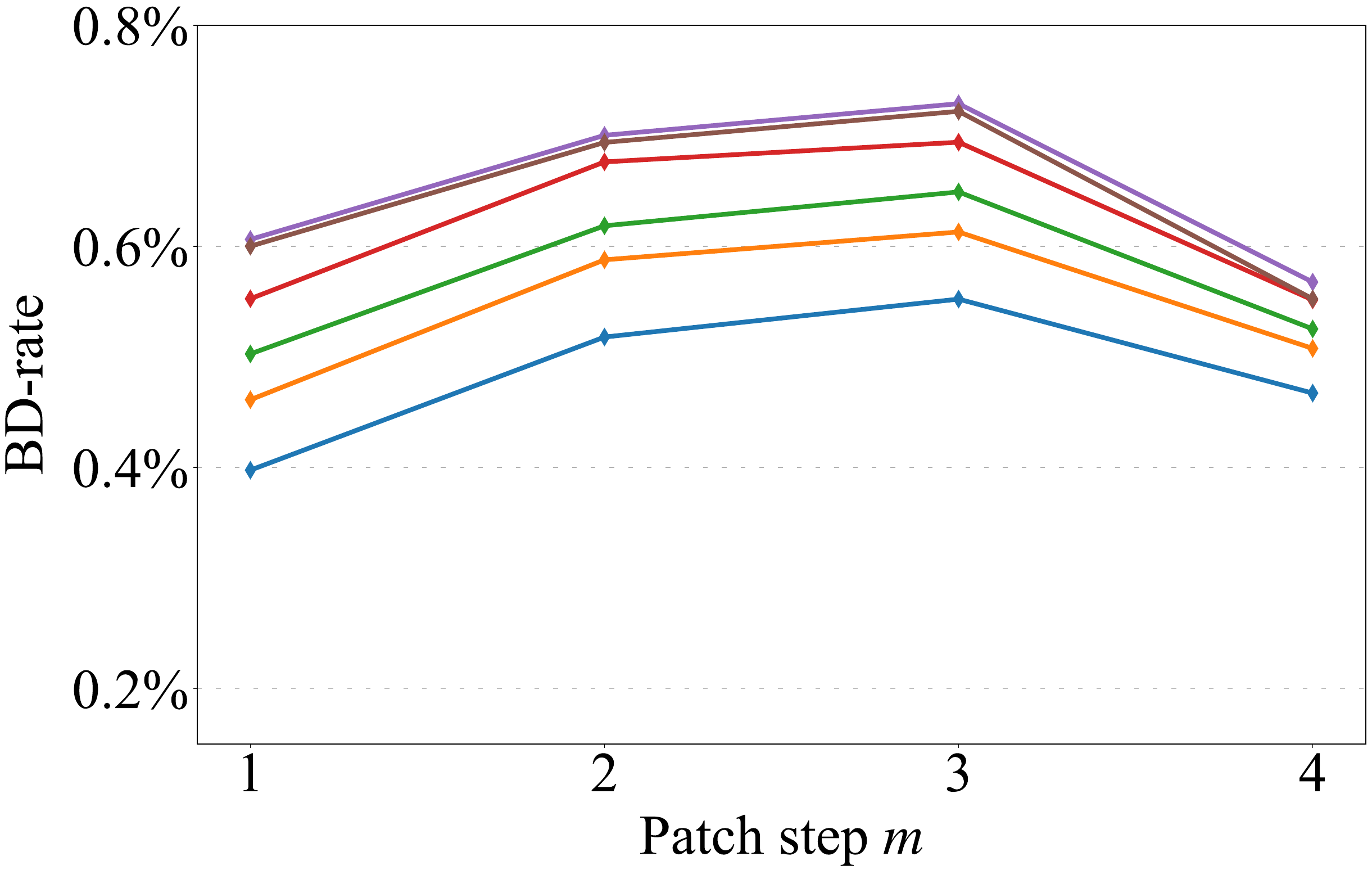}
			}
			\subfigure[]{
				\includegraphics[width=2.2in]{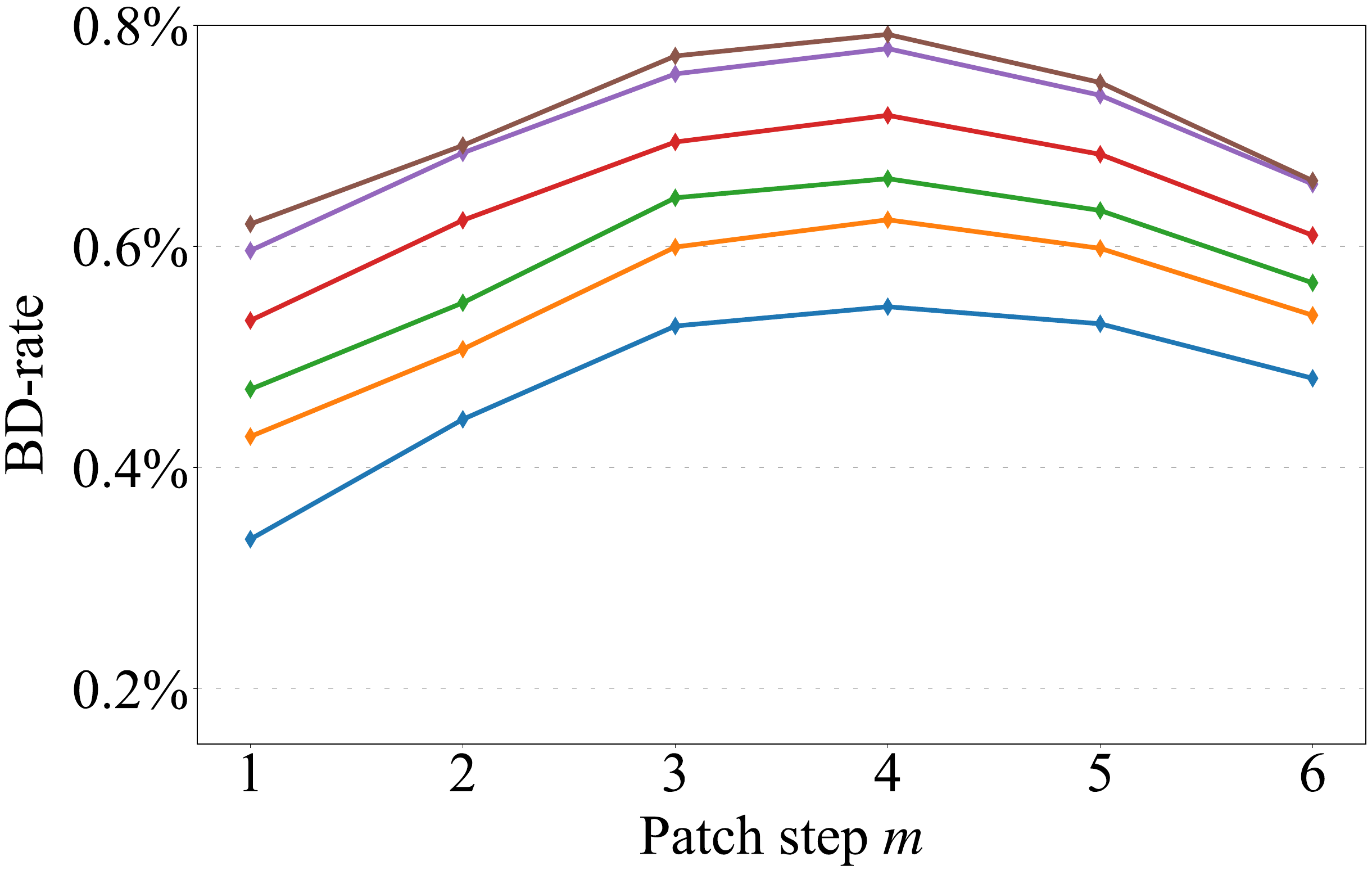}
			}
			\subfigure[]{
				\includegraphics[width=2.2in]{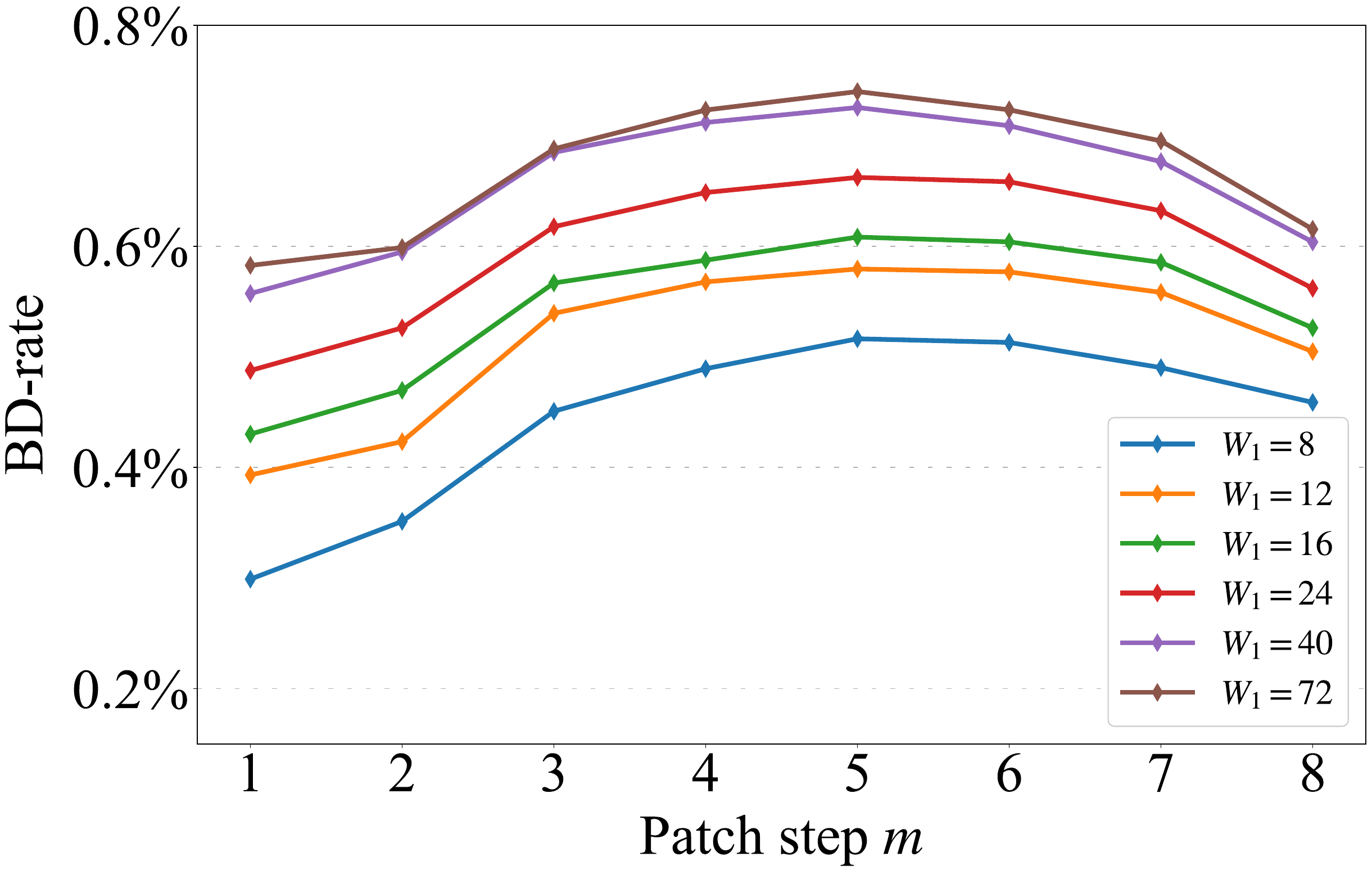}
			}
			\subfigure[]{
				\includegraphics[width=2.2in]{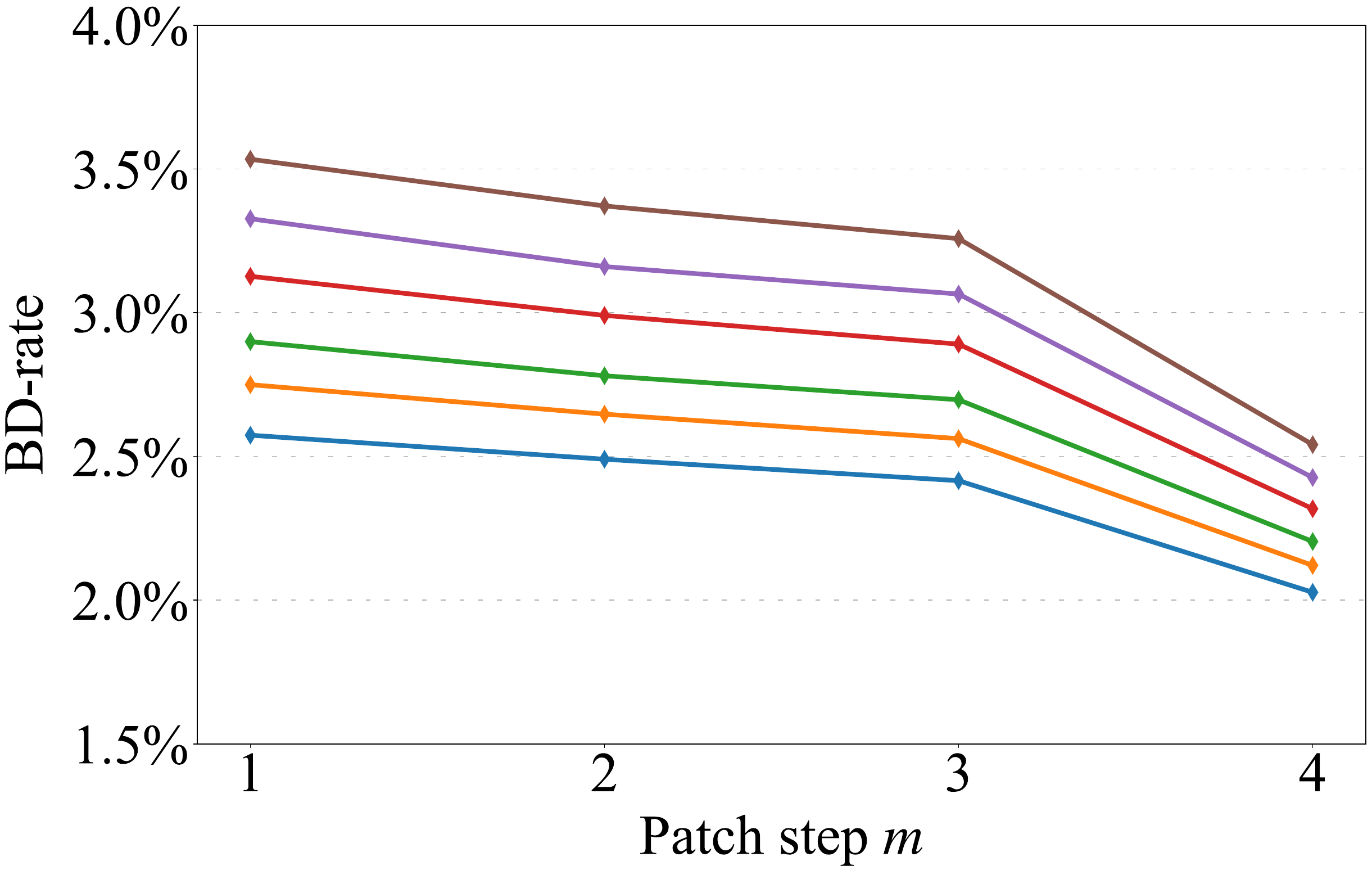}
			}
			\subfigure[]{
				\includegraphics[width=2.2in]{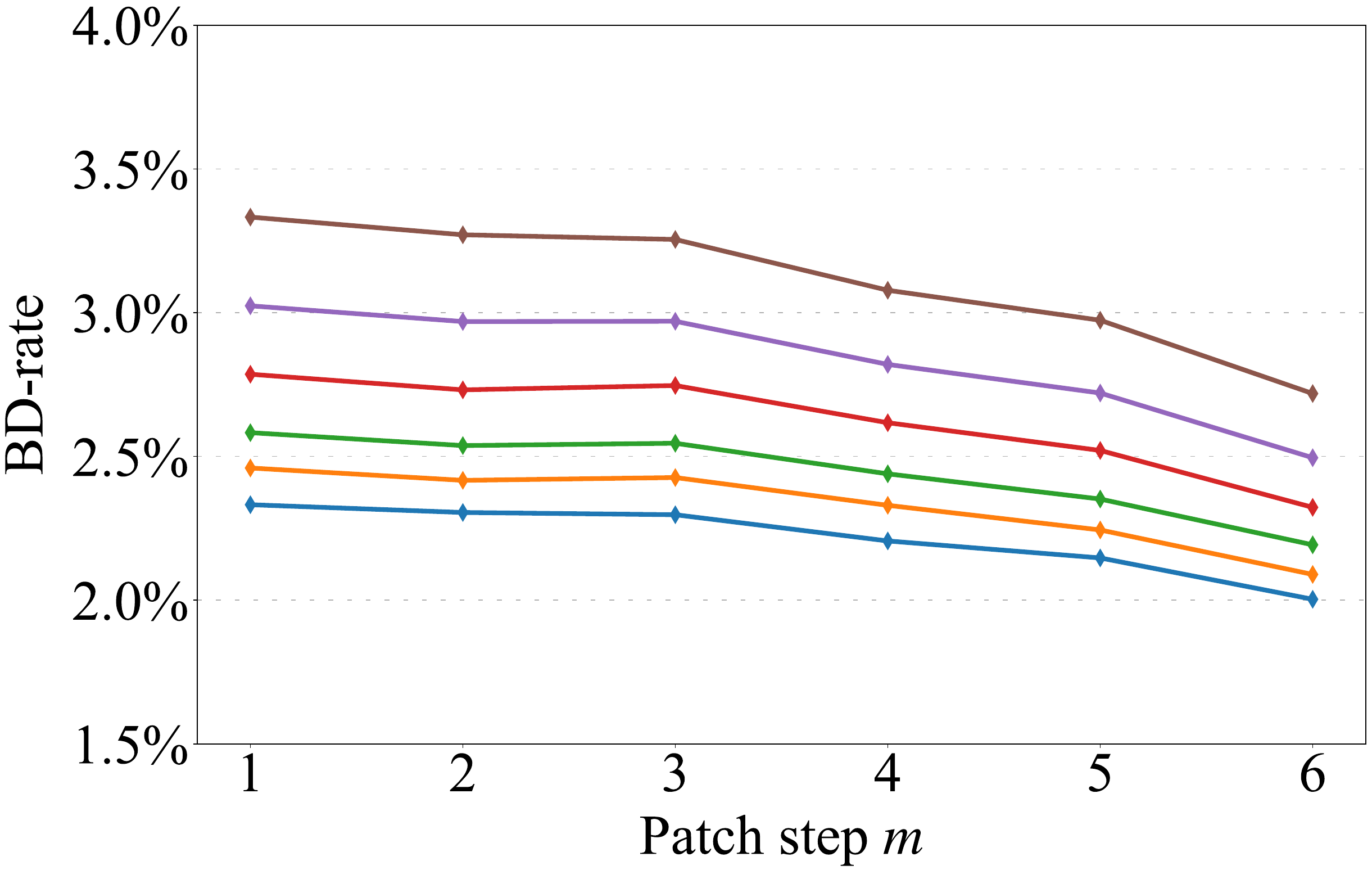}
			}
			\subfigure[]{
				\includegraphics[width=2.2in]{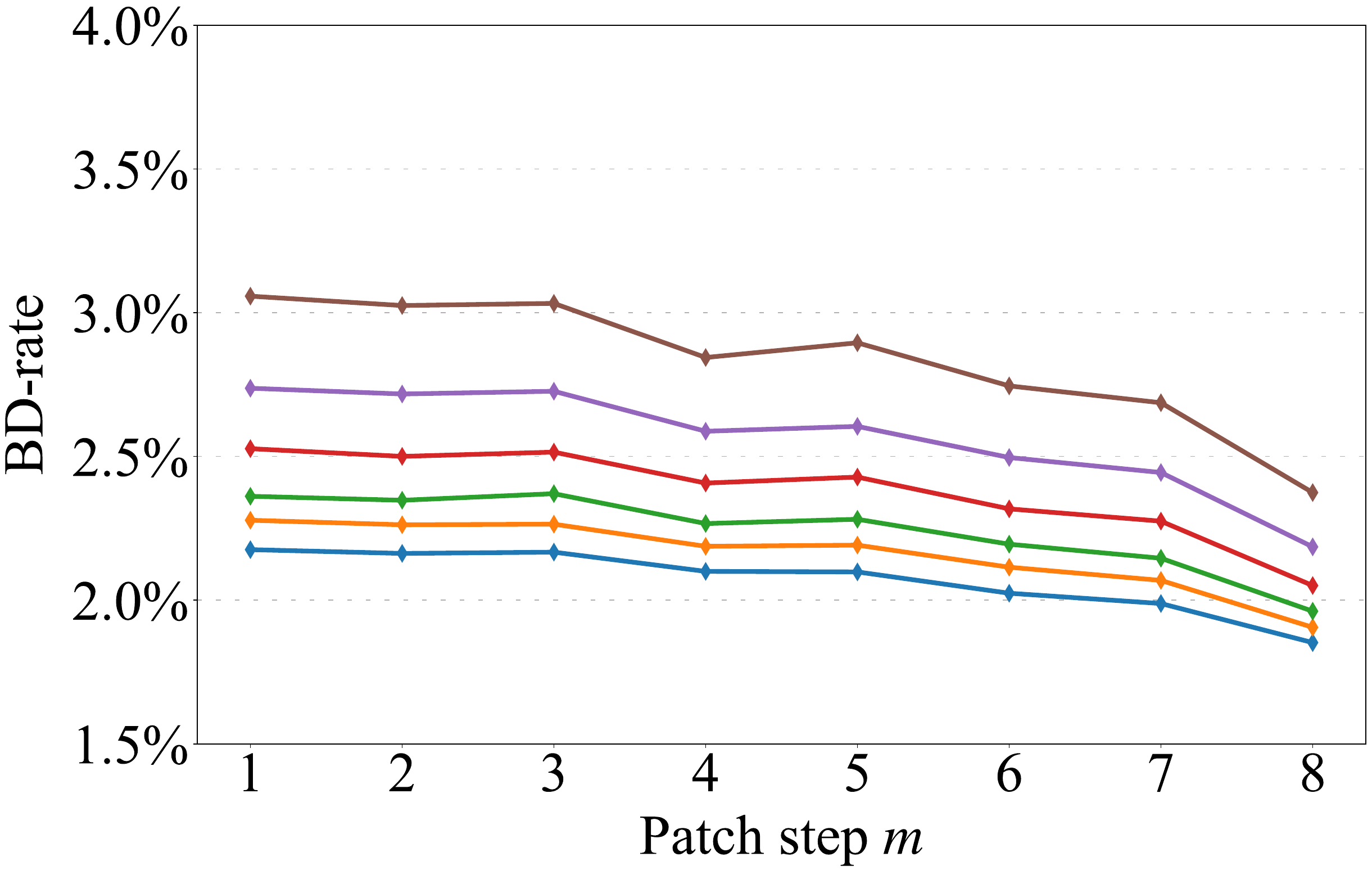}
			}
			\caption{Illustration of the influence of non-local reference sample derivation scheme on BD-rate under AI configuration. (a) $B_{s}=4$, natural sequence; (b) $B_{s}=6$, natural sequence; (c) $B_{s}=8$, natural sequence; (d) $B_{s}=4$, screen content sequence; (e) $B_{s}=4$, screen content sequence; (f) $B_{s}=8$, screen content sequence. The y-axis is the absolute value of BD-rate, for better visualization.}
			\label{fig::bdRateSearch}
		\end{center}
		\vspace{-3mm}
	\end{figure*}
	
	\begin{figure*}[!t]
		\begin{center}
			\noindent
			\subfigure[]{
				\includegraphics[width=2.2in]{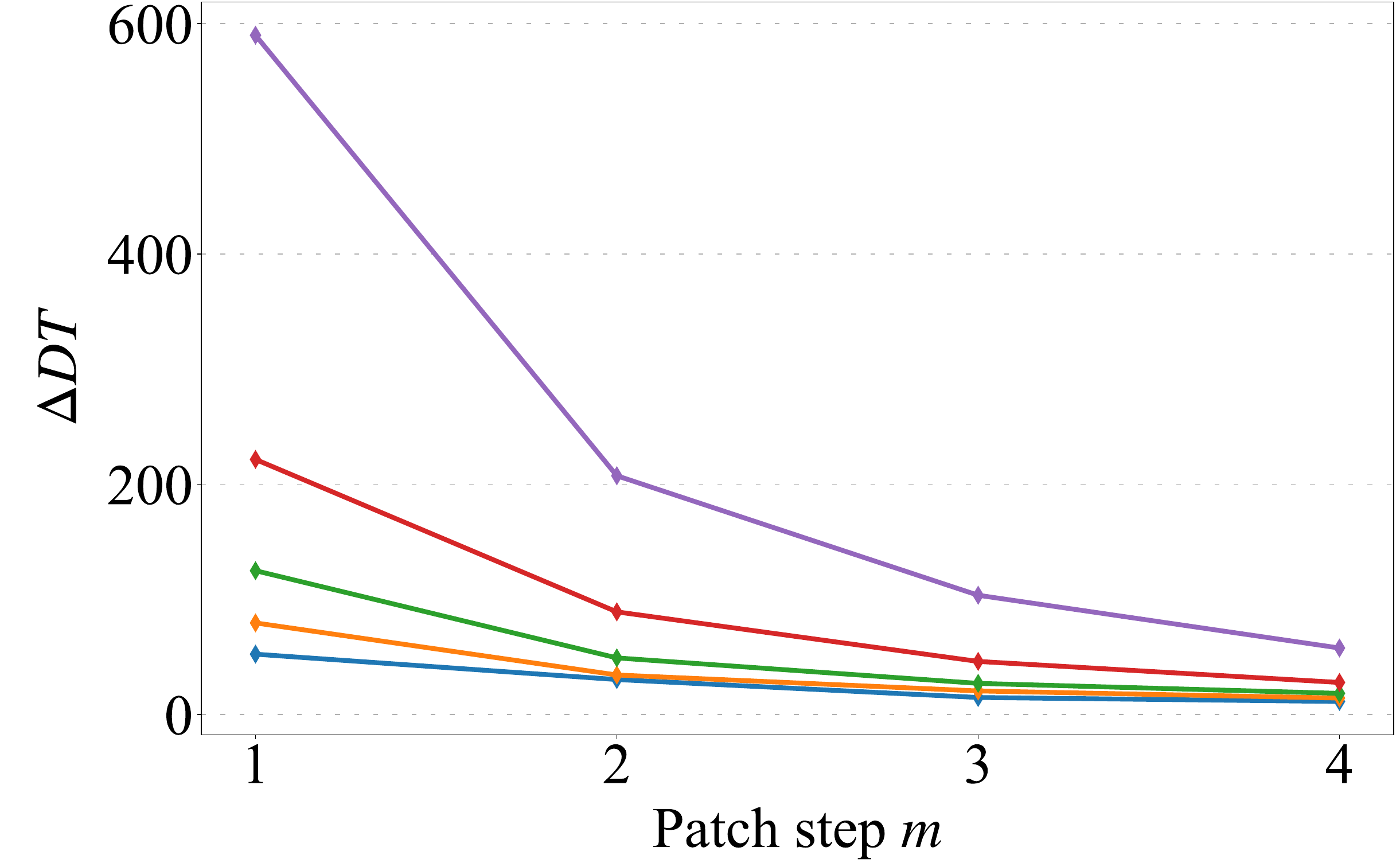}
			}
			\subfigure[]{
				\includegraphics[width=2.2in]{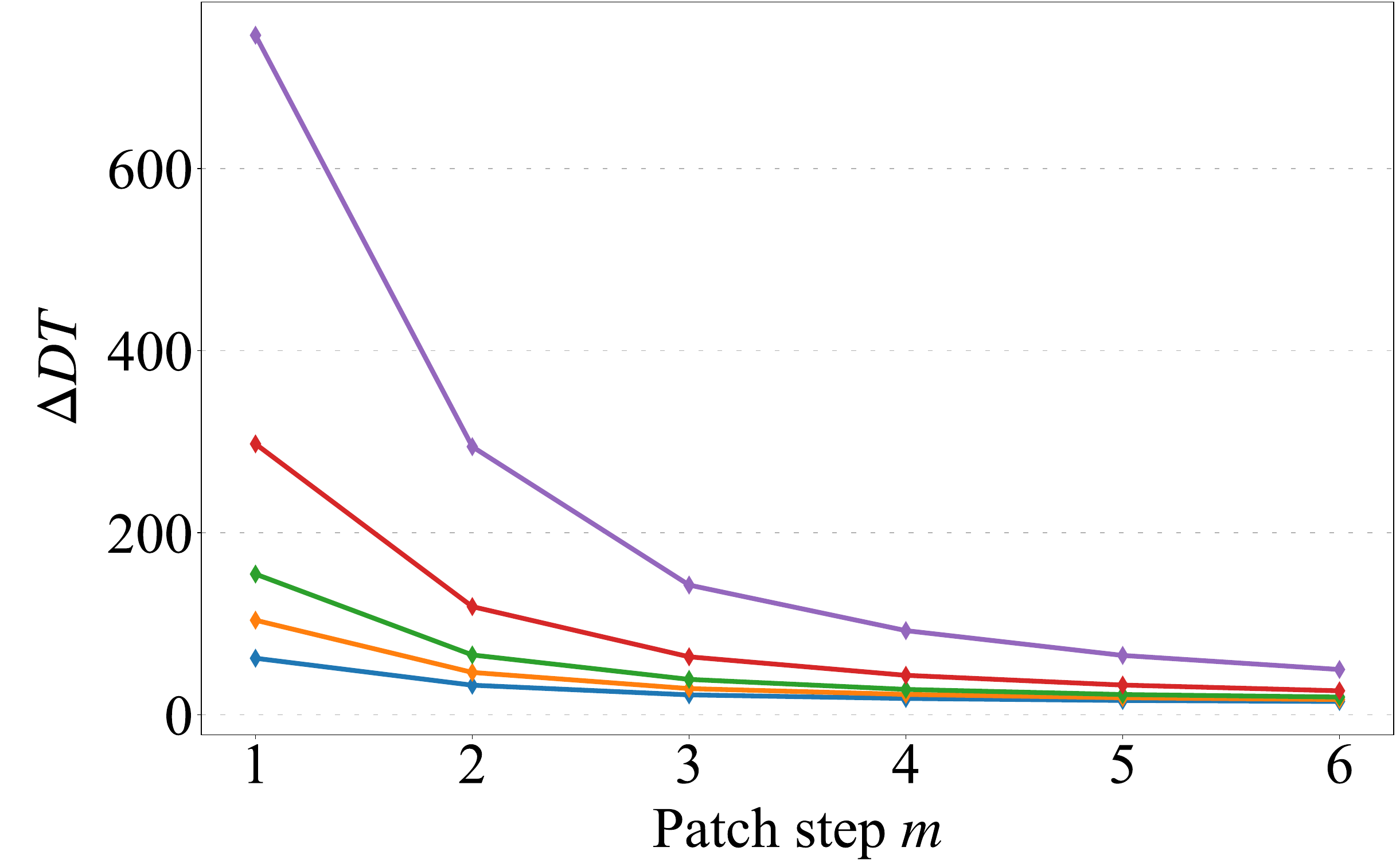}
			}
			\subfigure[]{
				\includegraphics[width=2.2in]{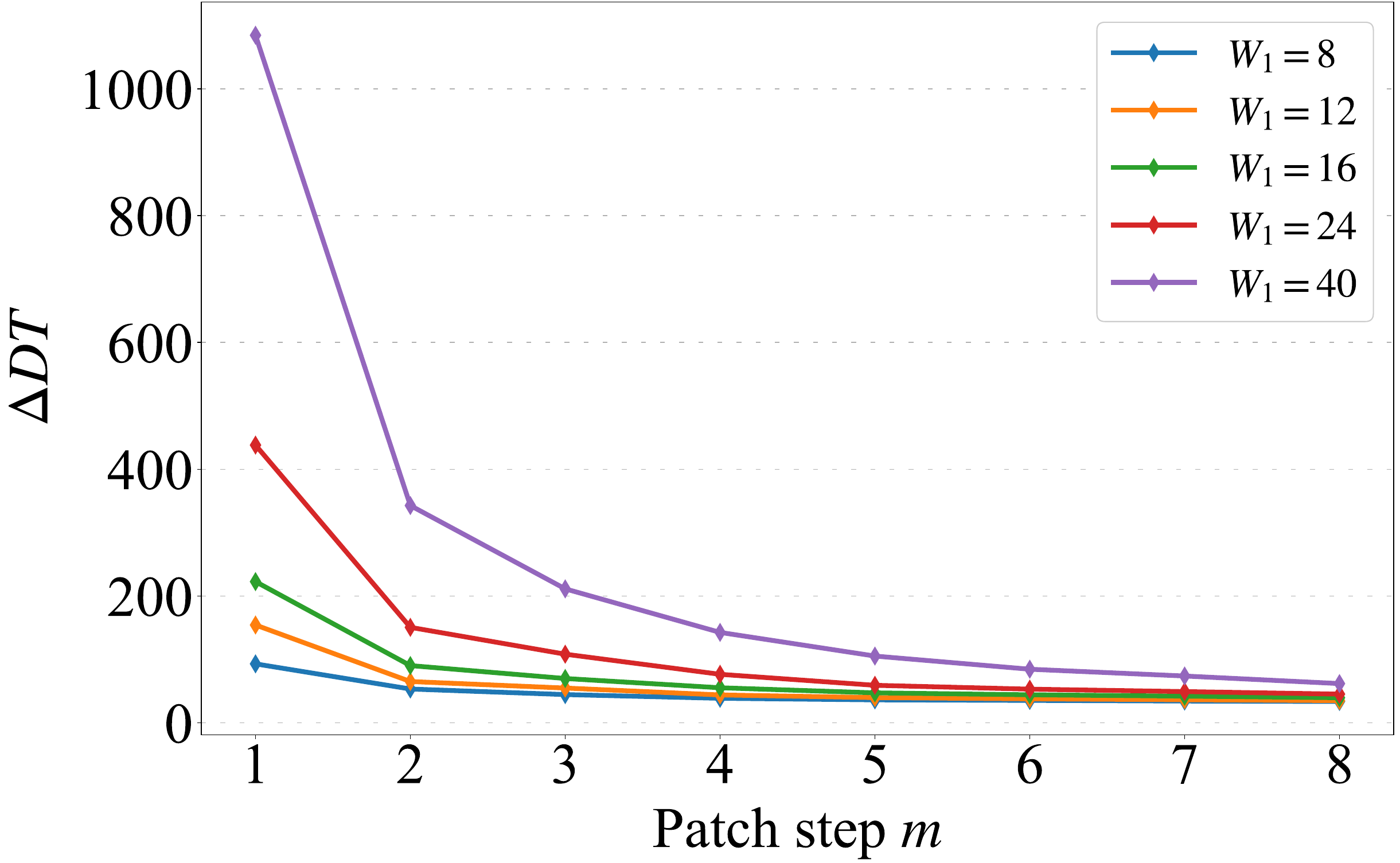}
			}
			\caption{Illustration of the influence of non-local reference sample derivation scheme on decoder complexity $\Delta DT$ under AI configuration. (a) $B_{s}=4$; (b) $B_{s}=6$; (c) $B_{s}=8$.}
			\label{fig::decTSearch}
		\end{center}
		\vspace{-3mm}
	\end{figure*}

	\subsection{Adaptive Filtering Operation}
	With the reference sample vector $\textbf{r}$, the classification results $\mathcal{C}$, and the decoded filter parameters $\textbf{w}$, the spatial filtering operation of DWF can be formulated by Eqn.(\ref{filtering}). However, the intensity of reference samples may be too different from the current sample in some cases. Subsequently, the filter coefficients may be unsuitable for other reference samples, resulting in poor performance. To address this issue, we propose to modify reference samples by limiting the intensity of outlier samples to a specific range. With the constrained reference samples, the filtering operation is formulated by,
	\begin{equation}
	   f(\textbf{p}, t) = z(\textbf{p}, t) + \sum_{j=0}^{N-1}w_{j}\times{\mathcal{T} (r_{j} - z(\textbf{p}, t))},
	\end{equation}
	\noindent where $\mathcal{T}(\cdot)$ is the constraint function, which is formulated by,
	
	\begin{equation}
	    \mathcal{T} (r_{j} - z(\textbf{p}, t)) = 
	    \begin{cases}
	    \xi_{j} & r_{j} - z(\textbf{p}, t) > \xi_{j},\\
	    -\xi_{j} & r_{j} - z(\textbf{p}, t) < -\xi_{j},\\
	    r_{j} - z(\textbf{p}, t) & \mathrm{otherwise.}
	    \end{cases}
	\end{equation}
	\noindent where $\xi_{j}$ is the constraint range. There are four candidates of $\xi$, i.e., 8, 32, 256, 1023, which can be selected by the encoder for each filter. The indexes of these selected ranges are transmitted to the decoder. The filter tap $N$ is set to 25. For a better trade-off between filter coefficient signaling and distortion reduction capability, we utilize a coefficient sharing scheme. In particular, the two neighboring reference samples share the same coefficients. By doing so, the number of filter coefficients can be reduced from $N$ to $N/2$. And the filtering operation can be formulated by,
    \begin{equation}
    \begin{split}
	   f(\textbf{p}, t) = z(\textbf{p}, t) & + \sum_{j=0}^{N/2-1}w_{j}\times{\mathcal{T} (r_{2j} - z(\textbf{p}, t))} \\ 
            & + \sum_{j=0}^{N/2-1}w_{j}\times{\mathcal{T} (r_{2j + 1} - z(\textbf{p}, t))}.
    \end{split}
    \end{equation}

 \begin{table}[t!]
		\begin{center}
			\caption{Syntax structure of the proposed DWF.} 
			\label{Syntax_pro}
			\begin{tabular}{l|c}
				\thickhline
				\textbf{Syntax}        & \textbf{Descriptor} \\
				\hline
				dwf\_parameter\_set~() \{\hspace{2.0cm}                       &       \\
				\hspace{0.3cm}\textbf{dwf\_luma\_new\_flag}           & u(1) \\
				\hspace{0.3cm}\textbf{dwf\_chroma\_new\_flag}         & u(1) \\
				\hspace{0.3cm}if(dwf\_luma\_new\_flag)\{                &       \\
				\hspace{0.6cm}\textbf{dwf\_num\_luma\_filters\_minus1}           & ue(v) \\
                    \hspace{0.6cm}if(dwf\_num\_luma\_filters\_minus1 $>$ 0)                &       \\
                    \hspace{0.9cm}for(i = 0; i $< K$; i++)                &       \\
                    \hspace{1.2cm}\textbf{dwf\_luma\_filter\_idx}[i]             & u(v)      \\
                    \hspace{0.6cm}for(i = 0; i $\leq$ dwf\_num\_luma\_filters\_minus1; i++)             &    \\
                    \hspace{0.9cm}for(j = 0; j $< N/2$; j++)\{             &    \\
                    \hspace{1.2cm}\textbf{dwf\_luma\_coeff\_abs}[i][j]             & ue(v)      \\
                    \hspace{1.5cm}if(dwf\_luma\_coeff\_abs[i][j] $\neq$ 0)                &       \\
                    \hspace{1.8cm}\textbf{dwf\_luma\_coeff\_sign}[i][j]             & u(1)      \\
                    \hspace{0.9cm}\}             &    \\

                    \hspace{0.6cm}\textbf{dwf\_luma\_constrain\_flag}           & u(1) \\
                    \hspace{0.6cm}if(dwf\_luma\_constrain\_flag $\neq$ 0)                &       \\
                    \hspace{0.9cm} for(i = 0; i $\leq$ dwf\_num\_luma\_filters\_minus1; i++) & \\
                    \hspace{1.2cm} for(j = 0; j $< N/2$; j++) & \\
                    \hspace{1.5cm} \textbf{dwf\_luma\_constrain\_idx}[i][j]           & u(2) \\
				\hspace{0.3cm}\}                                 &       \\

                    \hspace{0.3cm}                                 &       \\
                    
                    \hspace{0.3cm}if(dwf\_chroma\_new\_flag)\{                &       \\
                    \hspace{0.6cm}for(j = 0; j $< N/4$; j++)\{             &    \\
                    \hspace{0.9cm}\textbf{dwf\_chroma\_coeff\_abs}[j]             & ue(v)      \\
                    \hspace{1.2cm}if(dwf\_chroma\_coeff\_abs[j] $\neq$ 0)                &       \\
                    \hspace{1.5cm}\textbf{dwf\_chroma\_coeff\_sign}[j]             & u(1)      \\
                    \hspace{0.6cm}\}             &    \\

                    \hspace{0.6cm}\textbf{dwf\_chroma\_constrain\_flag}           & u(1) \\
                    \hspace{0.6cm}if(dwf\_chroma\_constrain\_flag $\neq$ 0)                &       \\
                    \hspace{0.9cm} for(j = 0; j $< N/4$; j++) & \\
                    \hspace{1.2cm} \textbf{dwf\_chroma\_constrain\_idx}[j]           & u(2) \\

                    \hspace{0.3cm}                                 &       \\
                    
                    \hspace{0.3cm}\}                                 &       \\
				\}                                               &       \\
				\thickhline 
			\end{tabular}
		\end{center}
		\vspace{-3mm}
	\end{table}
 
	\subsection{Syntax Design}
        For a frame in which DWF is conducted, the filter parameters can be signaled in the bitstream or derived from filter parameters of reference frames. The syntax design of the proposed DWF is illustrated in Table \ref{Syntax_pro}. If the frame-level DWF is on, the new filter flags of luma and chroma components are signaled by \textit{dwf\_luma\_new\_flag} and \textit{dwf\_chroma\_new\_flag}, respectively.  If the \textit{dwf\_luma\_new\_flag} is 1, filter parameters of luma component are signaled, where \textit{dwf\_num\_luma\_filters\_minus1}, \textit{dwf\_luma\_filter\_idx[i]}, \textit{dwf\_luma\_coeff\_abs[i][j]}, \textit{dwf\_luma\_coeff\_sign[i][j]}, \textit{dwf\_luma\_constrain\_flag}, and \textit{dwf\_luma\_constrain\_idx[i][j]} represent the number of DWF luma filters minus 1, the corresponding filter index of each luma class, the absolute value of DWF coefficients, the sign of DWF coefficients, the on/off flag of outlier data constraints, and the index of constraint parameters. $K$ indicates the number of classes defined in Section \ref{classification}. $N$ is the number of filter taps defined in Section \ref{formulation}. The parameters of chroma are similar to that of the luma component. The main difference is that there is only one filter designed for Cb and Cr components. In Table \ref{Syntax_pro}, u(1), u(2), u(v) and ue(v) represent 1 bit fixed-length coding, 2 bits fixed-length coding, variable-length coding, and unsigned integer 0-th order Exp-Golomb-coded syntax, respectively. Besides the frame-level parameters, there are CTU-level~(Coding Tree Unit) on/off flags to control DWF.

	\begin{table}[t!]
		\centering
		\begin{center}
			\caption{Illustration of the parameter settings for the investigation of non-local reference sample derivation scheme.} \label{table::Test_condition1}
			\begin{tabular}{c c}
				\thickhline
				\hline
				{\textbf{Parameter}}     &{\textbf{Settings}}\\
				\hline
				$W_1$                    & \{16, 24, 32, 48, 80, 144\}  \\
				$W_2$                    & \{0, 8, 16, 32\}           \\
				$N_{ref}$                & \{2, 4\}            \\
				$B_{s}$                  & \{4, 6, 8\}                  \\
				$m$                      & \{1, 2, ..., $B_{s}$\}       \\
				\thickhline
			\end{tabular}
		\end{center}
		\vspace{-3mm}
	\end{table}
	
		\begin{figure}[!t]
		\begin{center}
			\noindent
			\subfigure[]{
				\includegraphics[width=2.8in]{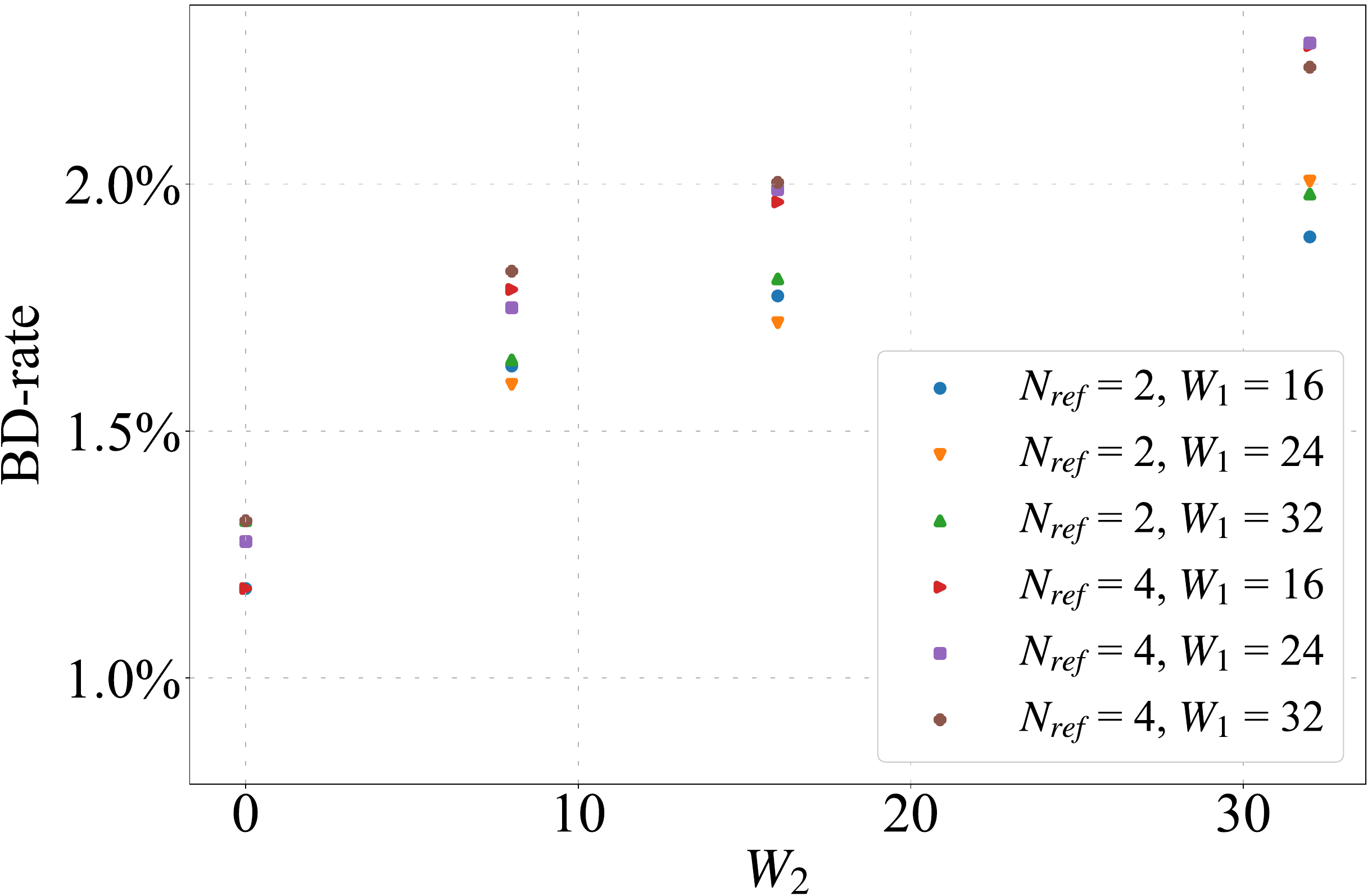}
			}
			\subfigure[]{
				\includegraphics[width=2.8in]{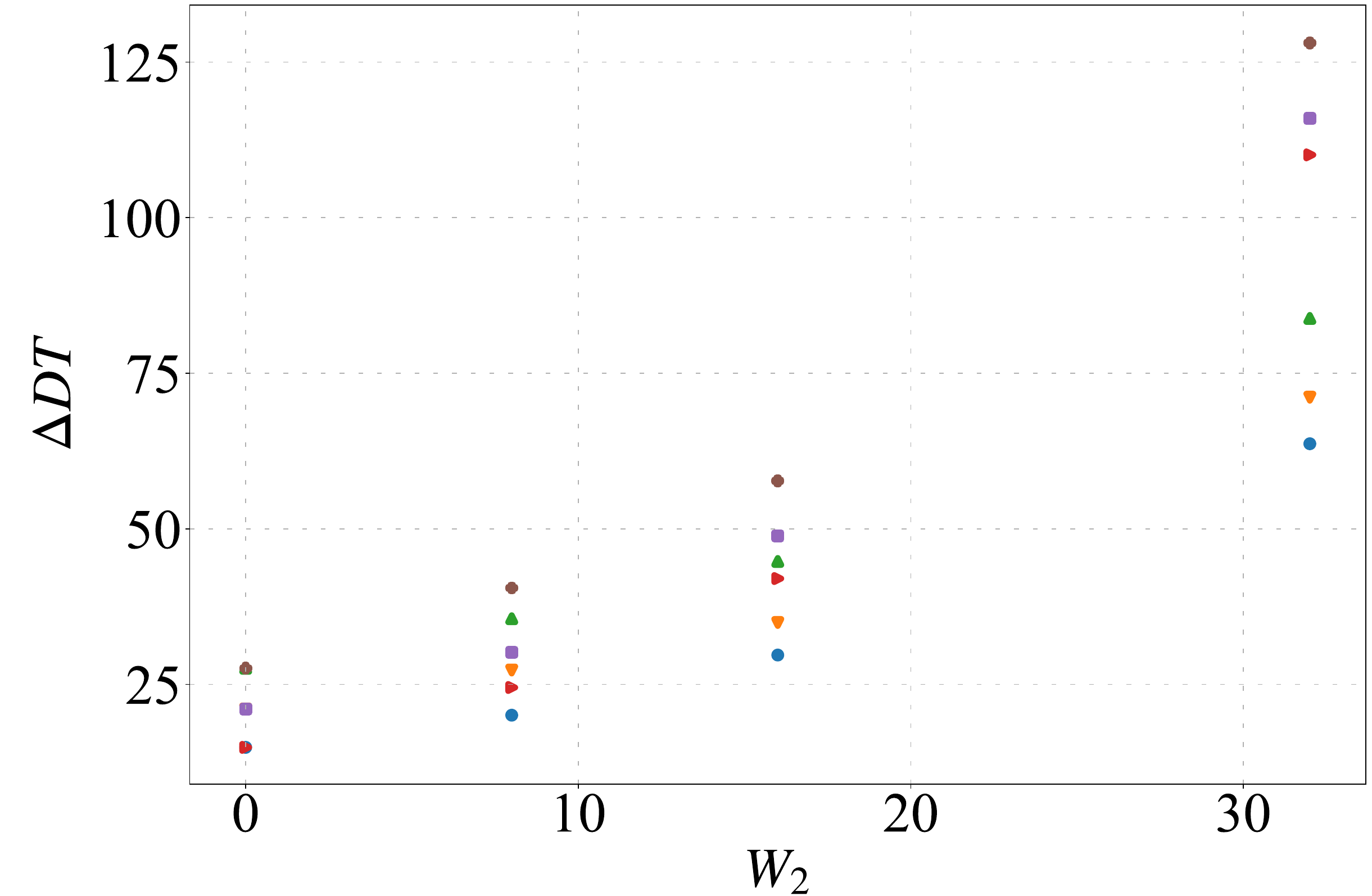}
			}
			\caption{Illustration of the influence of non-local reference sample derivation scheme on BD-rate and decoder complexity $\Delta DT$ under RA configuration. The y-axis of (a) is the absolute value of BD-rate, for better visualization.}
			\label{fig::refSearch}
		\end{center}
		\vspace{-3mm}
	\end{figure}

	\subsection{Implementation Study and Parameter Selection}
	For a better trade-off between coding gain and computational complexity, we investigate the influence of the block matching scheme. There are five factors: the search window $W_{1}\times W_{1}$ of the current frame and search window $W_{2}\times W_{2}$ of reference frames, the number of reference frames $N_{ref}$, the block size $B_{s}\times B_{s}$ and the block step $m$. The experimental settings are shown in Table \ref{table::Test_condition1}. The natural sequences used for the experiment are \textit{BQTerrace}, \textit{MarketPlace}, and \textit{Cactus}. The screen content sequences used for the experiment are \textit{Console}, \textit{Desktop}, and \textit{FlyingGraphics}. The first one second is encoded at QP \{22,27,32,37\}. The coding performance is measured by Bjontegaard's method~\cite{BDrate} in terms of BD-rate, and the negative BD-rate indicates the performance gain. For better visualization, we take the absolute value of BD-rate as the y-axis of Fig.~\ref{fig::bdRateSearch} and Fig.~\ref{fig::refSearch} (a). The decoding complexity is measured by,
	\begin{equation} \label{decT}
	\Delta DT = \frac{TDec_{\mathrm{test}}}{TDec_{\mathrm{anchor}}} \times 100\%,
	\end{equation}
	where $TDec_{\mathrm{test}}$ and $TDec_{\mathrm{anchor}}$ represent the consumed decoding time with the tested scheme and the anchor.
	
	We first explore the influence of $W_{1}$, $B_{s}$ and $m$ under AI configuration, which is depicted in Fig.~\ref{fig::bdRateSearch} and Fig.~\ref{fig::decTSearch}. An interesting phenomenon lies in that the coding performance first increases and subsequently declines with the decrease of $m$ for various block sizes $B_{s}$ of natural sequences. However, the coding performance exhibits an increasing trend with slight fluctuations for screen content sequences. The main reason lies in that as $m$ decreases, the extent of overlapping rises. The number of times that each point is filtered increases. For natural sequences with relatively rich textures, excessive overlapping may cause the over-smoothing issue. While for screen content sequences, the non-local similarity is relatively stronger~\cite{li2020unified}, so combining more filtered results can result in better coding performance. As a more extensive search range can improve the similarity between reference blocks and the current block, it is no surprise that the BD-rate shows a stable upward trend for different block sizes and block steps with the increment of $W_{1}$. For natural sequences, the increasing rate of BD-rate shows a downward trend as $W_{1}$ increases. While for screen content videos, this value is relatively stable. The fundamental reason is that with the increase of search range is more feasible for blocks in screen content videos to seek more similar blocks due to the strong non-local similarity. As for the computational complexity, $\Delta DT$ increases with the increase of search range $W_{1}$ and the extension of overlapping. For a better trade-off between BD-rate and computational complexity, we set the parameters as $B_{s}=6, m= 4$ for natural sequence, $B_{s} = 4, m = 3$ for screen content videos. For the coding of I-frames, $W_{1}$ is set to 32 for both natural and screen content sequences.

	With the selected $B_{s}$, $m$, and $W_{1}$ for I-frames, we further investigate the influence of $W_{1}$, $W_{2}$, and $N_{ref}$ under RA configuration, which is shown in Fig.~\ref{fig::refSearch}. It is observed that with the increment of the searching area, $\Delta DT$ shows a stable upward trend. However, BD-rate does not always increase with the increment of the search range. For block matching schemes with similar complexity, searching schemes with a smaller search range on the current frame, more reference frames, and a larger search range on reference frames show better BD-rate performance. For parameter settings with similar BD-rate, searching schemes with these features also show lower complexity. In view of this, we use four reference frames for RA configuration and eight reference frames for LDB configuration based on the hierarchical reference frame structure defined by VVC. The search range $W_{1}$ is set to be 24 for B-frames and P-frames. The search range $W_{2}$ on the reference frames is set to be 16. The parameter settings are detailed in Table \ref{table::paramSet}.
	
	\begin{table}[t!]
		\centering
		\begin{center}
			\caption{Illustration of the parameter settings in non-local reference sample derivation process.} \label{table::paramSet}
			\begin{tabular}{c c c c}
				\thickhline
				\hline
				                       &I-frame     & B-frame (RA)    & B-frame (LDB)\\
			    \hline
				$B_{s}$                & 6, 4       & 6, 4            & 6, 4 \\
				$m$                    & 4, 3       & 4, 3            & 4, 3 \\
				$W_{1}$                & 32         & 24              & 24 \\
				$W_{2}$                & -          & 16              & 16 \\
				$N_{ref}$              & -          & 4               & 8 \\
				\thickhline
			\end{tabular}
		\end{center}
		\vspace{-3mm}
	\end{table}

	\begin{table*}[ht!]
		\centering
		\begin{center}
			\caption{Experimental results of the proposed DWF~(Y component), Anchor: VTM-11.0.} 
			\label{table::Performance1}
			\setlength{\tabcolsep}{3.10mm}{
				\begin{tabular}{c|c|c c c|c c c|c c c}
					\thickhline
					\hline
					\multirow{2}{*}{\textbf{Class}}&
					\multirow{2}{*}{\textbf{Sequence}}&
					\multicolumn{3}{c|}{\rule{0pt}{8pt} \textbf{AI}} & \multicolumn{3}{c|}{\textbf{RA}}& \multicolumn{3}{c}{\textbf{LDB}}\\
					\cline{3-11} & & 
					\textbf{BD-Rate}& $\Delta \textit{\textbf{ET}}$ & $\Delta \textit{\textbf{DT}}$ & \textbf{BD-Rate}& $\Delta \textit{\textbf{ET}}$ & $\Delta \textit{\textbf{DT}}$ & \textbf{BD-Rate} & $\Delta \textit{\textbf{ET}}$ & $\Delta \textit{\textbf{DT}}$ \\
					
					\hline
					\multirow{3}{*}{A1}
					& \rule{0pt}{8pt} \textit{Tango2}               & -0.29\% & 130\%  &  765\%   & -0.38\% & 114\%  &  469\%    & -       & - & -        \\
					& \rule{0pt}{8pt} \textit{FoodMarket4}          & -0.56\% & 152\%  & 1022\%   & -0.42\% & 119\%  &  652\%    & -       & - & -        \\
					& \rule{0pt}{8pt} \textit{Campfire}             & -0.56\% & 116\%  & 1254\%   & -1.36\% & 107\%  & 1396\%    & -       & - & -        \\
					
					\hline
					\multirow{3}{*}{A2}
					& \rule{0pt}{8pt} \textit{CatRobot1}            & -0.62\% & 118\%  & 1512\%   & -0.90\% & 114\%  &  1056\%    & -       & - & -        \\
					& \rule{0pt}{8pt} \textit{DaylightRoad2}        & -0.21\% & 116\%  & 1611\%   & -0.57\% & 113\%  &   412\%    & -       & - & -        \\
					& \rule{0pt}{8pt} \textit{ParkRunning3}         & -1.03\% & 113\%  & 1375\%   & -1.30\% & 107\%  &  1120\%    & -       & - & -        \\
					
					\hline
					\multirow{6}{*}{B}
					& \rule{0pt}{8pt} \textit{MarketPlace}          & -0.31\% & 116\%  & 1286\%   & -1.21\% & 113\%  & 2269\%    & -1.43\% & 113\%  & 1264\%  \\
					& \rule{0pt}{8pt} \textit{RitualDance}          & -0.51\% & 123\%  &  968\%   & -0.57\% & 111\%  & 3654\%    & -0.78\% & 112\%  & 1298\%  \\
					& \rule{0pt}{8pt} \textit{Cactus}               & -0.50\% & 113\%  & 1269\%   & -1.02\% & 111\%  & 4210\%    & -2.24\% & 111\%  & 2700\%  \\
					& \rule{0pt}{8pt} \textit{BasketballDrive}      & -0.30\% & 117\%  &  585\%   & -0.65\% & 109\%  & 2658\%    & -1.15\% & 109\%  & 1475\%  \\
					& \rule{0pt}{8pt} \textit{BQTerrace}            & -1.08\% & 114\%  & 1010\%   & -3.08\% & 116\%  & 3897\%    & -4.06\% & 113\%  & 3745\%  \\
					
					\hline
					\multirow{5}{*}{C}
					& \rule{0pt}{8pt} \textit{BasketballDrill}      & -0.69\% & 113\%  & 1971\%   & -1.26\% & 109\%  & 1397\%    & -1.57\% & 108\%  & 3345\%  \\
					& \rule{0pt}{8pt} \textit{BQMall}               & -0.75\% & 112\%  & 1614\%   & -1.62\% & 110\%  & 1245\%    & -2.00\% & 109\%  & 4024\%  \\
					& \rule{0pt}{8pt} \textit{PartyScene}           & -0.13\% & 113\%  &  535\%   & -1.32\% & 110\%  & 2758\%    & -1.65\% & 111\%  & 3238\%  \\
					& \rule{0pt}{8pt} \textit{RaceHorsesC}          & -0.12\% & 111\%  &  500\%   & -0.77\% & 108\%  & 2055\%    & -0.75\% & 109\%  & 3145\%  \\
					 
					\hline
					\multirow{5}{*}{D}
					& \rule{0pt}{8pt} \textit{BasketballPass}       & -0.13\% & 112\%  &  696\%   & -0.93\% & 110\%  & 3065\%    & -1.36\% & 110\%  & 2936\%  \\
					& \rule{0pt}{8pt} \textit{BQSquare}             & -0.22\% & 115\%  &  924\%   & -4.32\% & 115\%  & 3241\%    & -4.82\% & 114\%  & 2349\%  \\
					& \rule{0pt}{8pt} \textit{BlowingBubbles}       & -0.09\% & 111\%  &  650\%   & -1.33\% & 110\%  & 3845\%    & -1.62\% & 110\%  & 2300\%  \\
					& \rule{0pt}{8pt} \textit{RaceHorses}           & -0.05\% & 112\%  &  512\%   & -0.69\% & 109\%  & 2874\%    & -1.00\% & 110\%  & 2239\%  \\
					
					\hline
					\multirow{3}{*}{E}
					& \rule{0pt}{8pt} \textit{FourPeople}           & -0.67\% & 115\%  & 2099\%   & -       & -       & -        & -1.86\% & 116\%  & 1129\%  \\
					& \rule{0pt}{8pt} \textit{Johnny}               & -1.29\% & 116\%  & 2554\%   & -       & -       & -        & -4.14\% & 119\%  & 2302\%  \\
					& \rule{0pt}{8pt} \textit{KristenAndSara}       & -0.83\% & 115\%  & 1312\%   & -       & -       & -        & -2.54\% & 118\%  & 1198\%  \\
					
					\hline
					\multirow{5}{*}{F}
					& \rule{0pt}{8pt} \textit{BasketballDrillText}  & -0.57\% & 107\%  & 1765\%   & -1.47\% & 100\%  & 2875\%    & -1.82\% & 108\%  & 965\%  \\
					& \rule{0pt}{8pt} \textit{ArenaOfValor}         & -0.46\% & 107\%  & 1045\%   & -1.20\% & 112\%  & 1956\%    & -1.45\% & 111\%  & 896\%  \\
					& \rule{0pt}{8pt} \textit{SlideEditing}         & -3.10\% & 107\%  & 2301\%   & -2.11\% & 125\%  & 1352\%    & -3.08\% & 119\%  & 1520\%   \\
					& \rule{0pt}{8pt} \textit{SlideShow}            & -1.16\% & 113\%  & 1425\%   & -1.06\% & 124\%  & 456\%     & -0.61\% & 118\%  & 485\%   \\
					
					\hline
					\multirow{5}{*}{G}
					& \rule{0pt}{8pt} \textit{ChineseEditing}       & -1.55\% & 107\%  & 2012\%   & -2.22\% & 121\%  & 2563\%    & -3.83\% & 118\%  & 1798\%  \\
					& \rule{0pt}{8pt} \textit{Console}              & -3.10\% & 110\%  & 2156\%   & -3.07\% & 112\%  & 4862\%    & -4.05\% & 113\%  & 3568\%  \\
					& \rule{0pt}{8pt} \textit{Desktop}              & -4.21\% & 108\%  & 2658\%   & -3.39\% & 120\%  & 3568\%    & -2.23\% & 119\%  & 3102\%  \\
					& \rule{0pt}{8pt} \textit{FlyingGraphics}       & -2.73\% & 108\%  & 2258\%   & -6.35\% & 108\%  & 5125\%    & -6.56\% & 110\%  & 4107\%  \\
					
					\hline
					\multirow{7}{*}{M}
					& \rule{0pt}{8pt} \textit{MissionControlClip2} & -1.77\% & 108\%  & 2234\%   & -2.15\% & 137\%  & 2108\%    & -3.80\% & 114\%  & 1889\%  \\
					& \rule{0pt}{8pt} \textit{MissionControlClip3} & -1.71\% & 108\%  & 2187\%   & -1.63\% & 125\%  & 1709\%    & -3.04\% & 123\%  & 1609\%  \\
					& \rule{0pt}{8pt} \textit{Map}                 & -2.37\% & 107\%  & 2598\%   & -2.96\% & 117\%  & 2545\%    & -4.41\% & 115\%  & 3309\%  \\
					& \rule{0pt}{8pt} \textit{Programming}         & -2.11\% & 108\%  & 2309\%   & -4.13\% & 114\%  & 3655\%    & -5.85\% & 113\%  & 4198\%  \\
					& \rule{0pt}{8pt} \textit{Robot}               & -0.26\% & 106\%  &  767\%   & -1.29\% & 114\%  & 1698\%    & -1.23\% & 112\%  & 1648\%  \\
					& \rule{0pt}{8pt} \textit{Web\_browsing}       & -5.79\% & 108\%  & 2334\%   & -6.69\% & 134\%  & 5108\%    & -5.28\% & 124\%  & 3868\%  \\
					
					\hline
					\multirow{10}{*}{Average}
					& \rule{0pt}{8pt} {Class A1}           & -0.47\% & 131\%  & 1014\%   & -0.72\% & 114\%  &  839\%    & -       & - & -                   \\
					& \rule{0pt}{8pt} {Class A2}           & -0.62\% & 116\%  & 1499\%   & -0.92\% & 111\%  &  867\%    & -       & - & -                   \\
					& \rule{0pt}{8pt} {Class B}            & -0.54\% & 116\%  & 1024\%   & -1.31\% & 112\%  & 3337\%    & -1.93\% & 112\%  & 2096\%         \\
					& \rule{0pt}{8pt} {Class C}            & -0.43\% & 112\%  & 1155\%   & -1.24\% & 109\%  & 1864\%    & -1.49\% & 109\%  & 3438\%         \\
					& \rule{0pt}{8pt} {Class D}            & -0.12\% & 113\%  & 696\%    & -1.82\% & 111\%  & 3256\%    & -2.20\% & 111\%  & 2456\%         \\
					& \rule{0pt}{8pt} {Class E}            & -0.93\% & 115\%  & 1988\%   & -       & -      & -         & -2.85\% & 118\%  & 1543\%         \\
					& \rule{0pt}{8pt} {Class F}            & -1.32\% & 108\%  & 1634\%   & -1.46\% & 117\%  & 1660\%    & -1.74\% & 114\%  & 967\%         \\
					& \rule{0pt}{8pt} {Class G}            & -2.90\% & 108\%  & 2271\%   & -3.76\% & 115\%  & 4030\%    & -4.17\% & 115\%  & 3144\%         \\
					& \rule{0pt}{8pt} {Class M}            & -2.33\% & 107\%  & 2072\%   & -3.14\% & 123\%  & 2804\%    & -3.94\% & 117\%  & 2754\%         \\
					\hline
					\multicolumn{2}{c|} {\rule{0pt}{8pt}Average (CTC)} & -0.58\% & 118\%  & 1291\%  & -1.09\% & 111\%  & 1950\% & -2.01\% & 112\%  & 2405\%  \\
					\hline
					\multicolumn{2}{c|} {\rule{0pt}{8pt}Average (All)} & -1.16\% & 113\%  & 1502\%  & -1.92\% & 115\%  & 2480\% & -2.67\% & 114\%  & 2388\%        \\
					
					\thickhline
			\end{tabular}}
		\end{center}
		\vspace{-3mm}
	\end{table*}
	
	\begin{figure*}[!t]
		\begin{center}
			\noindent
			\subfigure[]{
				\includegraphics[width=1.65in]{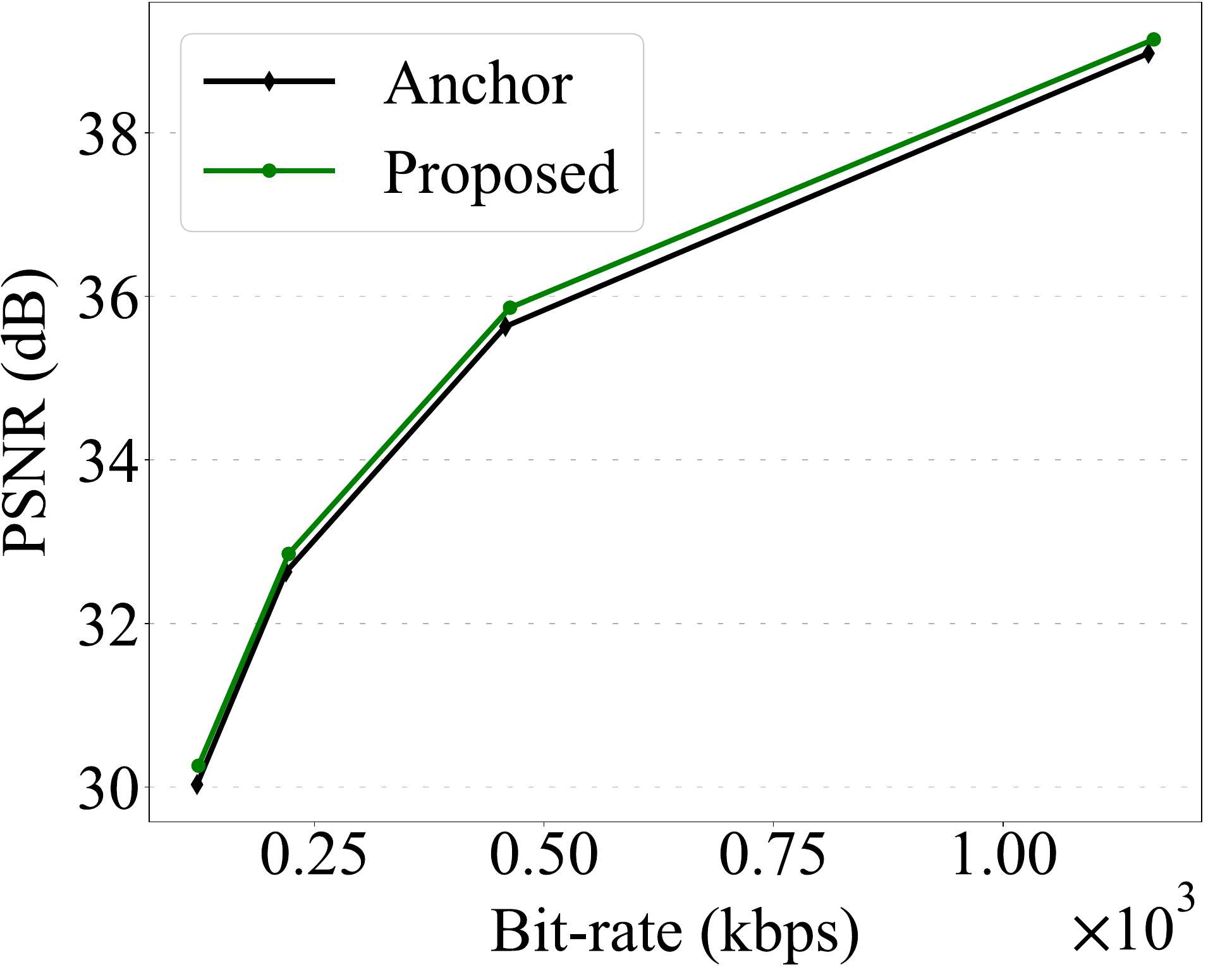}
			}
			\subfigure[]{
				\includegraphics[width=1.65in]{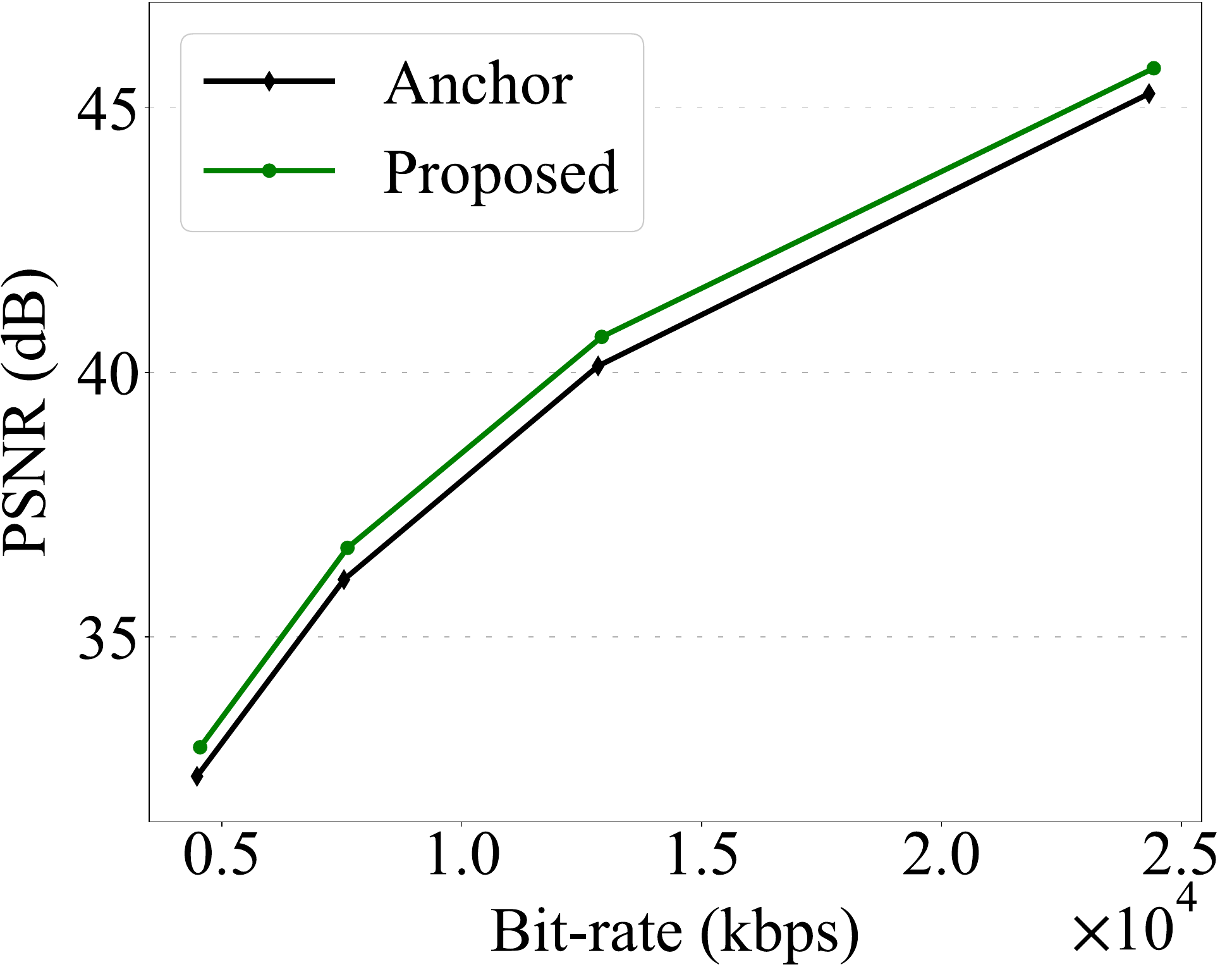}
			}
			\subfigure[]{
				\includegraphics[width=1.65in]{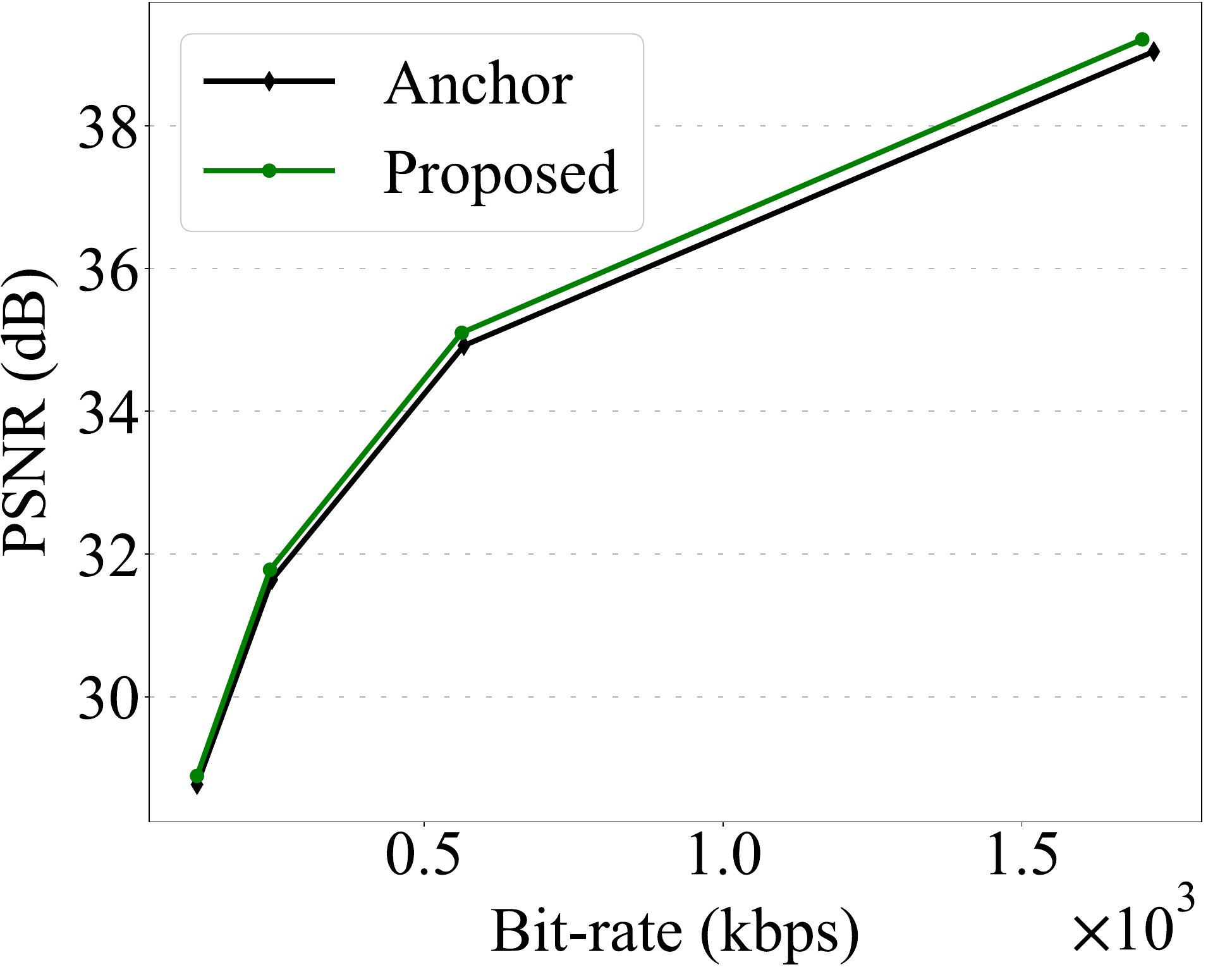}
			}
			\subfigure[]{
				\includegraphics[width=1.65in]{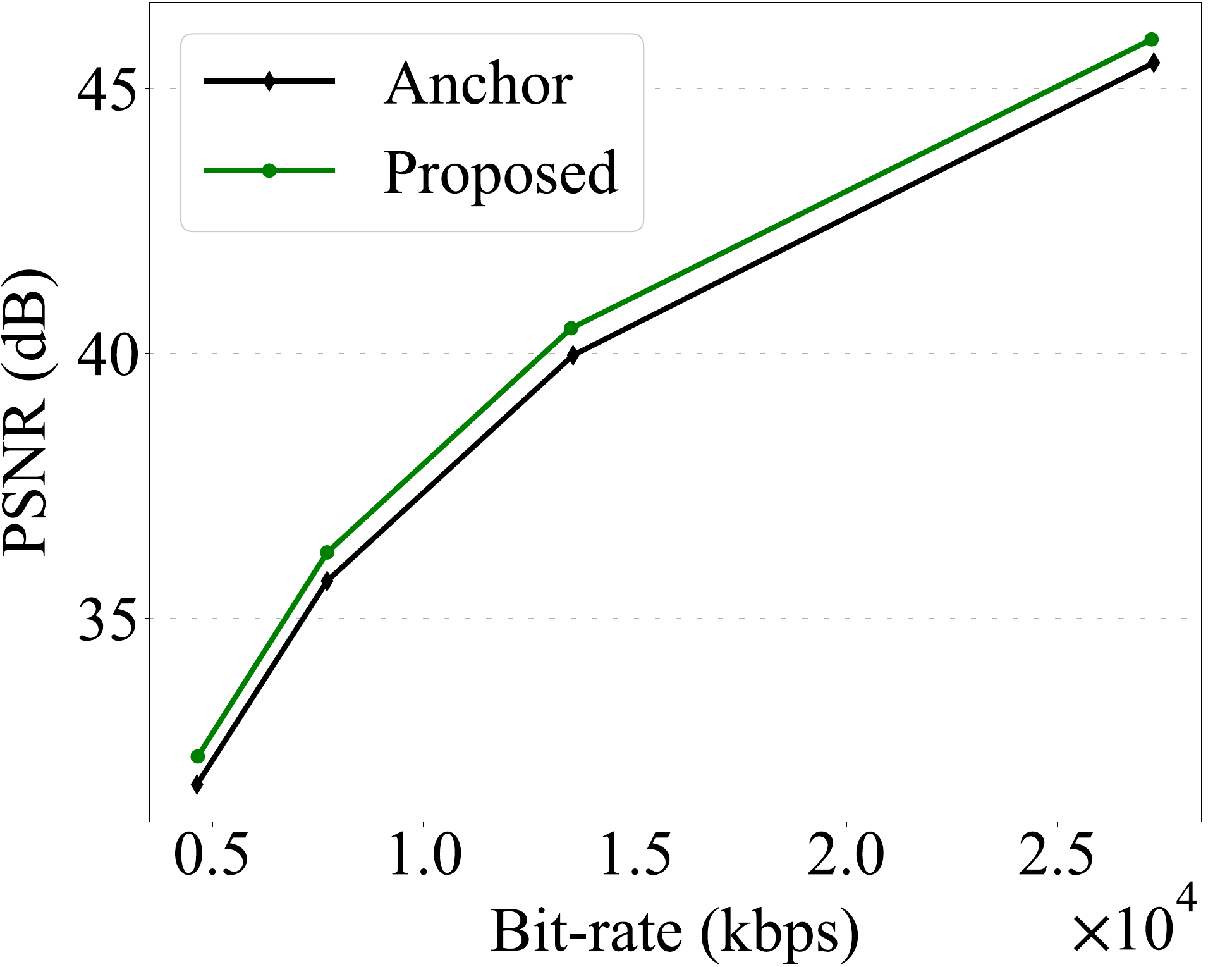}
			}
			\caption{Luma rate-distortion curves for different sequences compressed by VTM-11.0 with and without our proposed DWF under RA and LDB configurations. (a) \textit{BQSquare} with RA coding; (b) \textit{FlyingGraphics} with RA coding; (c) \textit{BQSquare} with LDB coding; (d) \textit{FlyingGraphics} with LDB coding.}
			\label{fig::RD-curve}
		\end{center}
		\vspace{-3mm}
	\end{figure*}
	
	\begin{figure*}[!t]
		\begin{center}
			\noindent
			\subfigure[]{
				\includegraphics[width=2.0in]{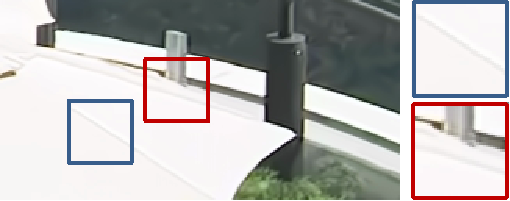}
			}
                \subfigure[]{
				\includegraphics[width=2.0in]{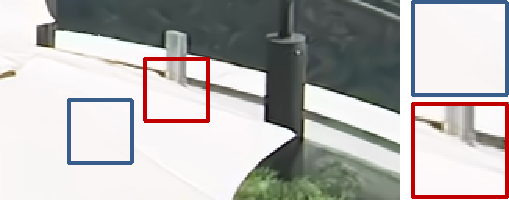}
			}
                \subfigure[]{
				\includegraphics[width=2.0in]{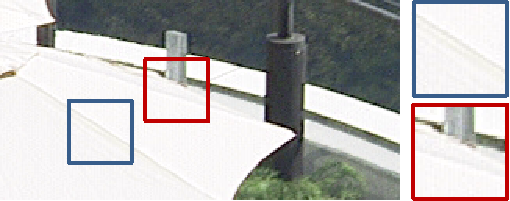}
			}
			\subfigure[]{
				\includegraphics[width=2.0in]{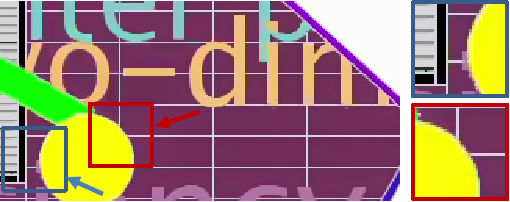}
			}	
			\subfigure[]{
				\includegraphics[width=2.0in]{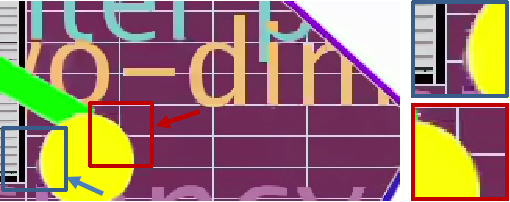}
			}
			\subfigure[]{
				\includegraphics[width=2.0in]{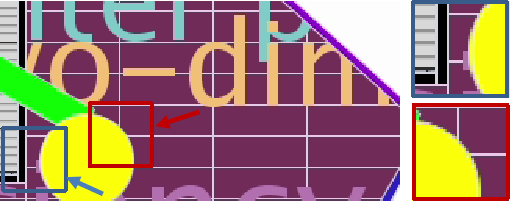}
			}
			\caption{Subjective quality comparison. (a), (b), (c) \textit{BQTerrace}, POC=77, QP=32, LDB; (d), (e), (f) \textit{FlyingGraphics}, POC=19, QP=32, LDB. For both test sequences, DWF is enabled in the left pictures and disabled in the middle pictures, and the right pictures are original pictures.}
			\label{fig::subjective}
		\end{center}
		\vspace{-3mm}
	\end{figure*}

\begin{table}[!t]
		\centering
		\begin{center}
			\caption{{Experimental results of the proposed DWF~(Cb component), Anchor: VTM-11.0.}}
			\label{table::PerformanceDWF-Cb}
			\setlength{\tabcolsep}{4.1mm}{
                    {
				\begin{tabular}{c|c c c}
					\thickhline
					\hline
					{\textbf{Class}}& {\textbf{AI}} & {\textbf{RA}}& {\textbf{LDB}} \\
					\hline
					{Class A1}      & -0.27\%   & -0.38\%     & -                         \\
					{Class A2}      & -0.37\%   & -0.52\%     & -                    \\
					{Class B}       & -0.45\%   & -0.15\%     & -0.93\%          \\
					{Class C}       & -1.09\%   & -0.61\%     & -1.12\%         \\
					{Class D}       & -0.28\%   & -0.38\%     & -0.32\%          \\
					{Class E}       & -1.54\%   & -           & -1.62\%          \\
					{Class F}       & -1.06\%   & -1.27\%     & -0.35\%      \\
					{Class G}       & -0.13\%   & -0.07\%     & -0.66\%        \\
					{Class M}       & -0.57\%   & -0.03\%     & -0.36\%      \\
					\hline
					{Average (CTC)} & -0.73\%   & -0.39\%     & -1.17\%        \\
					\hline
					{Average (All)} & -0.63\%   & -0.39\%     & -0.72\%       \\
					\thickhline
			\end{tabular}}}
		\end{center}
		\vspace{-3mm}
\end{table}

\begin{table}[!t]
		\centering
		\begin{center}
			\caption{{Experimental results of the proposed DWF~(Cr component), Anchor: VTM-11.0.}}
			\label{table::PerformanceDWF-Cr}
			\setlength{\tabcolsep}{4.1mm}{
                    {
				\begin{tabular}{c|c c c}
					\thickhline
					\hline
					{\textbf{Class}}& {\textbf{AI}} & {\textbf{RA}}& {\textbf{LDB}} \\
					\hline
					{Class A1}      & -0.71\%  & -0.71\%     & -                \\
					{Class A2}      & -0.74\%  & -0.64\%     & -                \\
					{Class B}       & -0.80\%  & -0.28\%     & -0.88\%          \\
					{Class C}       & -1.93\%  & -1.36\%     & -2.25\%          \\
					{Class D}       & -0.51\%  & -0.37\%     & -0.66\%          \\
					{Class E}       & -0.81\%  & -           & -0.97\%          \\
					{Class F}       & -1.38\%  & -1.52\%     & -0.91\%          \\
					{Class G}       & -0.25\%  & -0.10\%     & -0.77\%          \\
					{Class M}       & -0.41\%  & -0.59\%     & -0.06\%          \\
					\hline
					{Average (CTC)} & -1.03\%  & -0.73\%     & -1.36\%          \\
					\hline
					{Average (All)} & -0.82\%  & -0.68\%     & -0.88\%          \\
					\thickhline
			\end{tabular}}}
		\end{center}
		\vspace{-3mm}
\end{table}

\section{Experimental Results}
To evaluate the effectiveness of the proposed method, we integrate it into the VVC reference software VTM-11.0, in which the proposed DWF is applied as an in-loop filter located after SAO and before ALF.
 \subsection{Test Conditions}
	The videos recommended by JVET and several screen content videos~(Class G and Class M) are adopted in the experiment. Three configurations, AI, RA, and LDB, conforming to the common test condition~\cite{CTC} are employed in the simulation. For each sequence, two seconds are encoded at four typical QPs \{22, 27, 32, 37\}. The coding performance is measured by BD-rate. For the screen content videos in Class G and M, we follow the test conditions and configurations recommended by \cite{Q2013}. The encoding complexity is calculated by,
	\begin{equation}
	\Delta ET = \frac{TEnc_{\mathrm{test}}}{TEnc_\mathrm{{anchor}}} \times 100\%,
	\end{equation}
	\noindent where $TEnc_{\mathrm{test}}$ and $TEnc_\mathrm{{anchor}}$ represent the consumed encoding time with the tested scheme and the anchor. The decoding complexity is calculated by Eqn.~(\ref{decT}).

\subsection{DWF Performance}
	Table \ref{table::Performance1} shows the coding performance of the proposed DWF on the luma component compared to the VTM-11.0 under AI, RA, and LDB configurations. It is observed that the proposed DWF provides consistent coding gain at various resolutions. In particular, the proposed DWF achieves 0.58\%, 1.09\%, and 2.01\% BD-rate savings under AI, RA, and LDB configurations on average for CTC videos~(Class A1, A2, B, C, and E). When including all the test sequences, -1.16\%, -1.92\%, and -2.67\% coding gain can be observed under AI, RA, and LDB configurations, respectively. The highest coding gain is -5.79\%, -6.69\%, and -5.85\% under AI, RA, and LDB configurations. For natural sequences, the coding performance of \textit{BQTerrace}, \textit{BQSquare}, and \textit{Johnny} is considerably better than other sequences. The particular favorable results of these sequences are mainly attributable to the fact that there are numerous repeated patterns, such as the floor and fence in \textit{BQTerrace} and \textit{BQSquare}, and the striped suit in \textit{Johnny}. These repeated patterns can be effectively reconstructed by non-local reference samples. Although the camera motion in \textit{BQTerrace} and \textit{BQSquare} is more violent than \textit{Johnny}, their coding performance is comparable to \textit{Johnny}, which verifies the solidness of the proposed DWF. For screen content videos, the average bit-rate reductions are 2.56\%, 3.39\%, and 4.03\%, which are notably better than natural sequences. One explanation for this is that screen content videos have many repeated patterns. The other explanation is that these sequences tend to have relatively simple structure and texture, which means that the current classification method can well capture the features. Furthermore, the performance of DWF for RA and LDB is generally better than that for AI. One reason is that the temporal block matching under RA and LDB configurations can derive better similar blocks, leading to more reliable reference samples. The other reason is that the effect of DWF can be propagated to succeeding pictures under RA and LDB configurations, yielding more accurate inter-prediction.

    {Table \ref{table::PerformanceDWF-Cb} and Table \ref{table::PerformanceDWF-Cr} are the coding performance of the proposed DWF on Cb and Cr component, respectively. It can be observed that the proposed DWF achieves consistent coding performance improvement for Cb and Cr components. }
	
	\begin{figure}[!t]
		\begin{center}
			\noindent
			\includegraphics[width=2.7in]{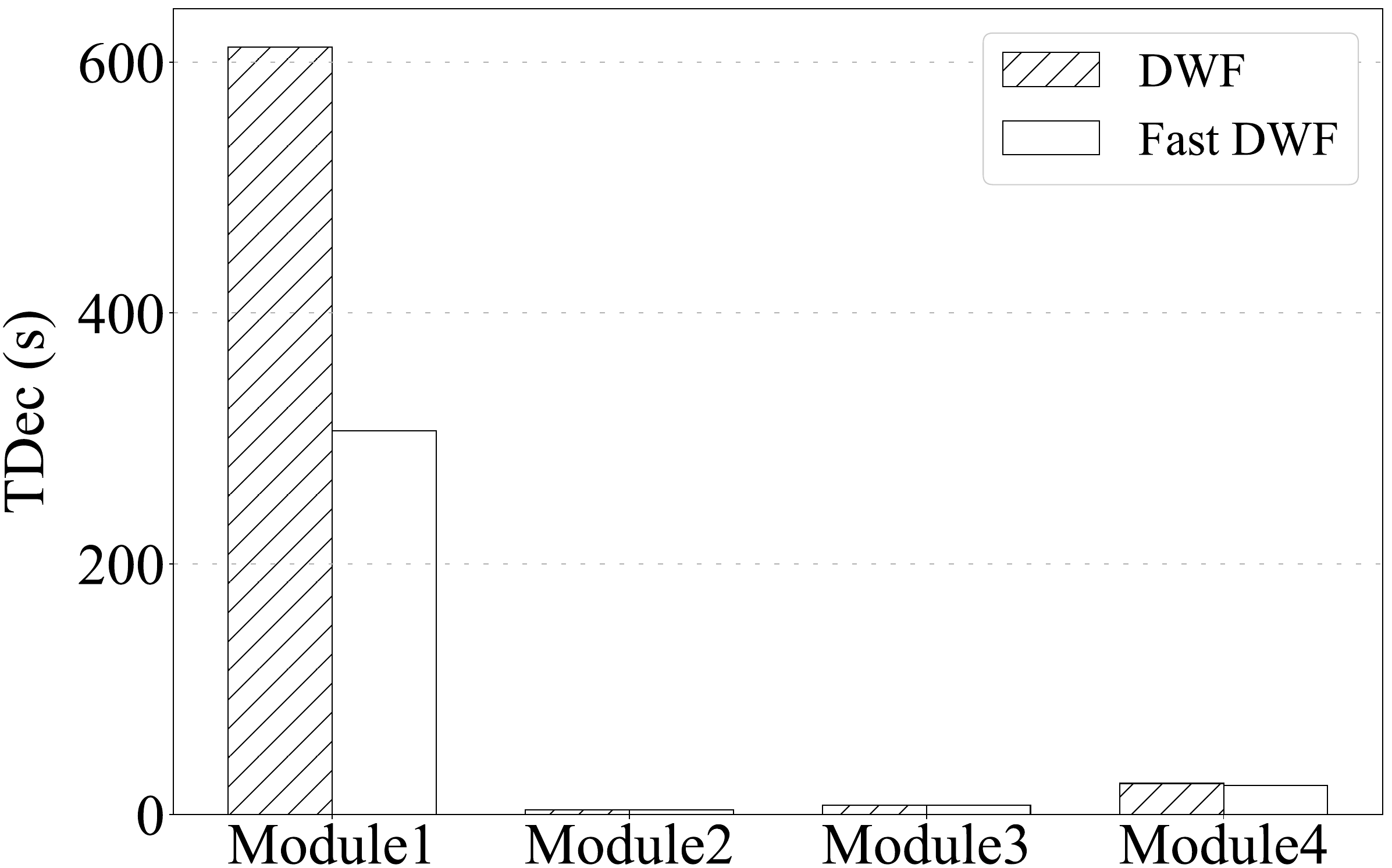}
			\caption{Decoder running time for the main modules of the proposed DWF. Module1: spatio-temporal reference sample derivation. Module2: content-aware filter adaptation. Module3: reference sample fusion. Module 4: adaptive filtering operation.}
			\label{fig::complexity}
		\end{center}
		\vspace{-3mm}
	\end{figure}
	
\begin{table*}[!t]
		\centering
		\begin{center}
			\caption{Experimental results of the proposed DWF with the fast algorithm~(Y component), Anchor: VTM-11.0.} 
			\label{table::PerformanceFast}
			\setlength{\tabcolsep}{4.1mm}{
				\begin{tabular}{c|c c c|c c c|c c c}
					\thickhline
					\hline
					\multirow{2}{*}{\textbf{Class}}&
					\multicolumn{3}{c|}{\rule{0pt}{8pt} \textbf{AI}} & \multicolumn{3}{c|}{\textbf{RA}}& \multicolumn{3}{c}{\textbf{LDB}}\\
					\cline{2-10} &
					\textbf{BD-Rate}& $\Delta \textit{\textbf{ET}}$ & $\Delta \textit{\textbf{DT}}$ & \textbf{BD-Rate}& $\Delta \textit{\textbf{ET}}$ & $\Delta \textit{\textbf{DT}}$ & \textbf{BD-Rate} & $\Delta \textit{\textbf{ET}}$ & $\Delta \textit{\textbf{DT}}$ \\
					\hline
					{Class A1}                    & -0.47\% & 131\%  & 1014\%   & -0.66\% & 107\%  &  424\%    & -       & - & -                   \\
					{Class A2}                    & -0.62\% & 116\%  & 1499\%   & -0.86\% & 106\%  &  441\%    & -       & - & -                   \\
					{Class B}                     & -0.54\% & 116\%  & 1024\%   & -1.25\% & 106\%  & 1733\%    & -1.76\% & 106\%  & 1113\%         \\
					{Class C}                     & -0.43\% & 112\%  & 1155\%   & -1.06\% & 105\%  & 948\%     & -1.32\% & 105\%  & 1735\%         \\
					{Class D}                     & -0.12\% & 113\%  &  696\%   & -1.71\% & 105\%  & 1796\%    & -1.87\% & 105\%  & 1396\%         \\
					{Class E}                     & -0.93\% & 115\%  & 1988\%   & -       & -      & -         & -2.64\% & 109\%  &  772\%         \\
					{Class F}                     & -1.32\% & 108\%  & 1634\%   & -1.33\% & 109\%  &  800\%    & -1.53\% & 107\%  &  453\%         \\
					{Class G}                     & -2.90\% & 108\%  & 2271\%   & -3.46\% & 108\%  & 2027\%    & -3.95\% & 107\%  & 1584\%         \\
					{Class M}                     & -2.33\% & 107\%  & 2072\%   & -2.96\% & 112\%  & 1420\%    & -3.74\% & 108\%  & 1394\%         \\
					\hline
					{Average (CTC)}               & -0.58\% & 118\%  & 1291\%   & -1.00\% & 106\%  & 1004\%     & -1.83\% & 106\%  & 1235\%         \\
					\hline
					{Average (All)}               & -1.16\% & 113\%  & 1502\%   & -1.78\% & 107\%  & 1275\%     & -2.46\% & 107\%  & 1231\%         \\
					
					\thickhline
			\end{tabular}}
		\end{center}
		\vspace{-3mm}
	\end{table*}

	Fig.~\ref{fig::RD-curve} shows the rate-distortion curves for test sequences, \textit{BQSquare}, and \textit{FlyingGraphics}, under RA and LDB configurations. We can observe that coding performance is significantly improved in a large bit-rate range with our proposed DWF. An interesting phenomenon is that the Peak signal-to-noise ratio~(PSNR) improvement of \textit{FlyingGraphics} is more stable than \textit{BQSquare}. For \textit{BQSquare}, the PSNR improvement is especially obvious for the high bit-rate range. This is mainly due to the fact that details are severely lost in compressed frames coded with low bit rates. These frames are difficult to be restored in video compression.  
	
	Fig.~\ref{fig::subjective} shows the subjective results for natural sequence \textit{BQTerrace} and screen content video \textit{FlyingGraphics}. The images in Fig.~\ref{fig::subjective}(a) and (d) are reconstructed by the VTM-11.0 with the proposed DWF, and the images in Fig.~\ref{fig::subjective}(b) and (e) are reconstructed by the VTM-11.0 without the proposed DWF, and the last two images are the original images. We can see that our proposed DWF can efficiently recover the destroyed details, such as the folds on the umbrella in \textit{BQTerrace} and the content next to the yellow circle in \textit{FlyingGraphics}.

\begin{table}[!t]
		\centering
		\begin{center}
			\caption{Experimental results of the proposed DWF + VTM-11.0 w/o ALF~(Y component), Anchor: VTM-11.0.}
			\label{table::PerformanceALF}
			\setlength{\tabcolsep}{4.1mm}{
				\begin{tabular}{c|c c c}
					\thickhline
					\hline
					{\textbf{Class}}& {\textbf{AI}} & {\textbf{RA}}& {\textbf{LDB}} \\
					\hline
					{Class A1}      & -0.03\%  & -0.32\%     & -                         \\
					{Class A2}      &  0.03\%  & -0.11\%     & -                    \\
					{Class B}       &  0.06\%  & -0.53\%     & -0.99\%          \\
					{Class C}       &  0.92\%  &  1.35\%     &  1.08\%         \\
					{Class D}       &  0.78\%  &  0.80\%     & -0.61\%          \\
					{Class E}       &  1.50\%  & -           &  0.73\%          \\
					{Class F}       & -0.44\%  & -0.75\%     & -0.71\%      \\
					{Class G}       & -2.37\%  & -3.39\%     & -3.34\%        \\
					{Class M}       & -2.00\%  & -2.91\%     & -3.19\%      \\
					\hline
					{Average (CTC)} & 0.47\%   & 0.10\%     &  0.13\%        \\
					\hline
					{Average (All)} & -0.32\%   & -0.89\%     & -1.21\%       \\

                        \hline
					{$\Delta ET$} & 113\% & 107\%  & 107\%     \\
					\hline
					{$\Delta DT$} & 1277\% & 1514\%  & 1425\%     \\
     
					\thickhline
			\end{tabular}}
		\end{center}
		\vspace{-3mm}
\end{table}

\begin{table}[!t]
		\centering
		\begin{center}
			\caption{Experimental results of the NLSF~\cite{K0160}~(Y component), Anchor: VTM-11.0.}
			\label{table::PerformanceNLSF}
			\setlength{\tabcolsep}{4.1mm}{
				\begin{tabular}{c|c c c}
					\thickhline
					\hline
					{\textbf{Class}}& {\textbf{AI}} & {\textbf{RA}}& {\textbf{LDB}} \\
					\hline
					{Class A1}                    & -0.70\%     & -0.65\%      & -                         \\
					{Class A2}                    & -0.66\%     & -0.66\%      & -                         \\
					{Class B}                     & -0.77\%     & -1.01\%      & -1.40\%            \\
					{Class C}                     & -1.01\%     & -0.75\%      & -1.12\%        \\
					{Class D}                     & -0.71\%     & -0.02\%      & -0.45\%          \\
					{Class E}                     & -2.09\%     & -            & -2.17\%            \\
					{Class F}                     & -0.83\%     & -0.50\%      & -1.24\%        \\
					{Class G}                     & -0.93\%     & -1.26\%      & -1.23\%       \\
					{Class M}                     & -0.87\%     & -0.56\%      & -1.07\%        \\
					\hline
					{Average (CTC)}               & -1.02\%     & -0.80\%      & -1.50\%        \\
					\hline
					{Average (All)}               & -0.93\%     & -0.68\%      & -1.20\%       \\
                    \hline
					{$\Delta ET$}               &  122\%  & 126\%  &  124\%      \\
					\hline
					{$\Delta DT$}               & 2574\%  &3265\%  & 3968\%      \\
					\thickhline
			\end{tabular}}
		\end{center}
		\vspace{-3mm}
\end{table}

\subsection{Complexity Analysis and Fast Algorithm}
We also evaluate the computational complexity by comparing the encoder and decoder running times. The executable files are compiled by Microsoft Visual Studio 2017, 64bit. The testing is conducted on Windows 10 operating system with an Intel(R) Core i7-8700 @3.20GHz CPU. In terms of computational complexity, on average 13\%, 15\%, and 14\% encoding time increment is observed for AI, RA, and LDB configurations, respectively. The decoding time is increased by 1402\%, 2380\%, and 2288\% on average. We investigate the percentage of running time for the main modules of the proposed DWF, which is depicted in Fig.~\ref{fig::complexity}. It is obvious that \textit{spatio-temporal reference sample derivation} is the most time-consuming module. Herein, the derivation of reference samples takes up more than 80\% of the running time, mainly due to the block matching process. For natural videos, there are 1024~($32\times 32$) patches in each CTU~(size $128\times 128$). For each patch, at most $1600$~($24\times 24+16\times 16\times 4$) patches need to be searched in the spatial and temporal search area. For each searched patch, the $L2$ norm needs to be calculated, which introduces 36 multiplication and 71 addition. In total, at most 58984240 multiplication and 116326400 addition are conducted in each CTU. For each sample, 3600 multiplication and 7100 addition are conducted at most.

\begin{table*}[!t]
		\centering
		\begin{center}
			\caption{Experimental results of the DWF with different local and non-local reference sample fusion method~(Y component), Anchor: VTM-11.0. $R_{N}$ is the number of non-local reference samples. $R_{L}$ is the number of local reference samples.} 
			\label{table::PerformanceLocalN}
 			\setlength{\tabcolsep}{4.1mm}{
				\begin{tabular}{c|c c c|c c c}
					\thickhline
					\hline
					\multirow{2}{*}{\textbf{Class}}&
					\multicolumn{3}{c|}{\rule{0pt}{8pt} \textbf{All Local: $R_{N}=0$}} & \multicolumn{3}{c}{\textbf{All Non-local: $R_{L}=0$}}\\
					\cline{2-7} &
					                  \textbf{AI}& \textbf{RA} & \textbf{LDB} & \textbf{AI} & \textbf{RA} & \textbf{LDB} \\
					\hline
					{Class A1}                    & -0.34\%  & -0.37\%  & -         & -0.30\%  & -0.53\%  & -    \\
					{Class A2}                    & -0.57\%  & -0.53\%  & -         & -0.35\%  & -0.57\%  & -   \\
					{Class B}                     & -0.33\%  & -0.60\%  & -1.04\%   & -0.41\%  & -0.89\%  & -1.60\% \\
					{Class C}                     & -0.05\%  &  0.04\%  & -0.02\%   & -0.31\%  & -1.05\%  & -1.44\% \\
					{Class D}                     &  0.01\%  & -0.34\%  & -0.24\%   & -0.02\%  & -1.43\%  & -2.02\% \\
					{Class E}                     & -0.12\%  & -        & -0.58\%   & -0.77\%  & -        & -2.57\% \\
					{Class F}                     & -0.86\%  & -0.34\%  & -0.82\%   & -0.84\%  & -1.03\%  & -1.53\% \\
					{Class G}                     & -2.30\%  & -2.27\%  & -2.56\%   & -1.72\%  & -2.56\%  & -2.99\%\\
					{Class M}                     & -1.83\%  & -2.07\%  & -2.24\%   & -1.22\%  & -1.67\%  & -2.44\%\\
					\hline
					{Average (CTC)}               & -0.27\%  & -0.37\%  & -0.59\%   & -0.42\%  & -0.80\%  & -1.79\% \\
					\hline
					{Average (All)}               & -0.79\%  & -0.90\%  & -1.17\%   & -0.70\%  & -1.27\%  & -2.07\% \\
					
					\thickhline
			\end{tabular}}
		\end{center}
		\vspace{-3mm}
\end{table*}

To reduce the complexity of block matching, we make further optimization for the proposed DWF with the two-step fast algorithm presented in \cite{meng2018optimized}. This algorithm is based on a fixed searching template according to image spatial statistical characteristics. Specifically, the first step is to derive several similar blocks with the highest similarity based on a diamond searching template. Then, take these similar blocks as the center blocks and search for similar blocks recursively by using the diamond template. Based on the results in \cite{meng2018optimized}, the two-step fast algorithm can save roughly 70\% decoding time of the block matching process. In this paper, the fast algorithm is only applied on inter-predicted frames, i.e., B-frames and P-frames. As shown in Table \ref{table::PerformanceFast}, the fast algorithm can achieve significant time savings with negligible performance loss. Compared to Table \ref{table::Performance1}, the optimized DWF achieves 49\% and 48\% time savings with 0.14\% and 0.21\% coding performance change. With the fast algorithm, the percentage of Module1 is lower than before, as depicted in Fig.~\ref{fig::complexity}. To further reduce the complexity, simplification approaches of block matching can be investigated in the future.

\subsection{Comparision with the State-of-the-art Methods}
To validate the effectiveness of the proposed DWF, we compare it with the state-of-the-art local and non-local methods, including ALF in VTM-11.0 and non-local structure based in-loop filter~(NLSF) proposed in \cite{K0160}. The results of ALF are shown in Table \ref{table::PerformanceALF}. In this experiment, the anchor is the VTM-11.0, and the test is the VTM-11.0 with DWF replacing ALF. It can be observed that compared to ALF, DWF shows -0.32\%, -0.89\%, and -1.21\% coding gain under AI, RA, and LDB configurations, respectively. For large-resolution and screen-content videos, the performance of DWF is better than ALF. The main reason lies in that these videos have more repetitive patterns, hence the non-local reference samples can improve the coding performance. For natural videos with small resolutions, the performance of ALF is better than DWF. The main reason lies in that the non-local characteristic of small-resolution videos is not as good as that of large-resolution videos. For better performance, more fine-tuned parameter signaling, more adaptive classification, and more flexible reference sample derivation methods can be investigated in the future.

The results of NLSF are shown in Table \ref{table::PerformanceNLSF}. In this experiment, the anchor is VTM-11.0, and the test is VTM-11.0 with NLSF located after SAO and before ALF. It can be observed that the proposed DWF achieves better coding gain with lower encoding and decoding complexity under RA and LDB configurations. The performance improvement is notably higher for screen content videos. The main reason is that the noise distribution of screen content videos is difficult to model, which means that the compression noise is difficult to be removed by off-line trained parameters of NLSF. The performance of DWF is inferior to NLSF under AI configuration for natural videos. The main reason may be that there is no reference frame that can be utilized in the reference sample derivation process under AI configuration. Hence, the similarity of selected blocks is relatively low compared to that of RA and LDB configurations. And when the non-local similarity is not very strong, SVD-based methods may perform better than Wiener-based methods.

\begin{figure}[!t]
		\begin{center}
			\noindent
			\includegraphics[width=3.0in]{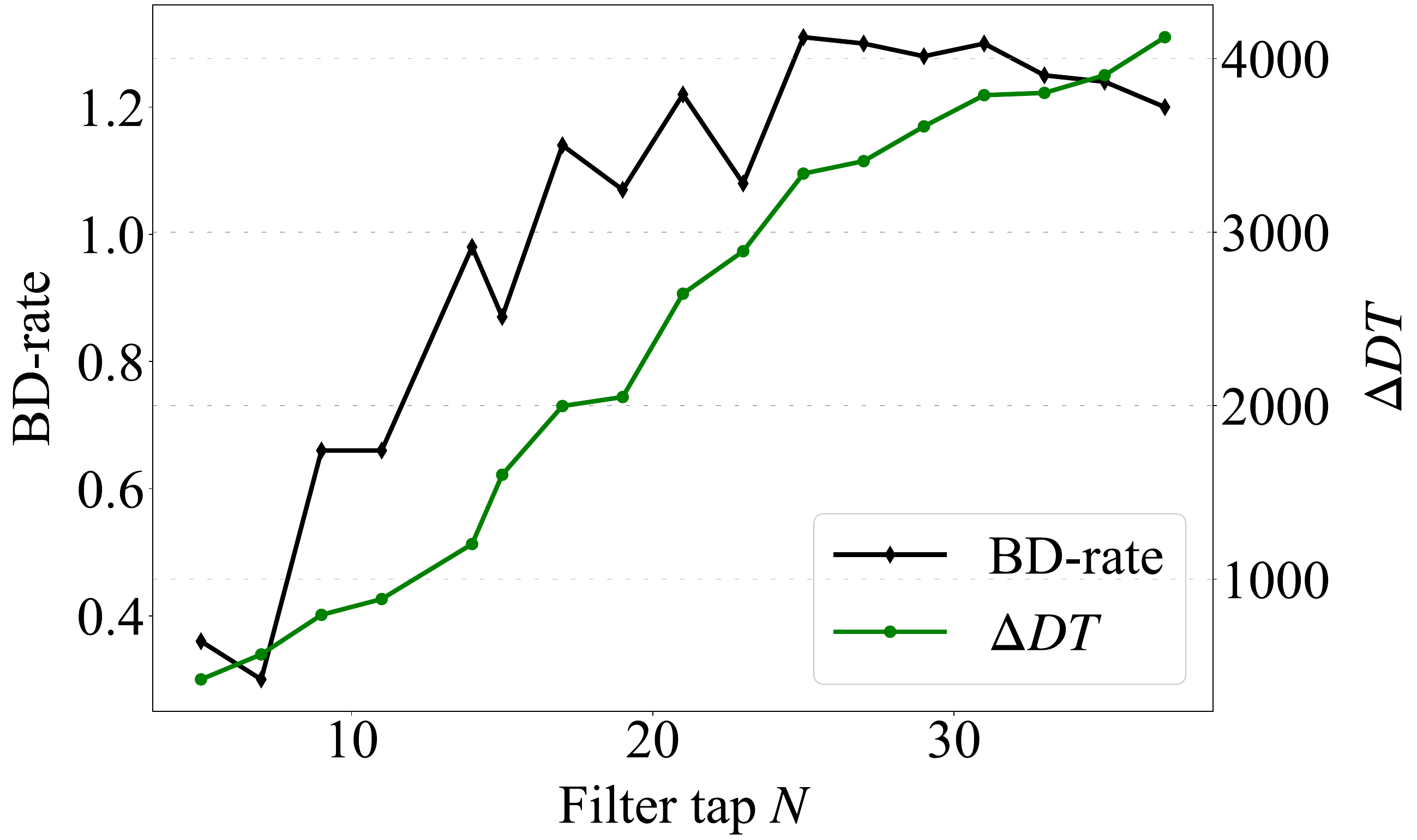}
			\caption{The influence of filter tap $N$ on BD-rate and decoding complexity $\Delta DT$. The left y-axis is the absolute value of BD-rate, for better visualization.}
			\label{fig::BDrate-complexity}
		\end{center}
		\vspace{-3mm}
	\end{figure}

\subsection{Ablation Study}
For a better understanding of the benefits of local and non-local fusion schemes, we conducted experiments by adjusting the local and non-local reference sample number $R_{L}$ and $R_{N}$ defined in Section \ref{noiseAna}, which is shown in Table \ref{table::PerformanceLocalN}. In this table, ``All Local'' indicates that $R_{N}$ is set to 0, and $R_{L}$ is set to 25. ``All Non-local'' indicates that $R{L}$ is set to 0, and $R_{N}$ is set to 25. It can be observed that all non-local settings can achieve better performance compared to all local settings. And both settings can not achieve competitive performance gain compared to the proposed DWF in Table \ref{table::Performance1}.

We also investigate the influence of filter tap $N$ on the BD-rate and decoder complexity, which is shown in Fig.~\ref{fig::BDrate-complexity}. In this experiment, test sequences are videos in Class B, and the configuration is RA. It can be observed that with the value of $N$ increasing, the coding gain shows an upward trend. However, when the filter taps $N$ is larger than 25, the coding gain shows a downward trend. One reason is that the increment of $N$ can increase the overhead caused by the signaling of filter coefficients. The other reason is that when $N$ is larger than 25, the similarity between the current block and its selected non-local similar blocks becomes lower. These unrelated blocks may bring negative effects on the overall performance.

\section{Conclusion}
    In this paper, a novel in-loop filtering method named DWF was proposed to improve the objective and subjective quality of compressed videos. We first identify the importance of the local and non-local combination of loop filters. Then we proposed an adaptive local and non-local reference sample derivation scheme. By incorporating the reference sample derivation method, a content-aware filter adaptation approach is designed, which combines the block-level and sample-level characteristics. We also investigate the influence of the reference sample derivation scheme on performance and complexity. Experimental results show that the proposed DWF can further improve the compression performance of VVC, especially for video sequences with lots of repetitive patterns, such as screen content videos. The analysis and observation in this paper contribute valuable directions for further improving the restoration capability of in-loop filters. As one of the first attempts to combine local and non-local samples, the proposed method could be improved in the future using GPU-based fast algorithms. For better coding performance, more advanced classification and more adaptive reference sample derivation approaches can also be explored.
    
\section*{ACKNOWLEDGMENT}
The authors would like to thank the associate editor and anonymous reviewers for their constructive comments that significantly helped them in improving the quality of the paper. The authors would also like to acknowledge Dr. Wenhao Zhang and Scott Labrozzi for the fruitful discussion.

\bibliographystyle{IEEEtran}
\bibliography{icme2020template}

@article{HEVC,
	title={Overview of the {H}igh {E}fficiency {V}ideo {C}oding ({HEVC}) standard},
	author={Sullivan, Gary J and Ohm, Jens-Rainer and Han, Woo-Jin and Wiegand, Thomas},
	journal={IEEE Transactions on Circuits and Systems for Video Technology},
	volume={22},
	number={12},
	pages={1649--1668},
	year={2012}
}

@article{VVC,
	title={Overview of the {v}ersatile {v}ideo {c}oding ({VVC}) standard and its applications},
	author={Bross, Benjamin and Wang, Ye-Kui and Ye, Yan and Liu, Shan and Chen, Jianle and Sullivan, Gary J and Ohm, Jens-Rainer},
	journal={IEEE Transactions on Circuits and Systems for Video Technology},
	volume={31},
	number={10},
	pages={3736--3764},
	year={2021},
	publisher={IEEE}
}

@article{VTM11.0,
	title={Algorithm description for {V}ersatile {V}ideo {C}oding and {T}est {M}odel 11 ({VTM} 11)},
	author={Chen, Jianle and Ye, Yan and Kim, Seung Hwan},
	journal={Joint Video Experts Team (JVET), doc. JVET-T2002},
	year={Oct. 2020}
}

@article{CTC,
	title={{JVET} common test conditions and software reference configurations for {SDR} video},
	author={Bossen, Frank and Boyce, Jill and S{\"u}hring, Karsten and Li, Xiang and Seregin, Vadim },
	journal={Joint Video Experts Team (JVET), doc. JVET-N1010},
	year={Mar. 2019}
}

@article{BDrate,
	title={Calculation of average {PSNR} differences between {RD}-curves},
	author={Bj{\o}ntegarrd, Gisle},
	journal={ITU-T SG16/Q6 (VCEG), doc. VCEG-M33},
	year={Apr. 2001}
}

@article{norkin2012hevc,
  title={{HEVC} deblocking filter},
  author={Norkin, Andrey and Bj{\o}ntegarrd, Gisle and Fuldseth, Arild and Narroschke, Matthias and Ikeda, Masaru and Andersson, Kenneth and Zhou, Minhua and Van der Auwera, Geert},
  journal={IEEE Transactions on Circuits and Systems for Video Technology},
  volume={22},
  number={12},
  pages={1746--1754},
  year={2012},
  publisher={IEEE}
}

@article{SAO,
	title={Sample {A}daptive {O}ffset in the {HEVC} standard},
	author={Fu, Chih-Ming and Alshina, Elena and Alshin, Alexander and Huang, Yu-Wen and Chen, Ching-Yeh and Tsai, Chia-Yang and Hsu, Chih-Wei and Lei, Shaw-Min and Park, Jeong-Hoon and Han, Woo-Jin},
	journal={IEEE Transactions on Circuits and Systems for Video Technology},
	volume={22},
	number={12},
	pages={1755--1764},
	year={2012},
	publisher={IEEE}
}

@article{tsai2013adaptive,
  title={Adaptive loop filtering for video coding},
  author={Tsai, Chia-Yang and Chen, Ching-Yeh and Yamakage, Tomoo and Chong, In Suk and Huang, Yu-Wen and Fu, Chih-Ming and Itoh, Takayuki and Watanabe, Takashi and Chujoh, Takeshi and Karczewicz, Marta and others},
  journal={IEEE Journal of Selected Topics in Signal Processing},
  volume={7},
  number={6},
  pages={934--945},
  year={2013},
  publisher={IEEE}
}

@inproceedings{karczewicz2016geometry,
	title={Geometry transformation-based adaptive in-loop filter},
	author={Karczewicz, Marta and Zhang, Li and Chien, Wei-Jung and Li, Xiang},
	booktitle={Picture Coding Symposium (PCS)},
	pages={1--5},
	year={2016},
	organization={IEEE}
}

@article{GALF-JVET,
	title={{CE}2.4.1.4: Reduced filter shape size for {ALF})},
	author={Karczewicz, Marta and Shlyakhov, Nikolay and Hu, Nan and others},
	journal={Joint Video Experts Team (JVET), doc. JVET-K0371},
	year={Jul. 2018}
}

@article{temporal-ALF,
	title={{CE}5: {C}oding tree block based adaptive loop filter ({CE}5-4)},
	author={Hu, Nan and Seregin, Vadim and Egilmez, Hilmi Enes and Karczewicz, Marta},
	journal={Joint Video Experts Team (JVET), doc. JVET-N0415},
	year={Mar. 2019}
}

@article{nonlinear-ALF,
	title={{CE}5: {R}esults of tests {CE}5-3.1 to {CE}5-3.4 on {N}on-Linear {A}daptive {L}oop {F}ilter},
	author={Taquet, Jonathan and Onno, Patrice and Gisquet, Christophe and Laroche, Guillaume},
	journal={Joint Video Experts Team (JVET), doc. JVET-N0242},
	year={Mar. 2019}
}

@inproceedings{CCALF,
	title={On Cross Component Adaptive Loop Filter for Video Compression},
	author={Misra, Kiran and Bossen, Frank and Segall, Andrew},
	booktitle={Picture Coding Symposium (PCS)},
	pages={1--5},
	year={2019},
	organization={IEEE}
}

@article{CCALF-JVET,
	title={{CC-ALF}: {I}ntegrated Text for the Cross Component Adaptive Loop Filter},
	author={Misra, Kiran and others},
	journal={Joint Video Experts Team (JVET), doc. JVET-Q0795},
	year={Jan. 2020}
}

@article{Q2013,
	title={{JVET} common test conditions and software reference configurations for non-4:2:0 colour formats},
	author={Chao, Yung-Hsuan and Sun, Yu-Chen and Xu, Jizheng and Xu, Xiaozhong},
	journal={Joint Video Experts Team (JVET), doc. JVET-Q2013},
	year={Jan. 2020}
}

@article{karczewicz2021vvc,
  title={{VVC} in-loop filters},
  author={Karczewicz, Marta and Hu, Nan and Taquet, Jonathan and Chen, Ching-Yeh and Misra, Kiran and Andersson, Kenneth and Yin, Peng and Lu, Taoran and Fran\c{c}ois, Edouard and Chen, Jie},
  journal={IEEE Transactions on Circuits and Systems for Video Technology},
  volume={31},
  number={10},
  pages={3907--3925},
  year={2021},
  publisher={IEEE}
}

@inproceedings{lu2020luma,
  title={Luma mapping with chroma scaling in versatile video coding},
  author={Lu, Taoran and Pu, Fangjun and Yin, Peng and McCarthy, Sean and Husak, Walt and Chen, Tao and Francois, Edouard and Chevance, Christophe and Hiron, Franck and Chen, Jie and others},
  booktitle={Data Compression Conference (DCC)},
  pages={193--202},
  year={2020},
  organization={IEEE}
}

@inproceedings{wennersten2017bilateral,
  title={Bilateral filtering for video coding},
  author={Wennersten, Per and Str{\"o}m, Jacob and Wang, Ying and Andersson, Kenneth and Sj{\"o}berg, Rickard and Enhorn, Jack},
  booktitle={Visual Communications and Image Processing (VCIP)},
  pages={1--4},
  year={2017},
  organization={IEEE}
}

@inproceedings{strom2019bilateral,
  title={Bilateral Loop Filter in Combination with SAO},
  author={Str{\"o}m, Jacob and Wennersten, Per and Enhorn, Jack and Liu, Du and Andersson, Kenneth and Sj{\"o}berg, Rickard},
  booktitle={Picture Coding Symposium (PCS)},
  pages={1--5},
  year={2019},
  organization={IEEE}
}

@article{P0078-hadamard,
	title={{CE}5-3: Combination of Hadamard filter and {SAO} ({CE}5-3.2, {CE}5-3.4)},
	author={Ikonin, Sergey and Stepin, Victor and Karabutov, Alexander and Nikolaeva, Sofya},
	journal={Joint Video Experts Team (JVET), doc. JVET-P0078},
	year={Oct. 2019}
}

@inproceedings{rasch2018signal,
  title={A signal adaptive diffusion filter for video coding},
  author={Rasch, Jennifer and Pfaff, Jonathan and Sch{\"a}fer, Michael and Schwarz, Heiko and Winken, Martin and Siekmann, Mischa and Marpe, Detlev and Wiegand, Thomas},
  booktitle={Picture Coding Symposium (PCS)},
  pages={131--133},
  year={2018},
  organization={IEEE}
}

@article{rasch2020signal,
  title={A Signal Adaptive Prediction Filter for Video Coding Using Directional Total Variation: Mathematical Framework and Parameter Selection},
  author={Rasch, Jennifer and Warno, Victor and Pfaff, Jonathan and Tischendorf, Caren and Marpe, Detlev and Schwarz, Heiko and Wiegand, Thomas},
  journal={IEEE Transactions on Image Processing},
  volume={29},
  pages={9678--9688},
  year={2020},
  publisher={IEEE}
}

@inproceedings{NLM,
	title={A non-local algorithm for image denoising},
	author={Buades, Antoni and Coll, Bartomeu and Morel, J-M},
	booktitle={Computer Society Conference on Computer Vision and Pattern Recognition (CVPR)},
	volume={2},
	pages={60--65},
	year={2005},
	organization={IEEE}
}

@article{NLMinHM,
	title={Inloop filter based on non-local means filter},
	author={Matsumura, Masaaki and Bandoh, Yukihiro and Takamura, Seishi and Jozawa, Hirohisa},
	journal={Joint Collaborative Team on Video Coding (JCT-VC), doc. JCTVC-E206},
	year={Mar. 2011}
}

@article{BM3D,
	title={Image denoising by sparse 3-{D} transform-domain collaborative filtering},
	author={Dabov, Kostadin and Foi, Alessandro and Katkovnik, Vladimir and Egiazarian, Karen},
	journal={IEEE Transactions on Image Processing},
	volume={16},
	number={8},
	pages={2080--2095},
	year={2007},
	publisher={IEEE}
}

@article{NALF,
	title={Low-Rank Based Nonlocal Adaptive Loop Filter for {H}igh {E}fficiency {V}ideo {C}ompression},
	author={Zhang, Xinfeng and Xiong, Ruiqin and Lin, Weisi and Zhang, Jian and Wang, Shiqi and Ma, Siwei and Gao, Wen},
	journal={IEEE Transactions on Circuits and Systems for Video Technology},
	volume={27},
	number={10},
	pages={2177--2188},
	year={2017},
	publisher={IEEE}
}

@inproceedings{zhang2015nonlocal,
	title={Nonlocal adaptive in-loop filter via content-dependent soft-thresholding for {HEVC}},
	author={Zhang, Xinfeng and Lin, Weisi and Wang, Shiqi and Ma, Siwei},
	booktitle={International Symposium on Multimedia (ISM)},
	pages={465--470},
	year={2015},
	organization={IEEE}
}

@article{ma2016nonlocal,
	title={Nonlocal in-loop filter: The way toward next-generation video coding?},
	author={Ma, Siwei and Zhang, Xinfeng and Zhang, Jian and Jia, Chuanmin and Wang, Shiqi and Gao, Wen},
	journal={IEEE MultiMedia},
	volume={23},
	number={2},
	pages={16--26},
	year={2016},
	publisher={IEEE}
}

@inproceedings{SANF,
	title={Structure-driven adaptive non-local filter for {H}igh {E}fficiency {V}ideo {C}oding ({HEVC})},
	author={Zhang, Jian and Jia, Chuanmin and Zhang, Nan and Ma, Siwei and Gao, Wen},
	booktitle={Data Compression Conference (DCC)},
	pages={91--100},
	year={2016},
	organization={IEEE}
}

@article{K0053,
	title={{CE}2: Noise Suppression Filter},
	author={Chernyak, Roman and Victor, Stepin and Ikonin, Sergey and Chen, Jianle},
	journal={Joint Video Experts Team (JVET), doc. JVET-K0053},
	year={Jul. 2018}
}

@article{K0160,
	title={{CE}2: Non-local Structure-based Filter},
	author={Meng, Xuewei and Jia, Chuanmin and Wang, Shanshe and Ma, Siwei and Zheng, Xiaozhen},
	journal={Joint Video Experts Team (JVET), doc. JVET-K0160},
	year={Jul. 2018}
}

@article{K0236,
	title={{CE}2.5.2: Non-local mean in-loop filter},
	author={Lai, Chen-Yen and Chen, Ching-Yeh and Huang, Yu-Wen and Lei, Shaw-Min},
	journal={Joint Video Experts Team (JVET), doc. JVET-K0236},
	year={Jul. 2018}
}

@inproceedings{meng2018optimized,
  title={Optimized non-local in-loop filter for video coding},
  author={Meng, Xuewei and Jia, Chuanmin and Wang, Shanshe and Zheng, Xiaozhen and Ma, Siwei},
  booktitle={Picture Coding Symposium (PCS)},
  pages={233--237},
  year={2018},
  organization={IEEE}
}

@article{zhang2017high,
	title={High-efficiency image coding via near-optimal filtering},
	author={Zhang, Xinfeng and Wang, Shiqi and Zhang, Yabin and Lin, Weisi and Ma, Siwei and Gao, Wen},
	journal={IEEE signal processing letters},
	volume={24},
	number={9},
	pages={1403--1407},
	year={2017},
	publisher={IEEE}
}

@inproceedings{zheng2011directional,
  title={Directional adaptive loop filter for video coding},
  author={Zheng, Yunfei and Yin, Peng and Xu, Qian and Sole, Joel and Lu, Xiaoan},
  booktitle={IEEE International Conference on Image Processing (ICIP)},
  pages={3501--3504},
  year={2011},
  organization={IEEE}
}

@inproceedings{erfurt2018multiple,
  title={Multiple feature-based classifications adaptive loop filter},
  author={Erfurt, Johannes and Lim, Wang-Q and Schwarz, Heiko and Marpe, Detlev and Wiegand, Thomas},
  booktitle={Picture Coding Symposium (PCS)},
  pages={91--95},
  year={2018},
  organization={IEEE}
}

@inproceedings{watanabe2009loop,
  title={In-loop filter using block-based filter control for video coding},
  author={Watanabe, Takashi and Wada, Naofumi and Yasuda, Goki and Tanizawa, Akiyuki and Chujoh, Takeshi and Yamakage, Tomoo},
  booktitle={IEEE International Conference on Image Processing (ICIP)},
  pages={1013--1016},
  year={2009},
  organization={IEEE}
}

@inproceedings{liu2012adaptive,
  title={Adaptive post-filtering based on Local Binary Patterns},
  author={Liu, Ying and He, Dake and Fieguth, Paul},
  booktitle={IEEE International Conference on Image Processing (ICIP)},
  pages={2905--2908},
  year={2012},
  organization={IEEE}
}

@article{jia2020fast,
  title={Fast Non-Local Adaptive In-Loop Filter Optimization on {GPU}},
  author={Jia, Chuanmin and Luo, Falei and Zhang, Xinfeng and Wang, Shiqi and Wang, Shanshe and Ma, Siwei},
  journal={IEEE Transactions on Multimedia},
  volume={23},
  pages={39--51},
  year={2020},
  publisher={IEEE}
}

@article{li2020unified,
  title={Unified Intra Mode Coding Based on Short and Long Range Correlations},
  author={Li, Junru and Wang, Meng and Zhang, Li and Zhang, Kai and Liu, Hongbin and Wang, Shiqi and Ma, Siwei and Gao, Wen},
  journal={IEEE Transactions on Image Processing},
  volume={29},
  pages={7245--7260},
  year={2020},
  publisher={IEEE}
}

@article{M0255,
	title={{AHG}11: {MMVD} without Fractional Distances for {SCC}},
	author={Liu, Hongbin and Zhang, Li and Zhang, Kai and others},
	journal={Joint Video Experts Team (JVET), doc. JVET-M0255},
	year={Jan. 2019}
}

@article{ma2022evolution,
  title={Evolution of {AVS} video coding standards: twenty years of innovation and development},
  author={Ma, Siwei and Zhang, Li and Wang, Shiqi and Jia, Chuanmin and Wang, Shanshe and Huang, Tiejun and Wu, Feng and Gao, Wen},
  journal={Science China Information Sciences},
  volume={65},
  number={9},
  pages={1--24},
  year={2022},
  publisher={Springer}
}

@inproceedings{meng2022parametric,
  title={Parametric Non-local In-loop Filter for Future Video Coding},
  author={Meng, Xuewei and Jia, Chuanmin and Zhang, Xinfeng and Lei, Meng and Wang, Shanshe and Li, Lin and Ma, Siwei},
  booktitle={Data Compression Conference (DCC)},
  pages={474--474},
  year={2022},
  organization={IEEE}
}

@inproceedings{lei2022coarse,
  title={Coarse-to-fine Prediction With Local and Nonlocal Correlations for Intra Coding},
  author={Lei, Meng and Meng, Xuewei and Jia, Chuanmin and Wang, Shanshe and Cheng, Zhipeng and Ma, Siwei},
  booktitle={Data Compression Conference (DCC)},
  pages={460--460},
  year={2022},
  organization={IEEE}
}

@article{meng2021spatio,
  title={Spatio-Temporal Correlation Guided Geometric Partitioning for Versatile Video Coding},
  author={Meng, Xuewei and Jia, Chuanmin and Zhang, Xinfeng and Wang, Shanshe and Ma, Siwei},
  journal={IEEE Transactions on Image Processing},
  volume={31},
  pages={30--42},
  year={2021},
  publisher={IEEE}
}

@article{zhang2022implicitly,
  title={Implicitly Selected Transform for {AVS3}},
  author={Zhang, Yuhuai and Zhang, Kai and Zhang, Li and Wang, Shanshe and Gao, Wen},
  journal={IEEE Transactions on Image Processing},
  volume={31},
  pages={1298--1310},
  year={2022},
  publisher={IEEE}
}

@inproceedings{meng2020edge,
  title={Edge-directed geometric partitioning for versatile video coding},
  author={Meng, Xuewei and Zhang, Xinfeng and Jia, Chuanmin and Li, Xia and Wang, Shanshe and Ma, Siwei},
  booktitle={IEEE International Conference on Multimedia and Expo (ICME)},
  pages={1--6},
  year={2020},
  organization={IEEE}
}

@article{lei2022joint,
  title={Joint Local and Nonlocal Progressive Prediction for Versatile Video Coding},
  author={Lei, Meng and Luo, Falei and Zhang, Xinfeng and Wang, Shanshe and Ma, Siwei},
  journal={IEEE Transactions on Image Processing},
  volume={31},
  pages={2824--2838},
  year={2022},
  publisher={IEEE}
}

@inproceedings{lin2022high,
  title={High-Order Intra Prediction for Future Video Coding},
  author={Lin, Kai and Zhang, Jiaqi and Li, Junru and Jia, Chuanmin and Gao, Wen},
  booktitle={2022 Data Compression Conference (DCC)},
  pages={349--358},
  year={2022},
  organization={IEEE}
}

@inproceedings{meng2021optimizedNNN,
  title={Optimized Adaptive Loop Filter in Versatile Video Coding},
  author={Meng, Xuewei and Zhang, Jiaqi and Jia, Chuanmin and Xinfeng, Zhang and Shanshe, Wang and Siwei, Ma},
  booktitle={Data Compression Conference (DCC)},
  pages={359--359},
  year={2021},
  organization={IEEE}
}

@article{han2021technical,
  title={A technical overview of AV1},
  author={Han, Jingning and Li, Bohan and Mukherjee, Debargha and Chiang, Ching-Han and Grange, Adrian and Chen, Cheng and Su, Hui and Parker, Sarah and Deng, Sai and Joshi, Urvang and others},
  journal={Proceedings of the IEEE},
  volume={109},
  number={9},
  pages={1435--1462},
  year={2021},
  publisher={IEEE}
}

\begin{IEEEbiography}[{\includegraphics[width=1in,height=1.25in,clip,keepaspectratio]{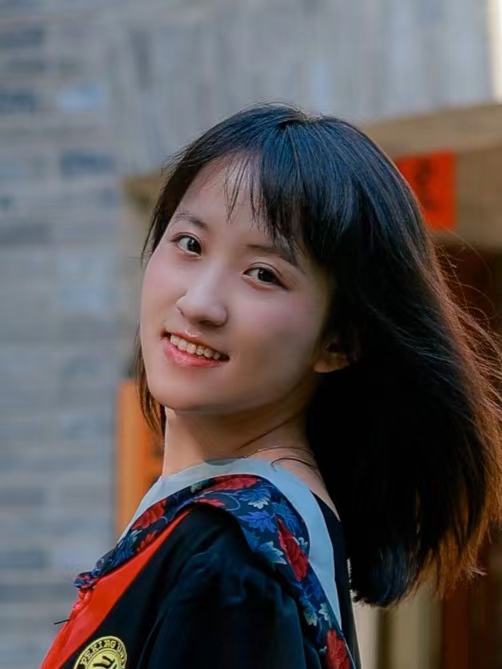}}]{Xuewei Meng}
			received the B.E. degree in communication engineering from Beijing University of Posts and Telecommunications, Beijing, China, in 2017 and the Ph.D. degree in computer application technology from Peking University, Beijing, China, in 2022. She is currently a senior software engineer with Core Media Technology, Disney Streaming. Her research interests include video processing, video compression, and video coding standard. She is actively participating in the research of \textit{Versatile Video Coding}~(VVC) standard. 
		\end{IEEEbiography}
	
		\begin{IEEEbiography}[{\includegraphics[width=1in,height=1.25in,clip,keepaspectratio]{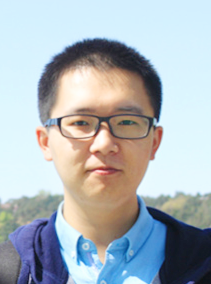}}]{Chuanmin Jia}
			received the B.E. degree in computer science from Beijing University of Posts and Telecommunications, Beijing, China, in 2015 and the Ph.D. degree in computer application technology from Peking University, Beijing, China, in 2020. He was a visiting student in New York University, USA, in 2018. He is currently an Assistant Professor with the Wangxuan Institute of Computer Technology, Peking University. His research interests include video compression and multimedia signal processing. He received the Best Paper Award of PCM 2017, Best Paper Award of IEEE MM 2018, and Best Student Paper Award of IEEE MIPR 2019.
		\end{IEEEbiography}
	
		\begin{IEEEbiography}[{\includegraphics[width=1in,height=1.25in,clip,keepaspectratio]{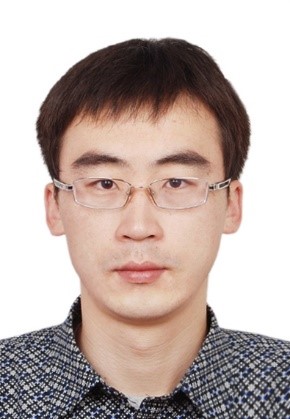}}]{Xinfeng Zhang} (M'16-SM'20) received the B.S. degree in computer science from the Hebei University of Technology, Tianjin, China, in 2007, and the Ph.D. degree in computer science from the Institute of Computing Technology, Chinese Academy of Sciences, Beijing, China, in 2014. From 2014 to 2017, he was a Research Fellow with the Rapid-Rich Object Search Lab, Nanyang Technological University, Singapore. From Oct. 2017 to Oct. 2018, he was a Post-Doctoral Fellow with  the School of Electrical Engineering System, University of Southern California, Los Angeles, CA, USA. From Dec. 2018 to Aug. 2019, he was a Research Fellow with the department of Computer Science, City University of Hong Kong. 
  
        He currently is an Assistant Professor with the School of Computer Science and Technology, University of Chinese Academy of Sciences. He authored more than 100 refereed journal/conference papers and received the Best Paper Award of IEEE Multimedia 2018, the Best Paper Award at the 2017 Pacific-Rim Conference on Multimedia (PCM) and the Best Student Paper Award in IEEE International Conference on Image Processing 2018. His research interests include video compression and processing, image/video quality assessment, and 3D point cloud processing.
			
		\end{IEEEbiography}

		\begin{IEEEbiography}[{\includegraphics[width=1in,height=1.25in,clip,keepaspectratio]{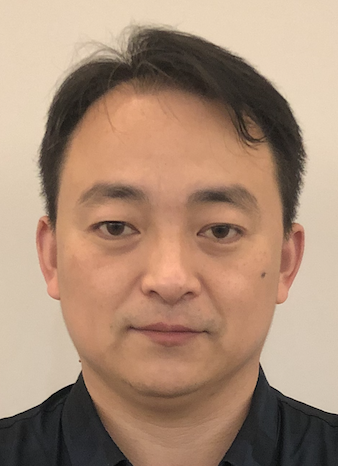}}]{Shanshe Wang}
			received the B.S. degree from the Department of Mathematics, Heilongjiang University, Harbin, China, in 2004, the M.S. degree in computer software and theory from Northeast Petroleum University, Daqing, China, in 2010, and the Ph.D. degree in computer science from the Harbin Institute of Technology. He held a postdoctoral position with Peking University from 2016 to 2018. He joined the School of Electronics Engineering and Computer Science, Institute of Digital Media, Peking University, Beijing, where he is currently a Research Associate Professor. His current research interests include video compression and image and video quality assessment.
		\end{IEEEbiography}
	
		\begin{IEEEbiography}[{\includegraphics[width=1in,height=1.25in,clip,keepaspectratio]{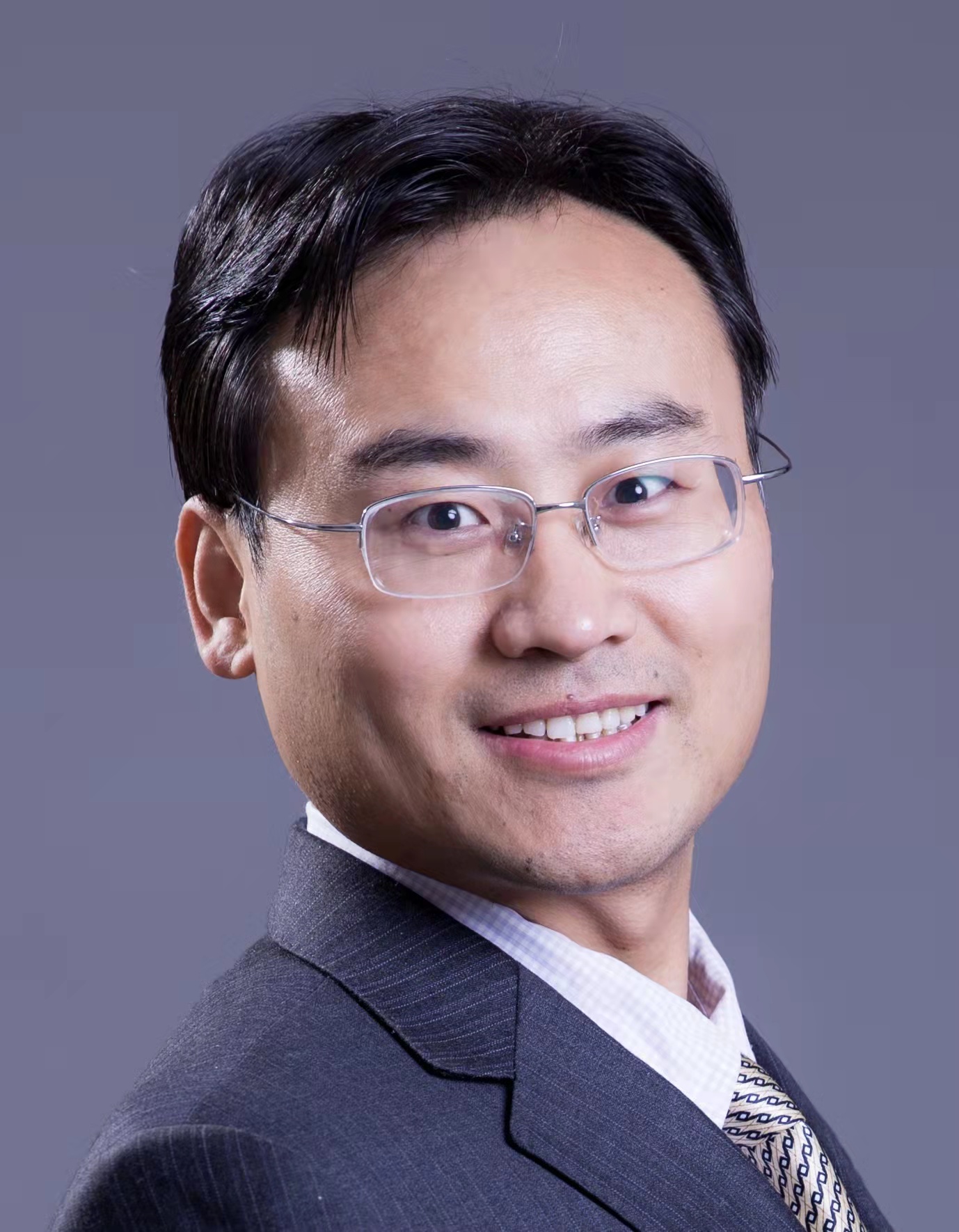}}]{Siwei Ma} (Senior Member, IEEE) received the B.S. degree from Shandong Normal University, Jinan, China, in 1999, and the Ph.D. degree in computer science from the Institute of Computing Technology, Chinese Academy of Sciences, Beijing, China, in 2005. He held a postdoctoral position with the University of Southern California, Los Angeles, CA, USA, from 2005 to 2007. He joined the School of Electronics Engineering and Computer Science, Institute of Digital Media, Peking University, Beijing, where he is currently a Professor. He has authored over 300 technical articles in refereed journals and proceedings in image and video coding, video processing, video streaming, and transmission. He served/serves as an Associate Editor for the IEEE TRANSACTIONS ON IMAGE PROCESSING, the IEEE TRANSACTIONS ON CIRCUITS AND SYSTEMS FOR VIDEO TECHNOLOGY, and the Journal of Visual Communication and Image Representation.
		\end{IEEEbiography}
\end{document}